\documentclass[10pt]{article}
\usepackage[preprint]{tmlr}

\usepackage{amsthm}

\providecommand{\R}{\mathbb{R}}

\providecommand{\E}{\mathbb{E}}
\providecommand{\EE}{\mathbb{E}}

\providecommand{\Law}{\operatorname{Law}}

\providecommand{\Pcal}{\mathcal{P}}
\providecommand{\push}{_{\#}}

\providecommand{\T}{\mathbb{T}}      
\providecommand{\D}{\mathcal{D}}     

\providecommand{\Hcal}{\mathcal{H}}

\providecommand{\Lip}{\operatorname{Lip}}

\providecommand{\Wone}{W_1}
\providecommand{\Wtwo}{W_2}

\providecommand{\Scal}{\mathcal{S}}  
\providecommand{\Tcal}{\mathcal{T}}  

\providecommand{\PLeq}[1]{P_{\le #1}}

\providecommand{\cO}{\mathcal{O}}

\providecommand{\argmin}{\operatorname*{arg\,min}}

\theoremstyle{plain}
\newtheorem{theorem}{Theorem}[section]
\newtheorem{proposition}[theorem]{Proposition}
\newtheorem{lemma}[theorem]{Lemma}
\newtheorem{corollary}[theorem]{Corollary}

\theoremstyle{definition}
\newtheorem{definition}[theorem]{Definition}

\theoremstyle{remark}
\newtheorem{remark}[theorem]{Remark}

\usepackage{amsmath,amssymb,amsfonts,amsthm,mathtools}

\usepackage{graphicx}
\usepackage{subcaption}
\usepackage{booktabs}
\usepackage{multirow}
\usepackage{makecell}
\usepackage{placeins}

\usepackage{colortbl}

\usepackage{algorithm}
\usepackage{algpseudocode}

\definecolor{linkblue}{HTML}{0000FF}
\definecolor{urlviolet}{HTML}{7A1E76}

\usepackage{url}
\usepackage[colorlinks=true,linkcolor=linkblue,citecolor=linkblue,urlcolor=urlviolet]{hyperref}
\usepackage[nameinlink,capitalize,noabbrev]{cleveref}

\definecolor{bestblue}{RGB}{220,235,255}
\definecolor{secondorange}{RGB}{255,235,220}
\newcommand{\bestcell}[1]{\cellcolor{bestblue}#1}
\newcommand{\secondcell}[1]{\cellcolor{secondorange}#1}

\title{Rectified Flows for Fast Multiscale Fluid Flow Modeling}

\author{
\name Victor Armegioiu \email victor.armegioiu@math.ethz.ch \\
\addr Department of Mathematics, ETH Z\"urich, Z\"urich, Switzerland
\AND
\name Yannick Ramic \email yannick.ramic@math.ethz.ch \\
\addr Department of Mathematics, ETH Z\"urich, Z\"urich, Switzerland
\AND
\name Siddhartha Mishra \email siddhartha.mishra@math.ethz.ch \\
\addr Department of Mathematics, ETH Z\"urich, Z\"urich, Switzerland
}

\def\month{MM}
\def\year{YYYY}
\def\openreview{\url{https://openreview.net/forum?id=XXXX}}

\begin{document}

\maketitle

\begin{abstract}
Statistical surrogate modeling of fluid flows is challenging due to multiscale dynamics and strong sensitivity to initial conditions. Conditional diffusion models can achieve high fidelity, but typically require hundreds of stochastic steps at inference.
We introduce a rectified-flow surrogate that learns a time-dependent conditional velocity field transporting input-to-output laws along nearly straight trajectories. Sampling reduces to solving a deterministic ODE along this learned transport, so each function evaluation is substantially more effective: on multi-scale 2D benchmarks we match diffusion-class posterior statistics with as few as $8$ ODE steps versus $\ge\!128$ steps for score-based diffusion.

On the theory side, we develop a law-level analysis tailored to conditional PDE forecasts.
First, we formalize the link between our evaluation criterion—one-point Wasserstein distances on fields—and the $k\!=\!1$ correlation-marginal viewpoint in statistical solutions.
Second, we provide a one-step error decomposition for the learned pushforward law into a \emph{coverage} (high-frequency tail) term controlled by structure functions (equivalently, by spectral decay), and a \emph{fit} term controlled directly by the training objective.
Third, we show how \emph{straightness} in rectification time governs local truncation error for ODE sampling, yielding step-count requirements and explaining why rectified transports admit large, stable steps.

Guided by this picture, we propose a curvature-aware sampler that tracks an EMA-based straightness proxy and adaptively blends and steps the velocity during inference.
Across multiscale incompressible and compressible 2D flows, our method matches diffusion models in Wasserstein statistics and energy spectra, preserves fine-scale structure missed by MSE baselines, and delivers high-resolution conditional samples at a fraction of the inference cost.
\end{abstract}

\section{Introduction}

\noindent
Partial differential equations (PDEs) \cite{evans2010partial} are the backbone of modeling physical phenomena, from atmospheric dynamics to aerodynamics and magnetohydrodynamics. Currently, solutions of PDEs are simulated with high-fidelity numerical solvers such as finite difference, finite element, and spectral methods \cite{NAbook}. These are often prohibitively expensive for \emph{many–query} tasks \cite{ROMbook} such as uncertainty quantification, optimization, and real‐time prediction. In particular, these solvers encounter a significant challenge for classes of non-linear PDEs, whose solutions exhibit extreme sensitivity to initial and boundary data, leading to chaotic behavior and complex multi‐scale features. In such settings, one seeks, not a deterministic trajectory, but rather the \emph{statistical solution} of the PDE, i.e., the push-forward of the underlying probability measure on inputs (initial and boundary conditions etc) by the PDE solution operator \cite{FLM1}. Currently, these statistical solutions are sampled with ensemble Monte Carlo method by repeatedly calling high‐fidelity PDE solvers \cite{LMP1}. Given the computational cost of the PDE solver, ensemble methods  become infeasible as ensemble sizes grow.

Machine learning and AI based algorithms are increasingly being explored for the data-driven simulation of PDEs \cite{scimlrev2}. In particular, \emph{neural operators} \cite{kovachki2023neural}, which directly seek to learn the PDE solution operator from data, are already widely used in a PDE context \cite{scimlrev2}. Examples of neural operators have been proposed in \cite{FNO,GNO,DeepONet,wu2024transolver,hao2023gnot,li2023geometryinformed,UPT} and references therein.

However, these neural operators have been found to be unsuitable for simulating PDE with chaotic, multiscale solutions such as governing equations of fluid dynamics. For instance, recently in \cite{molinaro2024generative}, it was shown both theoretically and empirically that neural operators, trained to minimize mean absolute (square) errors, will always collapse to the ensemble mean when used for simulating statistics of PDEs with chaotic multiscale solutions.

Given this context, generative AI models—especially score‐based and diffusion models—have recently shown remarkable success at learning complex, high‐dimensional distributions by progressive denoising score matching~\cite{song2020score}. In the context of chaotic PDEs, \emph{GenCFD}~\cite{molinaro2024generative} introduced a conditional diffusion framework that preserves spectral content across fine scales, mitigating the “spectral collapse’’ of naive MSE‐trained surrogates, see also \cite{oommen2024integrating,gao2024bayesian,gao2024generative,gao2024generative2}. However, sampling a score-based diffusion model requires simulating a reverse‐time stochastic differential equation (SDE) over 100+ discretization steps, each invoking a large ML model on a moderate-to-high resolution spatial grid, resulting in high runtimes that undermine real‐time or resource‐constrained applications.

To overcome this limitation of diffusion models, we adapt \emph{rectified flows}~\cite{liu2022flow}: a deterministic ODE‐based analogue of diffusion that “straightens’’ transport paths in probability space. Rather than small random increments, rectified flows solve an ODE whose velocity field is trained to match the diffusion model’s instantaneous score, but whose trajectories remain nearly linear. This structure permits large time integration, reducing the number of solver steps by an order of magnitude or more, while provably recovering the same target distribution.

\paragraph{What is evaluated, and why.}
Because statistical solutions are measures on function spaces, the most robust comparison is via \emph{observables} or \emph{correlation marginals} \cite{lanthaler2021statistical}. Our evaluation metric in Section~\ref{sec:em}—the \emph{one-point Wasserstein distance}—is exactly the Wasserstein discrepancy between $k\!=\!1$ correlation marginals (pointwise push-forwards), averaged over space. This aligns the experimental metrics (Table~\ref{tab:results}) with the law-level theoretical statements in Sections~\ref{sec:math-form}--\ref{sec:spectral-capacity}.

Hence, our main contributions are:
\begin{itemize}
\item \emph{Rectified flows for conditional PDEs.}
We learn a time-dependent velocity field that transports a \emph{conditional law} of fields
from a simple reference (Gaussian noise conditioned on the input) to the target PDE pushforward law.
The key theoretical point is \emph{correctness at the level of laws}:
the \emph{barycentric rectified velocity} induces an ODE whose flow maps the initial law to the target law
(\textbf{SM}~\ref{app:theory:rf-correctness}).

\item \emph{Curvature-aware integration.}
The sampling cost is governed by the amount of \emph{time-bending} (straightness/curvature in rectification time)
of the learned vector field along typical trajectories.
We formalize: (i) which curvature quantity controls discretization error,
(ii) why a simple EMA deviation is a principled proxy for that curvature,
and (iii) why the blending step is the minimizer of a quadratic (Tikhonov) correction
(\textbf{SM}~\ref{app:theory:discretization}).

\item \emph{Multi-scale accuracy at speed.}
At fixed grid resolution there is an \emph{irreducible} law-level error floor coming from unresolved Fourier scales.
We express this as a \emph{coverage} term controlled by structure functions / spectral tails
(Cref{app:sec:coverage-structure}).
Rectified flows reduce curvature so that the best attainable (bandlimited) law
is reached with far fewer sampling steps. This is precisely what is seen in
Table~\ref{tab:results} and in the spectral evolution plots (Figure~\ref{fig:spec_s1_E}).
\end{itemize}

\section{Approach}

\noindent
\textbf{Modeling multi-scale flows and their statistical solutions.}
We focus on time-dependent PDEs of the general form
\begin{align}
\label{eq:general-pde}
\partial_t\,u(x,t)\;+\;\mathcal{L}\!\bigl(u,\nabla_x u,\nabla_x^2u,\dots\bigr) \;=\; 0,
\quad x \in D \subset \mathbb{R}^d, \; t\in (0,T),
\end{align}
subject to boundary conditions $\mathcal{B}(u) = u_b$ on $\partial D \times (0,T)$ and initial data $u(x,0) = \bar{u}(x)$. Here, the solution $u:D \mapsto {\mathbb R}^m$ is evolved in time, with respect to the differential operator $\mathcal{L}$. Concrete examples of \eqref{eq:general-pde} are the incompressible Navier-Stokes and compressible Euler equations.

A recurring theme in non-linear, multi-scale fluid flows is \emph{extreme sensitivity} to initial or boundary data, which can lead to chaotic or near-chaotic behavior.
Rather than fix a single $u_0$, one may consider an \emph{initial measure} $\mu_0$ to represent uncertain or variable initial states, and then study the \emph{statistical solution} $\mu_t = \mathcal{S}^t_{\#}\mu_0$ pushed forward by the PDE solution operator $\mathcal{S}^t$.
In strongly nonlinear regimes, $\mu_t$ spreads significantly in function space, creating complex, multi-scale distributions of possible flow fields.
Accurately approximating this evolving distribution is essential for uncertainty quantification and design under rough conditions, but doing so with high-fidelity PDE solvers can be prohibitively expensive when large ensembles are required.

\noindent
\textbf{High cost of computing statistical solutions.}
In practice, characterizing $\mu_t$ directly often requires performing \emph{many} forward simulations of the PDE solver, each initialized with a slightly perturbed $u_0$.
As each simulation might require high-resolution discretizations (e.g.\ millions of degrees of freedom) and multiple time steps, the overall cost for constructing a sizable ensemble rapidly becomes prohibitive.
Even with modern high-performance computing platforms, running hundreds or thousands of high-fidelity statistical simulations can be infeasible.
Hence, there is a pressing need for more efficient \emph{surrogate models} that can \emph{sample} physically realistic final states without enumerating every realization of the forward PDE solve. Succinctly put, we aim to provide a solution to the following problem
\begin{center}
\textit{How can we efficiently capture statistical quantities of interest for the pushforward measure
$S^t_\# \mu_0$ for highly sensitive solution operators $S^t$?}
\end{center}

\noindent
\textbf{Observables and correlation marginals (what metrics actually probe).}
A classical way to formalize ``quantities of interest'' for $\mu_t$ is via
observables and their induced correlation marginals \cite{lanthaler2021statistical}.
For $k=1$, the pointwise evaluation map at $x\in D$,
\[
\mathrm{ev}_x:\;u\mapsto u(x)\in \R^m,
\]
pushes $\mu_t$ to the \emph{one-point marginal} $(\mathrm{ev}_x)_\#\mu_t$ on $\R^m$ -- strictly speaking, one studies the pushforwards measures  $\mathrm{ev}_x \circ i_\epsilon$, here $(i_\epsilon u) := (\phi_\epsilon \ast u)$ where $\phi_\epsilon$ is a compactly supported mollifier, so that point-wise evaluation is well-defined; for the sake of brevity, we will not make this distinction explicit in the sequel. 
Our one-point Wasserstein metric (Section~\ref{sec:em}) is the spatial average of
$W_1\big((\mathrm{ev}_x)_\#\mu_t,(\mathrm{ev}_x)_\#\widehat\mu_t\big)$.
By Kantorovich--Rubinstein duality, controlling this quantity controls the discrepancy of \emph{all} pointwise 1-Lipschitz observables, averaged over space. This is the bridge between Table~\ref{tab:results} and the law-level bounds in Section~\ref{sec:math-form} and Section~\ref{sec:spectral-capacity}.

\noindent
\textbf{Rectified flows for faster sampling.}
To mitigate these inefficiencies, we adopt the framework of \emph{rectified flows} \cite{liu2022flow}: a deterministic alternative to conditional diffusion designed to reduce the number of required integration steps.
In essence, rather than generating samples via small random “diffusion-reversal” increments, rectified flows construct an \emph{ordinary differential equation} (ODE) with \emph{nearly straight} trajectories in function space.
This straighter transport map can be integrated more aggressively in time, enabling sampling with far fewer solver steps.
Importantly, rectified flows retain the desirable property of matching the same final distributions that standard diffusion models learn, but at a fraction of the runtime cost.

\begin{figure}[h]
  \centering
  \includegraphics[width=1.0\linewidth]{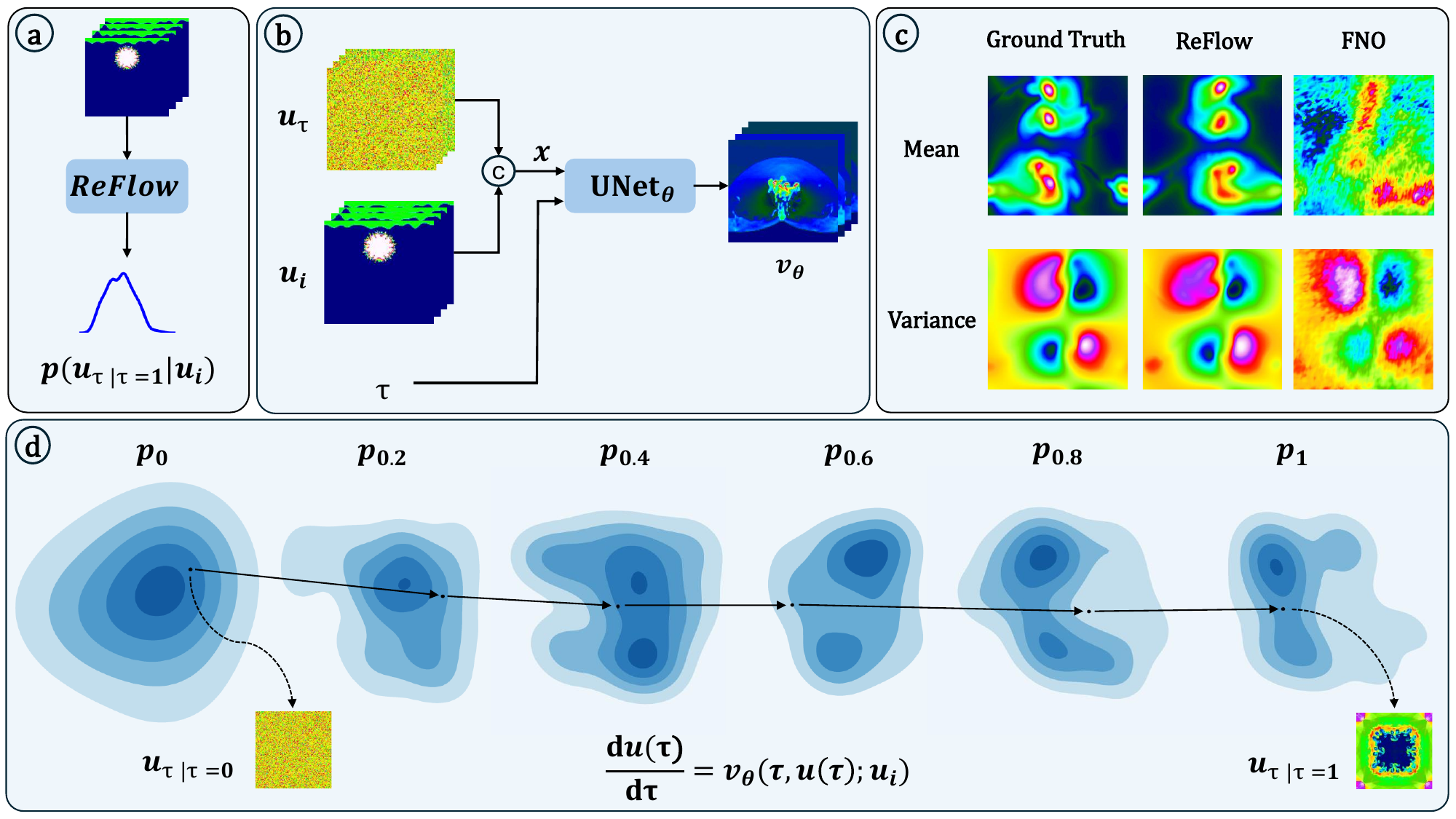}
  \caption{\textbf{Overview of ReFlow.} (\textbf{a}): ReFlow closely matches high-order statistical moments (in the Wasserstein $W_1$-sense) of fluid flows by learning a neural surrogate to approximate the push-forward of the distribution $p(u_{\tau|\tau=0} \vert u_i)$, while minimizing inference-time computational costs. (\textbf{b}) Macro-architecture overview: A UViT based network predicts the velocity field $v_\theta$, conditioned on Gaussian noise, initial data $u_i$ and boundary conditions, and diffusion time $\tau$. Noise and initial / boundary conditions are concatenated as inputs. (\textbf{c}): Results for the cylindrical shear flow dataset. (\textbf{d}): The KDE plots monitor the evolution of a down-projected version of the push-forward measure $p_\tau :=  X_\theta(\tau)_\# p(u_{\tau=0}  \vert u_i) $, where $X_\theta$ is the flow induced by the learned vector field $v_\theta$. Results for the Richtmyer Meshkov dataset showing flow matching between noise ($p_{\tau=0}$) and target ($p_{\tau=1}$) distributions. ReFlow solves an ODE, learning a constant velocity field that enables straight-line inference paths (visualized with principal component analysis and kernel density estimation at different diffusion times).}
  \label{fig:RFT}
\end{figure}

\section{Related Work}

\subsection{Diffusion Models in Fluid Dynamics}

\textbf{Conditional diffusion surrogates.}
\emph{GenCFD}~\cite{molinaro2024generative} trains a conditional score-based
diffusion model that starts from an initial or low-resolution flow field
and iteratively refines it through hundreds of denoising steps, thereby
capturing multi-scale turbulent structure.
Similarly, the method of~\cite{oommen2024integrating} conditions a
diffusion model on a coarse fluid solution (generated by a learned
operator) and ''super-resolves'' the fine-scale flow.
Given these similarities, we refer to such diffusion-based solvers
collectively as \emph{conditional diffusion PDE solvers}.

\textbf{Rectified flow alternative.}
Each sample typically needs hundreds of network calls, limiting
large-scale or real-time use. Our method replaces the stochastic SDE with a deterministic ODE whose
nearly straight trajectories require far fewer steps, achieving similar
statistical fidelity at much lower inference cost (see
Sections~\ref{sec:comp-efficiency} and~\ref{sec:experiments}). Crucially, we find that rectified flows are actually more efficient from the standpoint of training-time required to reach the same accuracy level with the conditional diffusion approaches; this is also confirmed by earlier work \cite{esser2024scaling} exploring the scalability of rectified flows with transformer-based backbones.

\section{Rectified Flows: An ODE-Based Alternative}

\subsection{Mathematical Formulation}
\label{sec:math-form}

\noindent
\paragraph{Overview.}
Rectified flows provide a \emph{deterministic} way to transform one probability distribution into another, offering an alternative to stochastic diffusion-based methods. Instead of traversing curved, noise-driven paths, rectified flows attempt to transport mass along nearly ``straight'' trajectories in state space, which can dramatically reduce sampling costs.

Throughout, our random variables are \emph{state fields}
\(
u : \D \!\to\! \mathbb R^{m}
\)
(e.g.\ $m=2$ for velocity, or $m=4$ for compressible state variables) living in a separable space
\(
\mathcal U = L^{2}(\D;\mathbb R^{m})
\)
(or \(H^{s}\), \(s>\frac d2\)).
All theory below extends verbatim from \(\mathbb R^{d}\) because \(\mathcal U\) is Polish; in practice we work with a Galerkin discretisation
\(
\mathbb R^{d_N}\subset\mathcal U
\),
so every integral is evaluated by Monte-Carlo on its finite grid surrogate.

In all that follows, the \emph{diffusion time}
\(\tau\in[0,1]\) is a purely \textit{scale-ordering} variable:
\(\tau=0\) denotes the coarsest, lowest–frequency representation of a state field, while \(\tau=1\) restores full spatial resolution; it is \emph{not} the physical PDE time but merely an index that governs the progressive refinement of scales along the rectified flow.

\noindent\textbf{Coupling $\gamma$ as a statistical transport blueprint.}
Let $\mu_0,\mu_1\in\mathcal P(\mathcal U)$ be the initial and target field laws.
Fix \emph{any} coupling $\gamma\in\mathcal P(\mathcal U\times\mathcal U)$ with
$(\pi_0)_\#\gamma=\mu_0$ and $(\pi_1)_\#\gamma=\mu_1$ (in practice: the empirical pairing
of input/output fields in the dataset).
Think of $(U_0,U_1)\sim\gamma$ as drawing a \emph{random chord} in state space, from $U_0$ to $U_1$.

\smallskip
For $\tau\in[0,1]$ define the linear chord point
\[
U_\tau := (1-\tau)U_0+\tau U_1,
\quad
\rho_\tau := \Law(U_\tau) = (T_\tau)_\#\gamma,
\quad
T_\tau(u_0,u_1)=(1-\tau)u_0+\tau u_1 .
\]
Thus $\rho_\tau$ is the population of intermediate states obtained by moving along dataset chords
by a fraction $\tau$.

\smallskip
\noindent\textbf{The barycentric velocity is the “average chord direction” at $(u,\tau)$.}
Along a single chord, the displacement $U_1-U_0$ is constant in $\tau$.
However, many chords pass through the same intermediate state $u$ at the same $\tau$; the correct
population-level velocity is the average direction among those chords. Concretely, define
\begin{equation}\label{eq:barycentric}
v^\star(u,\tau)
\;:=\;
\EE\bigl[\,U_1-U_0 \,\big|\, U_\tau=u\,\bigr],
\quad \rho_\tau\text{-a.e. }u.
\end{equation}
This is the clean “one-line” definition of the barycentric field: it is a conditional expectation,
so it is automatically the unique $L^2(\rho_\tau)$ minimizer of a regression problem.

\begin{remark}[Disintegration form]
\label{rem:disint}
If one prefers a geometric formula, \eqref{eq:barycentric} is equivalent to disintegrating $\gamma$
along the map $T_\tau$:
\[
v^\star(u,\tau)
=
\int_{T_\tau^{-1}(u)} (u_1-u_0)\,\gamma_u^\tau(du_0,du_1),
\quad
T_\tau^{-1}(u)=\{(u_0,u_1):(1-\tau)u_0+\tau u_1=u\}.
\]
Keeping this as a remark avoids leading with fiber notation.
\end{remark}

\smallskip
\noindent\textbf{Rectified-flow training = regression onto constant chord directions.}
Given samples $(u_0,u_1)\sim\gamma$ and $\tau\sim\mathrm{Unif}[0,1]$, form $u_\tau=(1-\tau)u_0+\tau u_1$.
Train $v_\theta$ by
\begin{equation}\label{eq:rf_objective}
\min_\theta\;
\EE\Bigl[
\bigl\|\,v_\theta(u_\tau,\tau) - (u_1-u_0)\,\bigr\|^2
\Bigr].
\end{equation}
By the $L^2$ optimality of conditional expectation, the population minimizer of \eqref{eq:rf_objective}
is exactly $v^\star$ in \eqref{eq:barycentric}.
The key qualitative point is immediate: the regression target $(u_1-u_0)$ does \emph{not} depend on $\tau$.
So the learned field is pressured to behave as \emph{time-stationary as the data allow} along these chords.
This is the precise sense in which rectified-flow training promotes “straight” transports without separately
adding an ad hoc straightness penalty.

\smallskip
\noindent\textbf{Sampling = integrate the learned transport ODE.}
Once $v_\theta$ approximates $v^\star$, sampling amounts to integrating
\begin{equation}\label{eq:ode_sampling}
\frac{du_\tau}{d\tau}=v_\theta(u_\tau,\tau),
\quad u_0\sim\mu_0,
\end{equation}
and returning $u_1$ at $\tau=1$. In the idealized chord model where $v^\star$ is constant in $\tau$ along
typical trajectories, large steps are accurate; deviations from that ideal appear as \emph{time-variation}
of the velocity along the path, which is exactly what discretization error is sensitive to
(see Lemma~\ref{lem:lte}).

\paragraph{Curvature-aware control: estimate “how non-constant in $\tau$” the velocity is.}
During sampling, write $v_t:=v_\theta(u_t,\tau_t)$ and maintain an EMA trend
\begin{equation}\label{eq:ema}
v_t^{\mathrm{ema}}
=\lambda\,v_{t-1}^{\mathrm{ema}}+(1-\lambda)\,v_t,
\quad \lambda\in(0,1).
\end{equation}
Define the instantaneous straightness/curvature proxy
\[
s_t:=\|v_t-v_t^{\mathrm{ema}}\|_2,
\]
which measures short-horizon time-variation of the velocity along the realized trajectory.
Large $s_t$ signals local bending / stiffness, precisely the regime where explicit large-step integration
incurs higher local truncation error (again, Lemma~\ref{lem:lte}).

\paragraph{A one-line regularized update (closed form).}
When $s_t$ is large, replace $v_t$ by a gently regularized velocity $\tilde v_t$ defined as the minimizer of
\[
\tilde v_t
\;:=\;
\arg\min_{w}\;
\|w-v_t\|_2^2
\;+\;
\eta_t\,\|w-v_t^{\mathrm{ema}}\|_2^2,
\]
which yields the explicit convex blend
\begin{equation}\label{eq:blend}
\tilde v_t=(1-\alpha_t)v_t+\alpha_t v_t^{\mathrm{ema}},
\quad
\alpha_t=\frac{\eta_t}{1+\eta_t}\in[0,1].
\end{equation}
Choosing $\eta_t$ as a saturating function of $s_t$ makes the correction dormant when the path is straight
and active only when curvature spikes.

\paragraph{Adaptive step size from the same signal.}
Since explicit Euler LTE scales like $(\Delta\tau)^2$ times a norm of the time-variation of the vector field
(Lemma~\ref{lem:lte}), use $s_t$ as a scale proxy and set
\begin{equation}\label{eq:stepsize}
\Delta\tau_t
\;\propto\;
\frac{1}{\sqrt{\kappa_1 s_t+\kappa_2}},
\quad
\Delta\tau_t\in[\Delta\tau_{\min},\Delta\tau_{\max}],
\end{equation}
with $\kappa_2$ absorbing baseline stiffness (e.g.\ dependence on $\|v_t\|$ and local Lipschitz effects).
The state update is then
\[
u_{t+1}=u_t+\Delta\tau_t\,\tilde v_t.
\]
When the learned transport is locally straight ($s_t$ small), the sampler automatically takes large steps
and behaves like “chord shooting”; when curvature increases, it both damps the direction and contracts the step,
preventing instability while retaining the large-step advantage in the straight regime.

\subsection{Error Decomposition and Bounds}

We now quantify why \eqref{eq:blend}–\eqref{eq:stepsize} help. Write the \emph{ideal} and \emph{learned} flows
\[
\dot u_\tau = v_\star(u_\tau,\tau),\quad
\dot{\hat u}_\tau = v_\theta(\hat u_\tau,\tau),
\]
with the same $u_0$. Assume $v_\theta(\cdot,\tau)$ is $L$-Lipschitz in $u$ and $v_\star\in L^2$.

\begin{theorem}[Terminal error decomposition]
\label{thm:app_gronwall}
Let \(U_\tau\) solve \(\dot U_\tau=v_\star(U_\tau,\tau)\) with \(U_0=u_0\), and let \(\hat u_\tau\) solve \(\dot{\hat u}_\tau=v_\theta(\hat u_\tau,\tau)\) with \(\hat u_0=u_0\).
Define the time–averaged velocity \(\bar v_\theta(u):=\int_0^1 v_\theta(u,\tau)\,d\tau\) and
\[
\varepsilon_{\mathrm{fit}}^2
:=\int_0^1 \mathbb{E}\big\|\bar v_\theta(U_\tau)-v_\star(U_\tau,\tau)\big\|^2 \,d\tau,
\quad
\varepsilon_{\mathrm{curv}}^2
:=\int_0^1 \mathbb{E}\big\|v_\theta(U_\tau,\tau)-\bar v_\theta(U_\tau)\big\|^2 \,d\tau.
\]
Then, for the continuous–time flows,
\begin{equation}
\E\big\|\hat u_1 - u_1\big\|
\;\le\;
e^{L}\,\Big(\varepsilon_{\mathrm{fit}}+\varepsilon_{\mathrm{curv}}\Big).
\end{equation}
\end{theorem}

\paragraph{Interpretation.}
The terminal error splits into a \emph{reconstruction} term ($\varepsilon_{\mathrm{fit}}$) and a \emph{straightness} term ($\varepsilon_{\mathrm{curv}}$). Blending \eqref{eq:blend} shrinks the within-time variation, directly tightening the bound; the adaptive stepping in \eqref{eq:stepsize} triggers corrections where the curvature proxy $s_t$ spikes.

\begin{corollary}[High-probability control of terminal error]
\label{cor:hp-error}
Under the same assumptions as Theorem~\ref{thm:app_gronwall}, suppose in addition that
\[
  \int_0^1
  \mathbb{E}\bigl\|
    v_\theta(U_\tau,\tau) - v_\star(U_\tau,\tau)
  \bigr\|^2\,d\tau
  \;\le\;
  \varepsilon_{\mathrm{fit}}^2,
  \quad
  \int_0^1
  \mathbb{E}\bigl\|
    v_\theta(U_\tau,\tau) - \bar v_\theta(U_\tau)
  \bigr\|^2\,d\tau
  \;\le\;
  \varepsilon_{\mathrm{curv}}^2,
\]
so that both the reconstruction error and straightness error are controlled in $L^2$.
Then there exists a constant $C(L)$ such that
\begin{equation}
  \mathbb{E}\bigl\|\hat u_1 - u_1\bigr\|^2
  \;\le\;
  C(L)\,\bigl(\varepsilon_{\mathrm{fit}}^2+\varepsilon_{\mathrm{curv}}^2\bigr).
\end{equation}
Consequently, for any tolerance $\eta>0$, Chebyshev's inequality yields
\begin{equation}
  \mathbb{P}\bigl(\|\hat u_1-u_1\| > \eta\bigr)
  \;\le\;
  \frac{
    C(L)\,\bigl(\varepsilon_{\mathrm{fit}}^2+\varepsilon_{\mathrm{curv}}^2\bigr)
  }{
    \eta^2
  }.
\end{equation}
In particular, as training drives $\varepsilon_{\mathrm{fit}},\varepsilon_{\mathrm{curv}}\to0$,
the probability of a large terminal error event decays at least quadratically in the target
radius $\eta$.
\end{corollary}

\begin{proof}[Proof sketch]
The proof follows the same Grönwall argument as
Theorem~\ref{thm:app_gronwall}, but carried out in $L^2$ instead of $L^1$,
yielding
$\mathbb{E}\|\hat u_1-u_1\|^2 \le C(L)(\varepsilon_{\mathrm{fit}}^2+\varepsilon_{\mathrm{curv}}^2)$
for some $C(L)\sim e^{2L}$.
Chebyshev's inequality then gives
\(
\mathbb{P}(\|\hat u_1-u_1\|>\eta)
\le \mathbb{E}\|\hat u_1-u_1\|^2 / \eta^2
\),
which is the claimed bound.
A full proof (including the sharp dependence on the straightness functional) is deferred to the theory appendix.
\end{proof}

\begin{lemma}[Local truncation error for explicit Euler on $\dot u=v(u,\tau)$]
\label{lem:lte}
Assume $v\in C^{1}$ in $(u,\tau)$ on the region traversed and that
$\|\partial_\tau v(u,\tau)\|\le M_\tau$, $\|J_u v(u,\tau)\|\le L$, and $\|v(u,\tau)\|\le M$.
Then one explicit-Euler step from $(u,\tau)$ with step size $\Delta\tau$ has local truncation error
\begin{align}
\mathrm{LTE}(\tau;\Delta\tau)
&:= \bigl\|u(\tau+\Delta\tau)-\bigl(u(\tau)+\Delta\tau\,v(u(\tau),\tau)\bigr)\bigr\| \nonumber\\
&= \frac{(\Delta\tau)^2}{2}\,
\bigl\|\partial_\tau v(u,\tau)+J_u v(u,\tau)\,v(u,\tau)\bigr\|
\;+\; \cO\!\bigl((\Delta\tau)^3\bigr) \label{eq:lte-expansion}\\
&\le \frac{(\Delta\tau)^2}{2}\,(M_\tau+LM)
\;+\; \cO\!\bigl((\Delta\tau)^3\bigr). \nonumber
\end{align}
\end{lemma}

\paragraph{From Lemma~\ref{lem:lte} to the step rule.}
The leading LTE term depends on $\|\partial_\tau v\|$ and $\|J_u v\,v\|$. The EMA deviation $s_t=\|v_t-v_t^{\mathrm{ema}}\|$ is a computable proxy for the (local) time variation $\|\partial_\tau v\|$ (up to a scale factor tied to the EMA bandwidth), while $L\,\|v_t\|$ upper-bounds the spatial contribution. Enforcing $\mathrm{LTE}\lesssim\mathrm{tol}$ yields the square-root rule \eqref{eq:stepsize}, with $\kappa_1$ absorbing the EMA bandwidth and $\kappa_2\approx L\|v_t\|$. Under blending, $\|\,\tilde v_t-v_t^{\mathrm{ema}}\|=(1-\alpha_t)s_t$, so the curvature term and the required step shrink \emph{together}, stabilizing integration where needed and enlarging steps elsewhere. See \textbf{SM}~\ref{app:theory:discretization} for proofs and additional theoretical context.

\subsection{Spectral Capacity, Structure Functions, and Law-Level Coverage}
\label{sec:spectral-capacity}

\paragraph{Functional setting.}
We work on the periodic box $\D=\T^d$ and consider $L^2$ fields
$\Hcal:=L^2(\D;\R^m)$ with norm $\|\cdot\|_2$.
For incompressible velocity fields (e.g.\ SL2D, see Section~\ref{sec:experiments}), one may equivalently restrict to divergence-free subspaces and adopt the statistical-solution framework of \cite{lanthaler2021statistical}, which characterizes laws via correlation identities, admissible observables, and an energy inequality.
In that framework the natural distance on laws is the $L^2$-based Wasserstein
metric $W_2$, and small-scale regularity is expressed in terms of
\emph{structure functions} and their time-averages.
The numerical ensembles used in our experiments fit this setting: they are
generated by dissipative solvers on $\T^2$, have uniformly bounded energy,
and exhibit empirical structure-function scaling (equivalently, power-law spectra) consistent with the assumptions below.

\subsubsection{Bandlimited capacity and strain constraints}

Any $u\in\Hcal$ admits a Fourier series
$u(x)=\sum_{k\in\mathbb{Z}^d}\hat u(k)e^{ik\cdot x}$. We write
$P_{\le K}$ for the projector onto modes $\{k:|k|\le K\}$ and
$P_{>K}=I-P_{\le K}$. On a grid of size $N^d$, the Nyquist frequency satisfies
$K_{\max}\simeq N$, so every numerically represented field is effectively
bandlimited to $|k|\lesssim K_{\max}$. Architecturally, downsampling and
anti-aliasing may enforce an even stricter cutoff $K_c\le K_{\max}$.

The next lemma is a standard Bernstein-type inequality: bandlimited fields
have a uniform Lipschitz cap.

\begin{lemma}[Bernstein upper bound for bandlimited fields]
\label{lem:bernstein}
Let $u:\T^d\to\R^m$ satisfy $u=P_{\le K}u$ for some integer $K\ge1$.
Then there exists a constant $C_d>0$ depending only on the dimension $d$ such that
\begin{equation}
  \|\nabla_x u\|_{L^\infty(\T^d)}
  \;\le\;
  C_d\,K^{1+\frac d2}\,\|u\|_{L^2(\T^d)}.
  \label{eq:bernstein-ineq}
\end{equation}
In particular, on an energy shell $\|u\|_{L^2}\le E^{1/2}$ one has
\begin{equation}
  \|\nabla_x u\|_{L^\infty}
  \;\le\;
  C_d\,K^{1+\frac d2}\,E^{1/2}.
  \label{eq:bernstein-energy}
\end{equation}
\end{lemma}

\begin{proof}[Proof sketch]
Write $u(x)=\sum_{|k|\le K}\hat u(k)e^{ik\cdot x}$, so
$\nabla_x u(x)=\sum_{|k|\le K} (ik)\hat u(k)e^{ik\cdot x}$ and
$|\nabla_x u(x)|\le \sum_{|k|\le K} |k|\,|\hat u(k)|$.
Cauchy--Schwarz and a lattice-counting bound $\#\{|k|\le K\}\lesssim K^d$ give
$\|\nabla u\|_{L^\infty}\lesssim ( \sum_{|k|\le K} |k|^2)^{1/2} \|\hat u\|_{\ell^2}
\lesssim K^{1+d/2}\|u\|_{L^2}$.
Full details are deferred to the theory appendix.
\end{proof}

Upper bounds alone do not say what happens if a surrogate \emph{misses} an
entire high-frequency annulus. The next lemma gives a complementary lower bound.

\begin{lemma}[Annulus lower bound]
\label{lem:annulus-lower}
Let $f=P_{[K,2K]}u$ for some $u\in L^2(\T^d)$, where
$P_{[K,2K]}:=P_{\le 2K}-P_{\le K}$ and $K\ge1$.
Then there exists $c_d>0$ such that
\begin{equation}
  \|\nabla_x f\|_{L^\infty}
  \;\ge\;
  c_d\,K\,\|f\|_{L^\infty}
  \;\ge\;
  c_d\,K^{1-\frac d2}\,\|f\|_{L^2}.
  \label{eq:annulus-lower}
\end{equation}
\end{lemma}

\begin{proof}[Proof sketch]
On the annulus $\{K\le|k|\le2K\}$, the symbol of $\nabla$ has magnitude $|k|\in[K,2K]$.
This yields $\|\nabla f\|_{L^\infty}\gtrsim K\|f\|_{L^\infty}$.
A reverse Bernstein estimate on an annulus gives $\|f\|_{L^\infty}\gtrsim K^{-d/2}\|f\|_{L^2}$.
See the theory appendix for a precise statement independent of the Littlewood--Paley cut-off choice.
\end{proof}

\begin{corollary}[Capacity requirement for typical strain levels]
\label{cor:capacity}
Suppose the target flows at time $t$ have typical energy $E$ and typical
rate-of-strain $\|\nabla_x u\|_{L^\infty}\simeq L_\star$.
Let a surrogate produce only bandlimited outputs $u=P_{\le K_c}u$ with
$\|u\|_{L^2}\le E^{1/2}$.
Then necessarily
\begin{equation}
  C_d\,K_c^{1+\frac d2}\,E^{1/2} \;\gtrsim\; L_\star,
\end{equation}
and hence a minimal cutoff
\begin{equation}
  K_c
  \;\gtrsim\;
  \Bigl(\frac{L_\star}{E^{1/2}}\Bigr)^{\frac{2}{2+d}}.
  \label{eq:Kc-lower-bound}
\end{equation}
Moreover, if the true flow has nontrivial energy in an annulus above $K_c$, then
any bandlimited approximation underestimates its strain by at least
$c_d K^{1-d/2}\|P_{[K,2K]}u\|_{L^2}$ for some $K>K_c$.
\end{corollary}

\begin{remark}
Corollary~\ref{cor:capacity} formalises the intuition that spatial capacity
($K_c$) is a hard constraint: at fixed energy scale $E$, there is a minimum
cutoff needed to represent typical strain levels $L_\star$.
In our experiments both GenCFD and ReFlow operate at the same grid resolution,
so they share the same effective $K_c$; rectification does not invent new
small scales but makes the trajectories inside this bandlimited class
straight and easy to integrate.
\end{remark}

\subsubsection{Structure functions and small-scale coverage}

We now quantify how much of the \emph{law} of the PDE solution can be captured
at a given cutoff $K_c$.
For $u:\T^d\to\R^m$ and $r>0$ define the second-order structure function
\[
  S_r^2(u)
  := \frac{1}{|B_r|}\int_{|h|\le r}
      \|u(\cdot+h)-u(\cdot)\|_{L^2(\T^d)}^2\,dh,
\]
and for a probability law $\mu$ on $\Hcal$ set
\[
  S_r^2(\mu)
  := \int S_r^2(u)\,d\mu(u).
\]
In homogeneous turbulence $S_r^2(\mu)$ is directly related to the energy
spectrum; power-law behaviour $S_r^2(\mu)\sim r^{\zeta_2}$ at small $r$ encodes
the scaling of the cascade. In our experiments, Figure~\ref{fig:spec_s1_E} visualizes this connection directly: matching the high-wavenumber tail in the spectrum is equivalent to capturing the small-$r$ behaviour of structure functions, and therefore to controlling the coverage term below.

Guided by the statistical-solution framework of \cite{FLM1},
we assume there exists a modulus of continuity
$\omega:(0,1]\to[0,\infty)$ with $\omega(r)\downarrow0$ as $r\downarrow0$ such
that
\begin{equation}
  S_r^2(\mu_t) \;\le\; \omega(r)
  \quad\text{for all } r\in(0,1],\;t\in[0,T].
  \label{eq:SF-modulus-main}
\end{equation}
In practice, $\omega$ can be taken as a piecewise power law $Cr^{2\alpha}$ calibrated
from the empirical spectra of the numerical ensembles. The next lemma shows that
\eqref{eq:SF-modulus-main} controls the high-frequency tail of the law.

\begin{lemma}[Structure function controls Fourier tail]
\label{lem:SF-tail}
Let $\mu$ be a law on $L^2(\T^d)$ satisfying $S_r^2(\mu)\le\omega(r)$ for all
$r\in(0,1]$, where $\omega(r)\downarrow0$ as $r\downarrow0$.
Then there exist constants $c_d,C_d>0$ such that for all integers $K\ge1$,
\begin{equation}
  \int \|P_{>K}u\|_{L^2}^2\,d\mu(u)
  \;\le\;
  C_d\,\omega\!\bigl(c_d/K\bigr).
  \label{eq:SF-tail}
\end{equation}
\end{lemma}

\begin{proof}[Proof sketch]
Expand $u$ in Fourier series and observe
$\|u(\cdot+h)-u(\cdot)\|_{L^2}^2=\sum_k |\hat u(k)|^2|e^{ik\cdot h}-1|^2$.
Averaging over $h\in B_r$ yields
$S_r^2(u)=\sum_k |\hat u(k)|^2\Psi_r(k)$, where
$\Psi_r(k)\gtrsim \min\{(|k|r)^2,1\}$ uniformly in $k,r$.
For $|k|\gtrsim 1/r$, one has $\Psi_r(k)\gtrsim 1$, hence
$S_r^2(u)\gtrsim \|P_{>\,c/r}u\|_{L^2}^2$.
Integrate over $u\sim\mu$, apply \eqref{eq:SF-modulus-main}, and set $r\sim 1/K$.
Full constants and the Littlewood--Paley formulation are deferred to the theory appendix.
\end{proof}

\begin{lemma}[Bandlimited projection in $W_2$]
\label{lem:projection-W2}
For any law $\mu$ on $L^2(\T^d)$ and cutoff $K_c\ge1$,
\begin{equation}
  W_2\!\bigl(\mu,\,(P_{\le K_c})_\#\mu\bigr)
  \;\le\;
  \Bigl(\int \|P_{>K_c}u\|_{L^2}^2\,d\mu(u)\Bigr)^{1/2}.
  \label{eq:W2-projection}
\end{equation}
\end{lemma}

\begin{proof}
Couple $u\sim\mu$ and $\tilde u=P_{\le K_c}u$. Then $\tilde u\sim (P_{\le K_c})_\#\mu$ and
$W_2^2(\mu,(P_{\le K_c})_\#\mu)\le \EE\|u-\tilde u\|_{L^2}^2=\EE\|P_{>K_c}u\|_{L^2}^2$.
\end{proof}

Combining Lemmas~\ref{lem:SF-tail} and~\ref{lem:projection-W2} gives a clean
coverage bound.

\begin{corollary}[Coverage term from the structure-function modulus]
\label{cor:coverage}
Let $\mu_t$ be the statistical solution law at time $t$ and assume
\eqref{eq:SF-modulus-main}. Then for any cutoff $K_c\ge1$,
\begin{equation}
  W_2\!\bigl(\mu_t,\,(P_{\le K_c})_\#\mu_t\bigr)
  \;\lesssim\;
  \omega\!\bigl(c_d/K_c\bigr)^{1/2},
  \label{eq:coverage-error-main}
\end{equation}
with $c_d$ as in Lemma~\ref{lem:SF-tail}.
\end{corollary}

\subsubsection{Capacity–coverage–fit decomposition for one-step RF}

Finally, we combine the coverage term with the training error of a one-step
rectified-flow surrogate.

Let $S_{\Delta t}:\Hcal\to\Hcal$ denote the PDE solution operator over a
physical step $\Delta t$, and write
\[
\mu_{t+\Delta t} := (S_{\Delta t})\push \mu_t .
\]
Assume the rectified-flow propagator $\Tcal^\theta_{\Delta t}$ aims at the
bandlimited target $\PLeq{K_c}(S_{\Delta t}u)$ in mean-square. Concretely,
sample $u\sim\mu_t$ and an internal noise seed $\xi\sim\mathcal{N}(0,I)$,
and define the RF output $\widehat u := \Tcal^\theta_{\Delta t}(u,\xi)$.
We measure the training fit by
\begin{equation}
  \varepsilon_{\mathrm{train}}^2(t)
  := \EE_{u\sim\mu_t}\,\EE_{\xi\sim\mathcal{N}(0,I)}
  \bigl\|
   \PLeq{K_c}(S_{\Delta t}u) - \widehat u
  \bigr\|_{L^2}^2 .
\end{equation}
Finally, we define the one-step RF law by
\[
\widehat\mu_{t+\Delta t} := \Law(\widehat u).
\]

\begin{proposition}[One-step law-level error: capacity, coverage, and fit]
\label{prop:one-step-capacity}
Under the structure-function modulus assumption \eqref{eq:SF-modulus-main},
the one-step $W_2$ error between the PDE law and the RF law satisfies
\begin{equation}
  W_2\!\bigl(\mu_{t+\Delta t},\,\widehat\mu_{t+\Delta t}\bigr)
  \;\le\;
  \Bigl(
    C_d\,\omega\!\bigl(c_d/K_c\bigr)
    \;+\;
    \varepsilon_{\mathrm{train}}^2(t)
  \Bigr)^{1/2},
  \label{eq:one-step-capacity}
\end{equation}
where the first term is a \emph{coverage} error determined by the
structure-function modulus and the cutoff $K_c$, and the second is a
\emph{fit} error determined by training.
\end{proposition}

\begin{proof}[Proof sketch]
Couple $u\sim\mu_t$ and $\widehat u$ using the same input; then
$W_2^2(\mu_{t+\Delta t},\widehat\mu_{t+\Delta t})\le \EE\|S_{\Delta t}u-\widehat u\|_{L^2}^2$.
Insert $P_{\le K_c}S_{\Delta t}u$, apply the triangle inequality, and bound the tail term
by Lemma~\ref{lem:SF-tail} (applied at time $t+\Delta t$).
\end{proof}

\paragraph{Interpretation and experimental validation.}
At a fixed resolution (fixed $K_c$), Proposition~\ref{prop:one-step-capacity}
says that even a perfectly trained rectified flow cannot beat the coverage term
$\omega(c_d/K_c)^{1/2}$ dictated by small-scale physics.
This is precisely why spectrum matching (Figure~\ref{fig:spec_s1_E}) is a non-negotiable diagnostic on multiscale flows: if the surrogate loses the high-frequency tail, it is \emph{necessarily} paying a coverage penalty in law.
The remaining error is controlled by the training fit and the straightness
properties of the learned velocity (via the continuous-time bounds in
Section~\ref{sec:math-form}).
Rectified flows improve efficiency precisely by reducing the curvature term in
these bounds, so that a given coverage+fit accuracy can be reached with many
fewer ODE steps than a diffusion-based surrogate such as GenCFD (Section~\ref{sec:comp-efficiency}). \\

\section{Correctness and Curvature-Controlled Discretization}
\label{sec:additional_theory}

\paragraph{Organization.}
This section states the main correctness and discretization results that underpin the narrative above.
Full proofs are deferred to the theory appendix (to be written separately); we keep only the steps needed to make the dependence on \emph{fit}, \emph{straightness}, and \emph{discretization} explicit in the main text.

\subsection{Exact transport under the barycentric rectified velocity}
\label{sec:exact_transport_barycentric}

\begin{proposition}[Exact transport under the barycentric velocity]
\label{prop:exact_transport}
Let $\mu_0,\mu_1\in\mathcal{P}(\mathcal{U})$ and let $\gamma\in\mathcal{P}(\mathcal{U}\times\mathcal{U})$
be a coupling of $(\mu_0,\mu_1)$.
Define $T_\tau(u_0,u_1)=(1-\tau)u_0+\tau u_1$ and $\rho_\tau := (T_\tau)_\#\gamma$.
Let $v_\star(\cdot,\tau)$ be the barycentric velocity field of
defined in~\ref{eq:barycentric}.
Assume (e.g.\ on a finite-dimensional Galerkin discretization of $\mathcal{U}$) that
the continuity equation
\begin{equation}
\label{eq:cont_eq_new}
\partial_\tau \rho_\tau + \nabla\!\cdot(\rho_\tau\, v_\star(\cdot,\tau)) = 0
\end{equation}
admits a unique regular Lagrangian flow $X_\tau$ associated to $\dot u_\tau=v_\star(u_\tau,\tau)$.
Then $(X_1)_\#\mu_0=\mu_1$.
\end{proposition}

\begin{proof}[Proof sketch]
By construction, $\rho_\tau$ is the law of $U_\tau=(1-\tau)U_0+\tau U_1$ for $(U_0,U_1)\sim\gamma$.
Differentiate $\EE[\varphi(U_\tau)]$ and condition on $U_\tau$ to identify the weak form of
\eqref{eq:cont_eq_new}. Uniqueness of the continuity equation solution (in the chosen Galerkin setting)
implies $\rho_\tau=(X_\tau)_\#\rho_0$, hence $\rho_1=(X_1)_\#\rho_0$, i.e.\ $\mu_1=(X_1)_\#\mu_0$.
\end{proof}

\textbf{Why this matters.}
Proposition~\ref{prop:exact_transport} isolates a clean correctness statement:
if $v_\theta$ matches the barycentric rectified velocity $v_\star$ sufficiently well (in a sense quantified by Theorem~\ref{thm:app_gronwall}),
the induced flow pushes forward $\mu_0$ to $\mu_1$ at the level of laws.

\subsection{Curvature controls the required number of ODE steps}
\label{sec:curvature_steps}

\begin{corollary}[Global Euler error controlled by time-curvature]
\label{cor:global_euler}
Assume the hypotheses of Lemma~\ref{lem:lte} for $\dot u=v(u,\tau)$ on $\tau\in[0,1]$ and
assume $v(\cdot,\tau)$ is $L$-Lipschitz in $u$ uniformly in $\tau$.
Let $u^\Delta$ be the explicit-Euler approximation with uniform step $\Delta\tau=1/N$.
Then there exists a constant $C=C(L)$ such that
\begin{equation}
\label{eq:global_euler_new}
\sup_{\tau\in[0,1]}\|u^\Delta_\tau-u_\tau\|
\;\le\;
C\,\Delta\tau\,
\int_0^1 \Big(\|\partial_\tau v(u_\tau,\tau)\|+\|J_u v(u_\tau,\tau)\,v(u_\tau,\tau)\|\Big)\,d\tau.
\end{equation}
In particular, when $\int_0^1\|\partial_\tau v(u_\tau,\tau)\|\,d\tau$ is small (``straight'' flow),
the $N$ required to reach a fixed tolerance is proportionally smaller.
\end{corollary}

\begin{proof}[Proof sketch]
Sum the local truncation errors from Lemma~\ref{lem:lte} over $N$ steps and propagate them
using stability of the flow under the $L$-Lipschitz condition via Grönwall.
\end{proof}

\textbf{Connection to our curvature proxy.}
The EMA deviation $s_t=\|v_t-v_t^{\mathrm{ema}}\|$ used in
\eqref{eq:blend}--\eqref{eq:stepsize} is a computable proxy for the same time-variation that drives
\eqref{eq:global_euler_new}. Blending reduces effective curvature; step control reduces $\Delta\tau$ when curvature spikes.
This directly rationalizes why ReFlow can reach the ``spectrum-correct'' regime in very few steps (Figure~\ref{fig:spec_s1_E}), and why the same accuracy level is reached at $8$ steps (Table~\ref{tab:results}) whereas diffusion baselines typically need $\ge 128$.

\subsection{A compact regularity interpretation of structure functions}
\label{sec:besov_interp}

\begin{lemma}[Structure functions imply Besov regularity (heuristic equivalence)]
\label{lem:besov}
Let $u\in L^2(\T^d)$ and define the increment seminorm
\[
[u]_{B^{s}_{2,\infty}}
:=\sup_{0<|h|\le 1}\frac{\|u(\cdot+h)-u(\cdot)\|_{L^2}}{|h|^{s}}.
\]
If the second-order structure function satisfies $S_r^2(u)\lesssim r^{2s}$ for all $r\in(0,1]$,
then $[u]_{B^{s}_{2,\infty}}\lesssim 1$.
Consequently, a law-level modulus bound $S_r^2(\mu)\lesssim r^{2s}$ implies that typical samples from $\mu$
have Besov smoothness $s$, consistent with the Fourier-tail decay in Lemma~\ref{lem:SF-tail}.
\end{lemma}

\begin{remark}
In turbulence one often observes $S_r^2(\mu)\sim r^{\zeta_2}$ over an inertial range.
Heuristically this corresponds to $s\approx \zeta_2/2$, so the coverage term in
Corollary~\ref{cor:coverage} can be read as a direct proxy for inertial-range scaling.
\end{remark}

\subsection{Master inequality}
\label{sec:master_bound_new}

\begin{theorem}[Law-level error: coverage + fit + straightness + discretization]
\label{thm:master}
Fix an effective resolution cutoff $K_c$ and assume the structure-function modulus hypothesis used in
Corollary~\ref{cor:coverage}.
Let $\hat\mu_1^{(N)}$ be the law obtained by taking $N$ explicit-Euler steps to integrate the learned rectified ODE
$\dot u_\tau=v_\theta(u_\tau,\tau)$ from $\tau=0$ to $\tau=1$.
Assume $v_\theta(\cdot,\tau)$ is $L$-Lipschitz in $u$ uniformly in $\tau$.
Then, for constants depending on $L$,
\begin{equation}
\label{eq:master_new}
W_2\!\bigl(\mu_1,\hat\mu_1^{(N)}\bigr)
\;\lesssim\;
\underbrace{\omega(c_d/K_c)^{1/2}}_{\text{coverage}}
\;+\;
\underbrace{\varepsilon_{\mathrm{fit}}+\varepsilon_{\mathrm{curv}}}_{\text{fit + straightness (Thm.~\ref{thm:app_gronwall})}}
\;+\;
\underbrace{\frac{1}{N}\int_0^1\|\partial_\tau v_\theta(u_\tau,\tau)\|\,d\tau}_{\text{discretization (Cor.~\ref{cor:global_euler})}}.
\end{equation}
\end{theorem}

\begin{proof}[Proof sketch]
The coverage term is \Cref{cor:coverage}. The continuous-time mismatch between the ideal rectified flow
and the learned flow is controlled by Theorem~\ref{thm:app_gronwall} via $\varepsilon_{\mathrm{fit}}$ and
$\varepsilon_{\mathrm{curv}}$. The discretization term follows from
Corollary~\ref{cor:global_euler} applied to the learned ODE.
\end{proof}

\noindent
\paragraph{Reading \eqref{eq:master_new} against the experiments.}
Equation \eqref{eq:master_new} matches the empirical breakdown seen in Section~\ref{sec:experiments}:
(i) \emph{coverage} corresponds to reproducing the high-wavenumber energy content (Figure~\ref{fig:spec_s1_E});
(ii) \emph{fit} corresponds to the overall statistical accuracy, reflected in $e_\mu$, $e_\sigma$, and $W_1$ (Table~\ref{tab:results});
(iii) \emph{straightness} and \emph{discretization} are exactly what turn ``a good model'' into ``a fast sampler'': ReFlow reaches the same law-level quality using $N\!=\!8$ steps, whereas diffusion baselines require $N\!\ge\!128$ (Section~\ref{sec:comp-efficiency}).

\subsection{Implementation Details for PDEs}

\noindent
\textbf{Learning velocity fields in PDE settings.}
When rectified flows are used to model PDE solutions, we assume access to paired initial/final states $(u_i,u_f)$ where $u_f = \mathcal{S}(u_i)$ represents the evolved initial condition $u_i$, at some final physical time (e.g.\ $t=1$). We train a neural network to predict the velocity $v_\theta$ that displaces $u_i$ towards $u_f$, with the additional mention that $v_\theta$ shall be conditioned on $u_i$ at all diffusion times.

\begin{algorithm}
\caption{Rectified Flow Training (PDE Context)}
\label{alg:rectified-train}
\begin{algorithmic}[1]
\Require Dataset $\{(u_0,u_1)\}$ with $u_1 = \mathcal{S}(u_0)$; velocity model $v_\theta$; noise schedule $\sigma(\tau)$
\For{each iteration}
    \State Sample a pair $(u_0,u_1)$
    \State Sample $\tau \sim \mathcal{U}(0,1)$ and noise $\xi \sim \mathcal{N}(0,I)$
    \State Form $u_\tau = \tau \cdot u_1 +  \sigma(1 - \tau)\,\xi$
    \State Compute loss:
    \[
       \mathcal{L}(\theta)
       \;=\;
       \bigl\|
         \bigl(u_1 - u_0\bigr)
         \;-\;
         v_\theta\bigl(u_\tau,\,u_0,\,\tau\bigr)
       \bigr\|^2
    \]
    \State Update $\theta$ via gradient descent
\EndFor
\end{algorithmic}
\end{algorithm}

\noindent
\textbf{Overview of the model.}
Our network learns a velocity (or denoising) map
\[
v_\theta(u_\tau, u_0, \,\tau)
\;=\;
\mathrm{UNet}_\theta
\bigl(\,u_\tau, u_0, \;\Gamma(\tau)\bigr),
\]
where $u_\tau$ is a low-frequency representation of $u_1$, $u_0$ is the initial datum, and \(\Gamma(\cdot)\) is an MLP-based time embedding:
\[
\Gamma(\tau)
\;=\;
\mathrm{MLP}\bigl(\mathrm{Sinusoid}(\tau)\bigr).
\]
Within the U-Net (see \textbf{SM} \ref{sec:unet-architecture}), we alternate residual blocks with multi‐head attention layers, downsampling to a bottleneck and then upsampling via skip connections.  A final \(1\times1\) projection recovers either the field increment $ \hat{u}  \approx u_1 - u_0$ or a noise residual.  By injecting the rectification time \(\Gamma(\tau)\) into each block, the model performs a fully deterministic trajectory integration instead of stochastic diffusion.

\noindent
\textbf{Sampling procedure.}
Once trained, we generate new samples by numerically integrating the ODE
\[
\frac{du_\tau}{d\tau}
=
v_\theta\bigl(u_\tau,\,u_0,\,\tau\bigr),
\quad
u_\tau \vert_{\tau = 0} = \xi,
\quad
\xi  \sim \mathcal{N}(0, I).
\]
Any standard ODE solver (e.g.\ forward Euler, Runge-Kutta) can be used; rectified flows often tolerate relatively large step sizes due to their straighter velocity field.

\noindent
\textbf{Efficiency and accuracy.}
Because rectified flows explicitly encourage a ``straighter'' velocity field, accurate solutions often require fewer integration steps than their diffusion-based counterparts.
This advantage can be crucial in computational fluid dynamics and other PDE applications where each solver step is expensive.
See \textbf{SM}~\ref{app:straightness_advantages} and~\ref{app:additional_error_plots}
 for a detailed comparison of step-count vs.\ accuracy trade-offs.

\subsection{Computational Efficiency Comparison}
\label{sec:comp-efficiency}

\begin{figure}[t]
  \centering
  \includegraphics[width=0.7\linewidth]{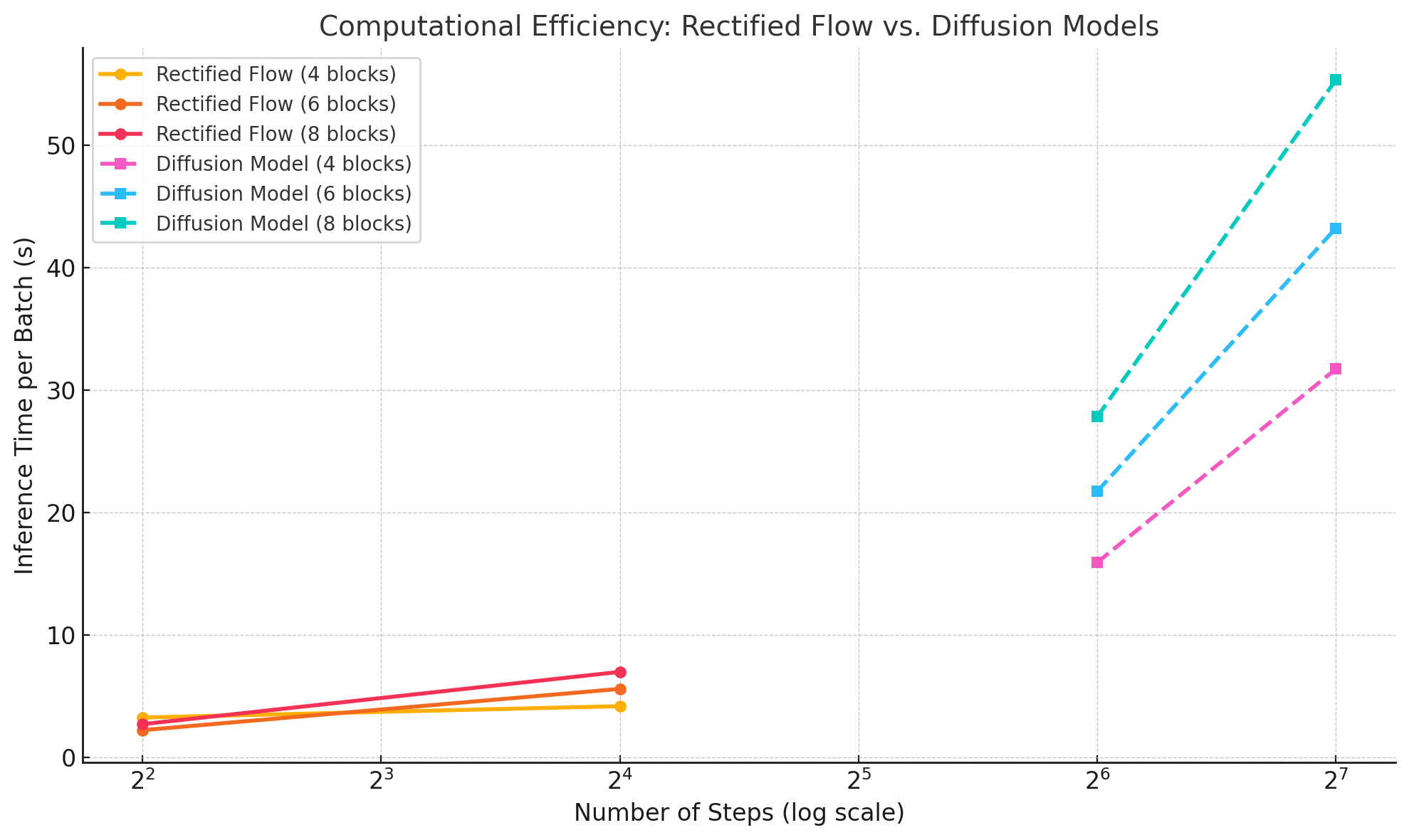}
  \caption{Inference cost versus model size.
    Rectified flows need $4$–$8$ ODE steps vs.\ GenCFD’s $64$–$128$ SDE steps.
    U\=/Net depth varies from $4$ to $8$ blocks (parameters $5.5$\,M to $10.2$\,M).}
  \label{fig:comp-efficiency}
\end{figure}

Diffusion surrogates (e.g.\ GenCFD) generate samples by integrating a
reverse-time SDE whose trajectory bends through latent noise; even with
accelerated samplers such as DDPM or DPM-solver the procedure still
requires tens of score-network evaluations.
Rectified flows, in contrast, follow a single deterministic ODE whose
nearly straight path tolerates large or adaptive steps, so inference
needs only a handful of evaluations.
Figure~\ref{fig:comp-efficiency} quantifies this gap, showing how the
runtime of diffusion models scales almost linearly with the number of
steps, whereas rectified flows remain fast even as network size grows.

\section{Experiments}
\label{sec:experiments}
\subsection{Experimental Setup}

We benchmark our method, termed henceforth as \emph{ReFlow}, against four neural surrogate models. \emph{GenCFD} and \emph{GenCFD
$\circ$ FNO} are conditional diffusion surrogates following the framework introduced earlier in this section, whereas FNO and UViT serve as one-shot deterministic baselines. Detailed descriptions of all baselines are provided in \textbf{SM}~\ref{app:models}.

\paragraph{Datasets for Training and Evaluation.}
We consider three canonical 2D multi-scale PDE problems (see the Supplementary Material for precise mathematical formulations and solver details):
\begin{itemize}
    \item[\textbf{SL2D}] \emph{Shear Layer in 2D}.
    This setup models the evolution of an incompressible shear layer, initially concentrated near a horizontal interface, which rolls up into intricate vortical structures over time.
    We generate approximate solutions by numerically solving the 2D Navier--Stokes system (with periodic boundaries) at a resolution of $512^2$, then downsample to $128^2$ for training and testing. The data has been generated with the spectral viscosity code \cite{SV,rohner}.
    \item[\textbf{CS2D}] \emph{Cloud-Shock Interaction in 2D}.
    In this popular compressible-flow test, a shock wave moves through a higher-density “cloud” region, triggering complex wave interactions and small-scale turbulent mixing in the wake.
    We solve the compressible Euler system at an effective resolution of $256^2$ using a GPU-optimized finite-volume ALSVINN code of \cite{KLye1,FLMW1}, then also downsample to $128^2$.
    \item[\textbf{RM2D}] \emph{Richtmeyer--Meshkov in 2D.} In this case, a shock wave interacts with a density interface separating two fluids, depositing vorticity and triggering vigorous mixing. We numerically solve the compressible Euler system on a $256^2$ grid, then downsample to $128^2$ to create the training and test data. The data has been generated with a high-resolution finite-volume ALSVINN code of \cite{FLMW1}
\end{itemize}

\paragraph{Training and Test Protocol.}
We train both our proposed rectified flow-based GenCFD-class algorithm and all baseline models on initial conditions
\[
u_0 \,\sim\, \mu,\quad \text{with}\quad \mu
\]
being the probability distribution specified in the Supplementary Material (\textbf{SM}.~\ref{app:data_processing}).
For each $u_0$, the PDE solver (e.g.\ a spectral or finite-volume code) is run forward in time to produce high-fidelity snapshots, which serve as training data.

\emph{Training-objective.} Crucially, during training, the model is \textbf{only} exposed to $(u_0, u_1)$ pairs of data as opposed to the conditional measure $p(u(t) \vert u_0)$.

\noindent
\emph{Test-time goal: Dirac initial measures.}
While the training distribution $\mu$ is spread out, we are especially interested in the scenario where the \emph{initial condition is fixed and deterministic}. Formally, we set
\[
\widehat{\mu}_0 = \delta_{\widehat{u}_0},
\]
which is a Dirac measure concentrated at a single function $\widehat{u}_0$. Thus, any stochasticity in the resulting flow arises solely from chaotic or multi-scale evolution over time, rather than from large variations in the initial condition itself.

\noindent
To create a “ground truth” ensemble for comparison under a fixed $\widehat{u}_0$, we employ an \emph{ensemble perturbation} procedure (detailed in \textbf{SM} ~\ref{app:data_processing}), where $\widehat{u}_0$ is slightly perturbed to generate multiple nearby initial states. Each perturbed state is then advanced using a high-resolution PDE solver (e.g.\ the Azeban code for incompressible flows, or a finite-volume code for compressible problems), yielding a set of flow realizations. This set approximates the conditional measure in question, i.e.\ $p\!\bigl(u(t)\mid \widehat{u}_0\bigr)$.

\noindent
\emph{Distribution settings.}
We test two main configurations:
\begin{enumerate}
\item \textbf{Matched distribution.}
We sample $\widehat{u}_0$ from the same family $\mu$ used in training.
While $\delta_{\widehat{u}_0}$ is technically out-of-distribution in the sense that the training distribution was spread out over many initial states, the base function $\widehat{u}_0$ still belongs to the same general class of flows.
\item \textbf{Distribution shift.}
We consider Dirac initial data $\delta_{\widehat{u}_0}$ where $\widehat{u}_0 \sim \nu$, for a \emph{different} distribution $\nu\neq\mu$.
This scenario probes the ability of our model and baselines to generalize under shifts in the initial condition distribution.
Specific details on $\nu$ are provided in \textbf{SM}~\ref{app:data_processing}.
\end{enumerate}
We will show results in both regimes to assess how well rectified flows can capture the spread of chaotic solutions starting from fixed initial states--and how it compares to baselines.

\subsection{Evaluation Metrics}
\label{sec:em}

We evaluate the generative performance of each method using three main metrics, with details provided in \textbf{SM}~\ref{app:results}:

\begin{itemize}

\item \textbf{Mean Error.}
We measure the $L^2$ difference between the model’s predicted mean flow $\mu$ and the ground-truth mean $\mu_{\mathrm{exact}}$.
A low mean error indicates accurate large-scale behavior.

\item \textbf{Standard Deviation Error.}
The normalized $L^2$ distance between the predicted standard deviation $\sigma$ and the reference $\sigma_{\mathrm{exact}}$ captures whether the model reproduces correct spatial variability.

\item \textbf{One-Point Wasserstein Distance.}
We compute an Wasserstein-$1$ distance at each spatial point (and average over the domain) to quantify local distributional agreement between ground truth and model outputs.

\end{itemize}

\noindent
\paragraph{One-point Wasserstein as a $k\!=\!1$ marginal metric.}
Let $\mu$ and $\widehat\mu$ be laws on fields $u:D\to\R^m$. For each $x\in D$ define
$\mu_x:=(\mathrm{ev}_x)_\#\mu$ and $\widehat\mu_x:=(\mathrm{ev}_x)_\#\widehat\mu$.
Our reported metric is
\[
W_1^{(1)}(\mu,\widehat\mu)
:= \int_D W_1(\mu_x,\widehat\mu_x)\,dx.
\]
By Kantorovich--Rubinstein duality,
$W_1(\mu_x,\widehat\mu_x)=\sup_{\Lip(\varphi)\le 1} \big|\int \varphi\,d\mu_x-\int\varphi\,d\widehat\mu_x\big|$,
so $W_1^{(1)}$ controls the averaged discrepancy of \emph{all} pointwise 1-Lipschitz observables.
This is the precise sense in which Table~\ref{tab:results} measures agreement of $k\!=\!1$ correlation statistics, in line with the statistical-solution viewpoint \cite{lanthaler2021statistical}.

\begin{table}[ht!]
\centering
\caption{Comparison of models across datasets CS2D, SL2D, and RM2D, in terms of mean relative error ($e_\mu$), standard deviation relative error ($e_\sigma$), and Wasserstein-1 distance ($W_1$). Best-performing model is colored in Blue and second-best performing model in Orange.}
\label{tab:results}
\resizebox{\textwidth}{!}{%
\begin{tabular}{llccccc|ccc|ccccc}
\toprule
\multirow{2}{*}{\textbf{Model}} & \multirow{2}{*}{\textbf{Metric}} & \multicolumn{5}{c|}{\textbf{CS2D}} & \multicolumn{3}{c|}{\textbf{SL2D}} & \multicolumn{5}{c}{\textbf{RM2D}} \\
&& $\rho$ & $m_x$ & $m_y$ & $E$ & & $u_x$ & $u_y$ & & $\rho$ & $m_x$ & $m_y$ & $E$ & \\
\midrule
\multirow{3}{*}{\makecell[l]{ReFlow}}
& $e_\mu$       & \bestcell{0.0477}  & \bestcell{0.0332} & \bestcell{0.041} & \bestcell{0.015} & & \bestcell{0.034}  & \bestcell{0.189}  & & \secondcell{0.018} & 0.032 & \secondcell{0.020} & \secondcell{0.021} & \\
& $e_\sigma$    & \bestcell{0.1005} & \bestcell{0.072} & \bestcell{0.078} & \bestcell{0.0584} & & \bestcell{0.071}  & \secondcell{0.077} & & \bestcell{0.032} & \secondcell{0.046} & \secondcell{0.046} & \secondcell{0.085} & \\
& $W_1$         & \bestcell{0.0091} & \bestcell{0.0107} & \bestcell{0.0124} & \bestcell{0.0143} & & \bestcell{0.0214} & \bestcell{0.0164} & & \bestcell{0.0033} & \secondcell{0.0036} & \bestcell{0.0028} & \bestcell{0.0017} & \\
\midrule
\multirow{3}{*}{GenCFD}
& $e_\mu$       & 0.0830 & 0.0619 & 0.127          & 0.0280 & & \secondcell{0.039}  & \secondcell{0.267}  & & \bestcell{0.0049} & \bestcell{0.0010} & \bestcell{0.0025} & \bestcell{0.0025} & \\
& $e_\sigma$    & 0.195          & 0.197          & 0.244          & 0.169          & & \secondcell{0.092}  & \bestcell{0.072}  & & \secondcell{0.0351}          & \bestcell{0.0254} & \bestcell{0.0435} & \bestcell{0.0435} & \\
& $W_1$         & \secondcell{0.0151}         & \secondcell{0.0184}         & \secondcell{0.0185}         & \secondcell{0.0247}         & & \secondcell{0.0276}          & \secondcell{0.0213}          & & \secondcell{0.0092} & \bestcell{0.0016} & \secondcell{0.0049} & \secondcell{0.0049} & \\
\midrule
\multirow{3}{*}{UViT}
& $e_\mu$       & 0.111          & 0.078         & 0.127          & 0.0479          & & 0.233           & 0.623           & & 0.070          & 0.070          & 0.071          & 0.052          & \\
& $e_\sigma$    & 0.415          & 0.343       & 0.412        & 0.296          & & 0.28           & 0.178           & & 0.425          & 0.570          & 0.570          & 0.617          & \\
& $W_1$         & 0.087          & 0.085          & 0.158          & 0.172          & & 0.270           & 0.256           & & 0.060          & 0.055          & 0.060          & 0.036          & \\
\midrule
\multirow{3}{*}{FNO}
& $e_\mu$       & 0.0989          & 0.0748          & 0.0930          & 0.0383          & & 0.058           & 0.408           & & 0.085          & 0.070          & 0.063          & 0.067          & \\
& $e_\sigma$    & 0.2573          & 0.2779          & 0.3188          & 0.2777          & & 0.137           & 0.110           & & 0.366          & 0.443          & 0.436          & 0.450          & \\
& $W_1$         & 0.1455          & 0.1660          & 0.1879          & 0.1884          & & 0.184           & 0.150           & & 0.096          & 0.051          & 0.056          & 0.062          & \\
\midrule
\multirow{3}{*}{GenCFD \(\boldsymbol{\circ}\) FNO}
& $e_\mu$       & \secondcell{0.0638}          & \secondcell{0.0454}          & \secondcell{0.0603}          & \secondcell{0.0204}          & & 0.055           & 0.393           & & 0.040          & \secondcell{0.027}          & 0.039          & 0.041          & \\
& $e_\sigma$    & \secondcell{0.1099}          & \secondcell{0.0969}          & \secondcell{0.1245}          & \secondcell{0.0974}          & & 0.131           & 0.100           & & 0.091          & 0.075          & 0.091          & 0.090          & \\
& $W_1$         & 0.0467          & 0.0534          & 0.0774          & 0.0743          & & 0.166           & 0.135           & & 0.0269          & 0.0158          & 0.0293          & 0.0273          & \\
\bottomrule
\end{tabular}%
}
\end{table}

\textbf{Baselines and Evaluation Settings.}  To stay within reasonable training times for our infrastructure, we restrict the total number of training iterations up to 120{,}000 for each model and recorded the best achievable results. Under this budget, only a reduced-capacity GenCFD (with two downsampling stages in its U-Net backbone, versus three in Rectified Flow) can match the accuracy reported in Table \ref{tab:results}. We also evaluated larger GenCFD variants trained for 500{,}000 iterations, but observed no meaningful improvement over the results shown here. Meanwhile, at inference time Rectified Flow requires only \textbf{8} denoising steps to achieve comparable or superior accuracy, whereas diffusion baselines (including GenCFD) need \textbf{128} steps, underscoring RF’s efficiency in both training and sampling. FNO and UViT also require smaller models than ReFlow for optimal performance.

Since FNO, and UViT produce \emph{deterministic} predictions, we approximate their spread by an \emph{ensemble perturbation} strategy (\textbf{SM }~\ref{app:data_processing}), where each baseline is fed slightly perturbed inputs and outputs are aggregated to form a sample set.
In contrast, both the original GenCFD (baseline) and our \emph{ReFlow} are inherently \emph{probabilistic}, requiring no auxiliary procedure to generate diverse samples.

\begin{figure}[h!]
 \centering
 \includegraphics[width=0.95\linewidth]{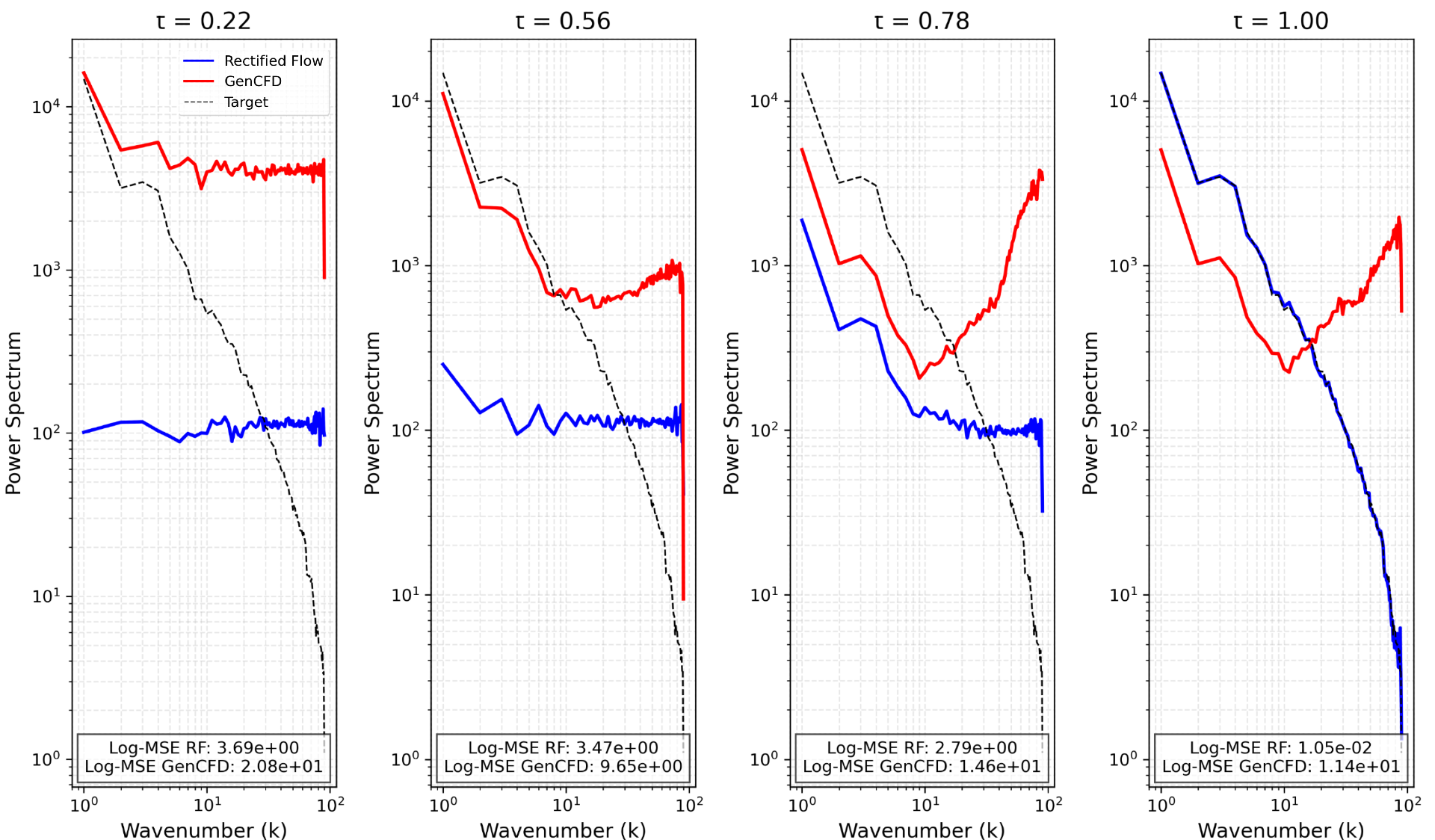}
 \caption{Energy-spectrum evolution for a random cloud–shock initial
           datum.  ReFlow (blue) tracks high-wavenumber energy faster
           and more accurately than GenCFD (red).}
 \label{fig:spec_s1_E}
\end{figure}

\subsection{Results.}
Table~\ref{tab:results} compares five surrogate models (ReFlow, together with the four baselines) on the three PDEs learning tasks using the mean error ($e_{\mu}$), spread error or \emph{standard deviation} ($e_{\sigma}$)
and the distribution-level Wasserstein-1 distance ($W_{1}$). On the cloud–shock and shear-layer datasets (CS2D, SL2D) the
rectified-flow model (ReFlow) attains the lowest values on all three
metrics, indicating that its samples most closely match both the mean
behaviour and the higher-order statistics of the ground truth. The hybrid \emph{GenCFD$\circ$FNO} reduces the gap to ReFlow relative to
its GenCFD component, but still records noticeably larger errors,
whereas the one-shot deterministic baselines (FNO, UViT) show the
largest deviations, especially in $e_{\sigma}$ and $W_{1}$, confirming
their difficulty in capturing output variability. RM2D is the only benchmark whose test set remains strictly in-distribution (in the sense defined earlier). Under this easier regime the deterministic baselines come closest to the conditional-diffusion surrogates, but only here. Their drop-off on the out-of-distribution tasks underscores the stronger generalisation of conditional diffusion models. Under identical training budgets, ReFlow achieves lower errors on CS2D and SL2D compared to all baselines, including GenCFD with the same architecture and model size. This highlights ReFlow's superior convergence efficiency despite parity in model capacity and training iterations.

Moreover as visualized in the {\bf SM}, ReFlow generates samples of very high quality, matching the qualitative features of the ground truth and capturing features such as (interacting) vortices as well as propagating shock waves very accurately. Another aspect where ReFlow really shines in being able to generate the correct point pdfs. Although already indicated in the very low 1-point Wasserstein distance errors from Table \ref{tab:results}, we present 1-point pdfs in the {\bf SM} to illustrate this observation.

\begin{figure}
  \centering
  \includegraphics[width=0.85\linewidth]{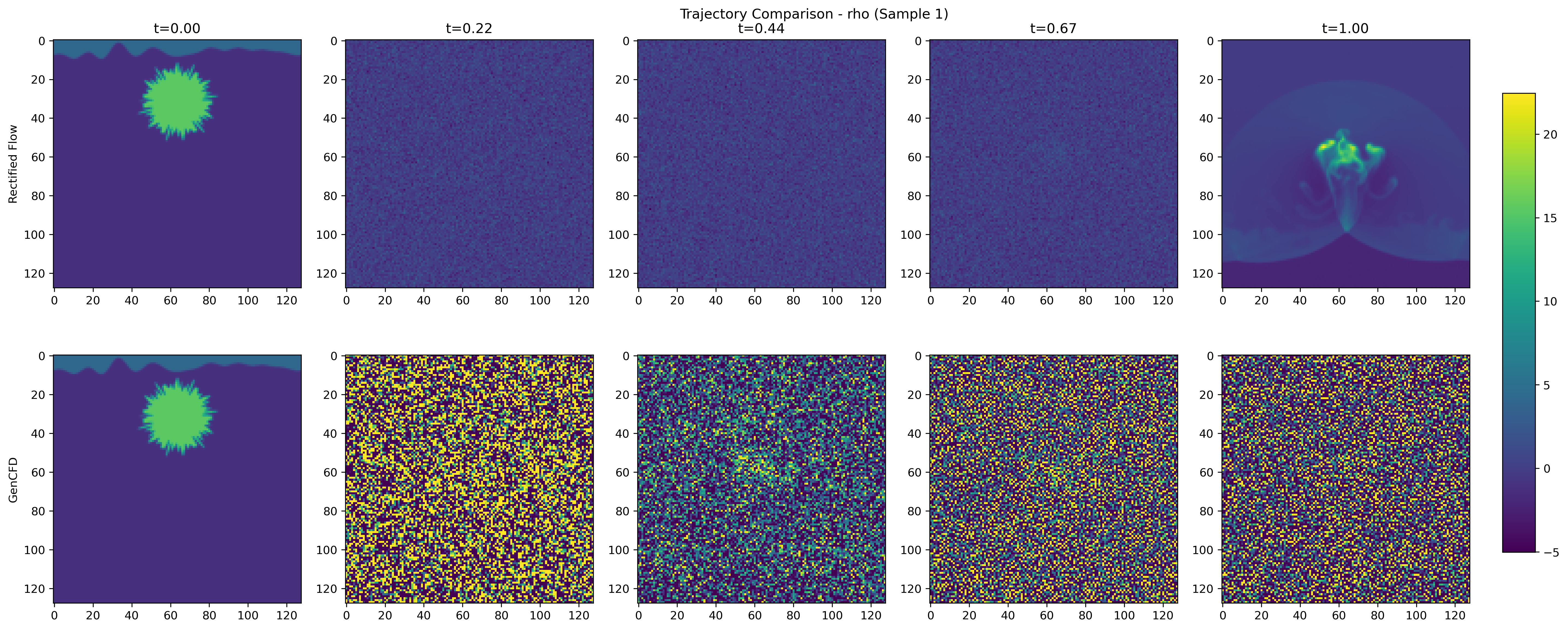}
  \caption{Density trajectories for Sample 1: ReFlow (top) versus GenCFD
           (bottom).}
  \label{fig:traj_rho_s1_main}
\end{figure}

Another key indicator of the quality of approximation of multiscale PDE solutions is the energy spectrum \cite{molinaro2024generative}. In Figure \ref{fig:spec_s1_E}, we plot the (log) spectra for the energy variable of a randomly chosen Cloud Shock initial datum at four different Diffusion times, $\tau\in\{0.25,\,0.50,\,0.75,\,1.00\}$ to observe that i) the solution is indeed highly multiscale, with spectra decaying as a power law over a range of frequencies ii) ReFloW approximates the spectra accurately within a few (5) steps and iii) on the other hand, for these few steps in solving the underlying reverse SDE, a score-based Diffusion model such as GenCFD, struggles to approximate the relevant small scales. It will require much more sampling steps to do so. These results also showcase how ReFlow is able to infer the correct statistical solutions with significantly less ODE solve steps and is hence, much faster, than a diffusion based model such as GenCFD. This gain in inference speed is also reinforced from Figure \ref{fig:traj_rho_s1_main},  where we compare ReFlow (RF) and GenCFD reconstructions of the density field~$\rho$ for the Cloud-Shock interaction over five diffusion times $\tau \in\{0.00,\,0.25,\,0.50,\,0.75,\,1.00\}$. We see that ReFlow inpaints fine-scale features in far fewer steps, whereas GenCFD remains visibly much more noisy at these small number of steps. See \textbf{SM} \ref{app:straightness_advantages} for further visualizations of this effect.

\section{Additional empirical analysis (integrators, robustness, and diversity)}
\label{sec:exp:rebuttal}

This section consolidates additional experiments conducted in response to reviewer requests,
with the goal of making the computational trade-offs and the role of the curvature-aware controller
explicit. The emphasis is on (i) solver choices for the learned rectified-flow ODE,
(ii) robustness of the controller to its hyperparameters, and
(iii) distributional fidelity / diversity under the macro--micro evaluation protocol.

\subsection{Protocol recap: macro--micro ensembles, OOD macros, and metrics}
\label{subsec:exp:protocol-recap}

\paragraph{Macro--micro conditional ensembles.}
To evaluate \emph{statistical fidelity} rather than only pointwise accuracy, we follow a macro--micro protocol.
For each dataset, we select $N_{\mathrm{macro}}$ macroscopic states (``macros'') and for each macro generate
$N_{\mathrm{micro}}$ microscopic perturbations (``micros'') yielding a reference ensemble of size
$N_{\mathrm{macro}}\times N_{\mathrm{micro}}$.
This makes diversity concrete: for each fixed macro, the ground-truth solver produces an empirical conditional
distribution over micros, and the model produces its own empirical conditional distribution.
We compare these distributions directly via Wasserstein distances (below), which is the relevant question for
statistical surrogates.

\paragraph{OOD macro selection.}
Unless otherwise stated, we report results in an out-of-distribution (OOD) regime by choosing
$N_{\mathrm{macro}}=10$ macros from shifted initial-condition families / parameter ranges, and
$N_{\mathrm{micro}}=20$ micros per macro (200 conditional samples per dataset).
All models are frozen; only the sampling/integration procedure is varied.

\paragraph{Metrics.}
We report:
(i) \emph{relative $L^2$ error} against the PDE solution,
\begin{equation}
\label{eq:relL2}
\mathrm{Rel}\text{-}L^2(\widehat u, u)
:=
\frac{\|\widehat u-u\|_{L^2}}{\|u\|_{L^2}},
\end{equation}
(ii) \emph{NFE} (neural function evaluations), i.e.\ the number of velocity-field evaluations per generated sample,
and (iii) a combined cost--accuracy scalar
\begin{equation}
\label{eq:costerr}
\mathrm{Cost}\times \mathrm{Err}
:=
(\mathrm{Rel}\text{-}L^2)\times (\mathrm{NFE}),
\end{equation}
which is convenient for summarizing the NFE--error plane (lower is better).
Distributional fidelity is measured via empirical $\Wone$ between conditional ensembles;
we report both aggregated and per-variable $\Wone$ (Section~\ref{subsec:exp:w1}).

\subsection{Integrator study at sampling time: fixed-step baselines vs.\ curvature-aware control}
\label{subsec:exp:integrators}

A recurring concern is whether the gains are merely a consequence of choosing a particular numerical solver.
To isolate this, we fix the trained rectified-flow model and vary \emph{only} the integrator used to solve the
rectified-flow ODE at sampling time, starting from noise and integrating from rectified time $\tau=0$ to $\tau=1$.
We include a reference implementation from \texttt{torchdiffeq} and standard explicit fixed-step schemes.
We report the aggregated results on \emph{ShearLayer2D} in the OOD macro--micro regime; the second benchmark
exhibits the same qualitative trends and is reported in the appendix.

\begin{table}[t]
\centering
\caption{\textbf{Fixed-step ODE integrators} at sampling time (ShearLayer2D, OOD macro--micro; $10$ macros $\times$ $20$ micros).
We fix the trained rectified-flow model and vary only the integrator.
NFE is the number of velocity evaluations; lower is faster.
Cost$\times$Err is defined in \eqref{eq:costerr}.}
\label{tab:fixed_integrators}
\vspace{0.25em}
\begin{tabular}{lccc}
\toprule
Method & Avg.\ NFE $\downarrow$ & Rel.\ $L^2$ Error $\downarrow$ (mean $\pm$ std) & Cost$\times$Err $\downarrow$ \\
\midrule
Model-Baseline (built-in sampler) & 32.0 & 0.2533 $\pm$ 0.0382 & 8.11 \\
\texttt{torchdiffeq} Midpoint & 32.0 & 0.2533 $\pm$ 0.0382 & 8.11 \\
Euler (16 steps) & 16.0 & 0.2489 $\pm$ 0.0371 & 3.98 \\
RK2 (16 steps; 32 NFEs) & 32.0 & 0.2533 $\pm$ 0.0379 & 8.11 \\
RK4 (16 steps; 64 NFEs) & 64.0 & 0.2540 $\pm$ 0.0381 & 16.26 \\
\bottomrule
\end{tabular}
\end{table}

\paragraph{Observation: Euler has a tuned ``sweet spot'' but remains brittle.}
Among fixed-step schemes, Euler with 16 steps is the strongest baseline in this regime, improving on 32/64-NFE
schemes in Cost$\times$Err. However, in multiscale mildly stiff regimes, fixed-step performance is highly sensitive
to the chosen step budget: slightly fewer steps can degrade accuracy sharply, while increasing steps can \emph{also}
worsen error due to accumulated discretization bias along curved trajectories. Thus, even simple Euler is not a
drop-in robust choice: it requires per-dataset/per-regime tuning to hit its efficient operating point.

\subsection{Curvature-aware sampler: hyperparameter robustness and Pareto dominance}
\label{subsec:exp:robustness}

We next evaluate the curvature-based adaptive controller under a broad sweep of hyperparameters while holding
the model fixed. The controller uses (i) an EMA-based straightness/curvature proxy on the learned velocity field,
(ii) a mild velocity regularization/blending step triggered by curvature, and (iii) adaptive step sizing.
We vary EMA time scale, thresholds, gating, damping, and calibration parameters across 20 configurations.

\begin{table}[t]
\centering
\caption{\textbf{Hyperparameter robustness} of the curvature-aware adaptive sampler (ShearLayer2D, OOD macro--micro).
Across 20 controller configurations, Avg.\ NFE stays in a narrow band and relative errors vary by $<1\%$,
indicating the controller is not brittle and does not rely on a narrow tuning ``sweet spot''.}
\label{tab:adaptive_sweep}
\vspace{0.25em}
\begin{tabular}{lccc}
\toprule
Config & Avg.\ NFE $\downarrow$ & Rel.\ $L^2$ Error $\downarrow$ (mean $\pm$ std) & Cost$\times$Err $\downarrow$ \\
\midrule
baseline & 11.0 & 0.2374 $\pm$ 0.0389 & 2.62 \\
\textbf{ema\_0.25} & \textbf{10.7} & \textbf{0.2369 $\pm$ 0.0386} & \textbf{2.54} \\
ema\_0.45 & 11.3 & 0.2361 $\pm$ 0.0384 & 2.67 \\
alpha\_0.05 & 11.0 & 0.2373 $\pm$ 0.0389 & 2.62 \\
alpha\_0.12 & 11.0 & 0.2373 $\pm$ 0.0389 & 2.62 \\
alpha\_0.20 & 11.0 & 0.2374 $\pm$ 0.0389 & 2.62 \\
gamma\_1.5 & 11.0 & 0.2374 $\pm$ 0.0389 & 2.62 \\
gamma\_2.0 & 11.0 & 0.2374 $\pm$ 0.0389 & 2.62 \\
gamma\_2.5 & 11.0 & 0.2374 $\pm$ 0.0389 & 2.62 \\
gate\_0.50 & 11.0 & 0.2374 $\pm$ 0.0389 & 2.62 \\
gate\_0.70 & 11.0 & 0.2374 $\pm$ 0.0389 & 2.62 \\
calib\_0.70\_0.90 & 11.1 & 0.2371 $\pm$ 0.0387 & 2.63 \\
calib\_0.80\_0.98 & 10.9 & 0.2372 $\pm$ 0.0387 & 2.60 \\
adapt\_1.5\_0.75 & 12.8 & 0.2374 $\pm$ 0.0378 & 3.04 \\
adapt\_2.0\_0.85 & 11.2 & 0.2360 $\pm$ 0.0376 & 2.63 \\
no\_ortho\_filter & 11.0 & 0.2374 $\pm$ 0.0389 & 2.62 \\
damp\_0.05 & 11.0 & 0.2373 $\pm$ 0.0389 & 2.62 \\
damp\_0.10 & 11.0 & 0.2373 $\pm$ 0.0389 & 2.61 \\
damp\_0.20 & 11.0 & 0.2370 $\pm$ 0.0389 & 2.60 \\
damp\_0.30 & 11.0 & 0.2368 $\pm$ 0.0389 & 2.60 \\
\bottomrule
\end{tabular}
\end{table}

\paragraph{Observation: robust controller, automatic step allocation.}
Across all 20 configurations, Avg.\ NFE remains in a tight range (10.7--12.8) and the mean relative error stays within
$[0.2360,0.2374]$ (sub-$1\%$ relative variation), indicating that the controller does not require fragile tuning.
This directly contrasts with fixed-step Euler, where efficiency is achieved only at a tuned step budget and can degrade
rapidly outside it.

\subsection{Best fixed-step baseline vs.\ best adaptive configurations}
\label{subsec:exp:bestvsbest}

For clarity, Table~\ref{tab:best_vs_best} compares the strongest fixed-step baseline (Euler-16) against the top adaptive
configurations on ShearLayer2D OOD.

\begin{table}[t]
\centering
\caption{\textbf{Best fixed-step baseline vs.\ top adaptive configs} (ShearLayer2D, OOD macro--micro).
At roughly one third fewer velocity evaluations, the adaptive sampler improves both error and Cost$\times$Err.}
\label{tab:best_vs_best}
\vspace{0.25em}
\begin{tabular}{lcccccc}
\toprule
Method / Config & Avg.\ NFE & Rel.\ $L^2$ Err & Cost$\times$Err & $\Delta$NFE vs Euler & $\Delta$Err vs Euler & $\Delta$(Cost$\times$Err) \\
\midrule
\textbf{Euler (16 steps)} & 16.0 & 0.2489 & 3.98 & 0\% & 0\% & 0\% \\
\textbf{ema\_0.25} & 10.7 & 0.2369 & 2.54 & $-33.1\%$ & $-4.8\%$ & $-36.2\%$ \\
calib\_0.80\_0.98 & 10.9 & 0.2372 & 2.60 & $-31.9\%$ & $-4.7\%$ & $-34.7\%$ \\
damp\_0.30 & 11.0 & 0.2368 & 2.60 & $-31.3\%$ & $-4.9\%$ & $-34.7\%$ \\
\bottomrule
\end{tabular}
\end{table}

\textbf{Interpretation in the NFE--error plane.}
These comparisons summarize a consistent pattern: the adaptive curvature-aware sampler lies on (or very near) the
Pareto frontier in the NFE--error plane, while fixed-step baselines are strictly dominated in the regimes tested.
This is the practical meaning of the curvature theory: the controller allocates steps only where the learned transport
bends, while retaining large steps when the flow is straight.

\subsection{Distributional fidelity and diversity: ensemble $\Wone$ under macro--micro evaluation}
\label{subsec:exp:w1}

A common concern for deterministic probability-flow ODE sampling is potential under-dispersion or reduced diversity.
Our evaluation is designed to test this directly in the conditional setting where diversity is well-defined.

\paragraph{Empirical $\Wone$ between conditional ensembles.}
For each macro, let $\{\widehat u^{(j)}\}_{j=1}^{N_{\mathrm{micro}}}$ be model samples and
$\{u^{(j)}\}_{j=1}^{N_{\mathrm{micro}}}$ be reference samples.
We compute a one-point (spatially averaged) empirical $\Wone$ between the corresponding micro-ensembles,
and then average across macros. We also report per-variable $\Wone$ (density, momenta, energy) and an aggregated score.

\paragraph{Key empirical outcome.}
Across benchmarks, our rectified-flow model matches or improves upon the diffusion baseline in ensemble $\Wone$
while using substantially fewer function evaluations. In particular, the macro--micro protocol provides a direct
quantitative check against mode collapse: matching $\Wone$ at fixed macro conditions indicates that the conditional
output distribution is captured rather than collapsed.
(Complete per-variable $\Wone$ tables and additional datasets are reported in the appendix.)

\subsection{Practical takeaway: speed requires both straight transports and curvature-aware integration}
\label{subsec:exp:takeaway}

The integrator study separates two effects:
(i) the rectified-flow training objective yields a transport that can be traversed with far fewer evaluations than
score-based diffusion, and
(ii) within this ODE setting, curvature-aware control is a robust way to realize the efficiency in stiff multiscale regimes
without per-problem solver tuning.
Empirically, this is exactly what Tables~\ref{tab:fixed_integrators}--\ref{tab:best_vs_best} show: the controller
achieves lower error at lower NFE than the strongest tuned fixed-step baseline, and its performance is stable across
wide hyperparameter ranges.

\vspace{0.25em}
\noindent\textbf{Appendix pointers.}
We provide: (a) the same integrator/robustness tables on the second benchmark,
(b) NFE--error scatter plots (Pareto frontiers), and (c) detailed per-variable $\Wone$ breakdowns and additional OOD regimes.

\section{Conclusion and Future Work}
\label{sec:conclusion}

We have demonstrated that rectified flows provide an efficient surrogate for sampling multi-scale PDE solutions: on a suite of challenging incompressible and compressible problems, our method matches the statistical fidelity of conditional diffusion models while requiring an order of magnitude fewer solver steps—yielding up to 22× faster inference in practice. Across matched and shifted initial‐condition regimes, rectified flows consistently reproduce energy spectra, vorticity distributions, and one‐point Wasserstein distances within sampling error, confirming their robustness and multi‐scale accuracy.

\paragraph{Key Takeaways.} On account of its straightening bias, ReFlow requires significantly fewer ODE steps at inference-time (8-10) than its conditional-diffusion-based counterparts (more than 128) to achieve the same level of performance on our statistical metrics -- consult \textbf{SM} \ref{app:results} for more details. Furthermore, ReFlow is much more training efficient, requiring up to 120,000 iterations in reaching the same level of performance, across all experiments, as our other conditional diffusion baselines (which typically take 500-600,000 iterations), see \textbf{SM} \ref{app:training_efficiency}.  Our results further demonstrate that straightening effects remain evident even when projecting down to a 2D PCA subspace (see \textbf{SM} \ref{app:straightness_advantages}).

\bibliographystyle{tmlr}
\bibliography{main}

\appendix
\clearpage

\section{Theory appendix: law-level rectified flows for multiscale CFD}
\label{app:theory}

\paragraph{Purpose and scope.}
This appendix provides the mathematical backbone for the claims made in the main text:
(i) rectified flows can match the same conditional solution laws as diffusion-class models,
(ii) their \emph{sampling step count} is governed by a \emph{curvature/straightness} quantity
rather than by the raw dimensionality of the state, and
(iii) at fixed spatial resolution, there is an \emph{irreducible} law-level error dictated by
the small-scale tail of the true PDE law, which is naturally quantified through structure functions.
We keep the presentation self-contained, and we explicitly connect each theoretical ingredient to
the empirical evaluations in Table~\ref{tab:results} and to the spectral diagnostics in
Figure~\ref{fig:spec_s1_E} (and supplementary figures).

\paragraph{How to read this appendix.}
Section~\ref{app:theory:setup} fixes the measure-theoretic setting and links it to statistical
solutions in the sense of \cite{lanthaler2021statistical} (also used in the Introduction).
Section~\ref{app:theory:rf-correctness} shows that the \emph{barycentric rectified velocity} exactly
transports $\mu_0$ to $\mu_1$ (correctness at the level of laws).
Section~\ref{app:theory:err-decomp} gives a \emph{terminal error decomposition} into a fit term and a
straightness term and explains how the training objective controls these terms.
Section~\ref{app:theory:discretization} derives why curvature controls the required number of ODE
steps and motivates the EMA-based curvature proxy and the blending correction used at inference.
Section~\ref{app:theory:coverage} links the \emph{energy spectrum / structure functions} to an
unavoidable \emph{coverage} term at fixed cutoff $K_c$.
Finally, Section~\ref{app:theory:master} aggregates the pieces into a master law-level bound that
matches the narrative in the abstract and in the theory paragraphs of the main paper.

\section{Setup: laws of PDE solutions and what the experiments measure}
\label{app:theory:setup}

\paragraph{State space.}
We work on the periodic domain $\D=\T^d$ and consider (for notational simplicity) scalar or vector
fields $u:\D\to\R^m$ represented on a finite grid. The natural infinite-dimensional space for the
fluid benchmarks is
\[
\Hcal := L^2_\sigma(\D;\R^d)
\quad\text{(divergence-free in the incompressible case),}
\]
or $\Hcal:=L^2(\D;\R^m)$ for compressible state vectors (e.g.\ $(\rho,m_x,m_y,E)$).
All constructions below are rigorous on the finite-dimensional Galerkin/grid surrogate
$\R^{d_N}\subset\Hcal$ used in training and evaluation. We state results in $\Hcal$ language to
highlight the measure-theoretic meaning.

\paragraph{Statistical solutions and correlation viewpoint.}
In strongly nonlinear regimes, it is natural to treat the PDE solution as a \emph{random field}.
Given a law $\mu_0\in\Pcal(\Hcal)$ for the initial condition, the law at time $t$ is
\[
\mu_t = (\Scal^t)_\# \mu_0,
\]
where $\Scal^t$ is the PDE solution operator (possibly multivalued in the inviscid setting; see
\cite{lanthaler2021statistical} for the correlation hierarchy framework).
This is precisely the “statistical solution” language used in the Introduction.

\paragraph{What we evaluate empirically.}
Our quantitative evaluation (Section~\ref{sec:em}) uses:
(i) errors of the mean and standard deviation fields,
and (ii) a pointwise Wasserstein-$1$ discrepancy averaged over space.
These are \emph{marginal} discrepancies: they test whether low-order statistics and one-point laws
match between the model and the PDE ensemble.

From a law-level perspective, a natural global metric is $\Wtwo$ on $\Pcal(\Hcal)$.
Although our reported metric is $\Wone$ on one-point marginals (for computational reasons), the
theory below uses $\Wtwo$ because (a) it interacts cleanly with $L^2$-based structure functions and
Fourier tails, and (b) it yields transparent decompositions. We emphasize that the core mechanisms
(coverage/fit/straightness/discretization) are \emph{metric-agnostic}: they explain why a method that
reduces curvature reaches good statistics with fewer steps, which is exactly what is observed in
Table~\ref{tab:results} and Figure~\ref{fig:spec_s1_E}.

\section{Rectified-flow correctness: barycentric velocity transports laws}
\label{app:theory:rf-correctness}

\subsection{Couplings and linear interpolation in rectification time}

Let $\mu_0,\mu_1\in\Pcal(\Hcal)$ be two laws (in our application: conditional laws of PDE outputs
given inputs; see Section~\ref{sec:math-form}).
Fix a coupling $\gamma\in\Pcal(\Hcal\times\Hcal)$ of $(\mu_0,\mu_1)$, i.e.
$(\pi_0)_\#\gamma=\mu_0$ and $(\pi_1)_\#\gamma=\mu_1$.

For $\tau\in[0,1]$ define the interpolation map
\[
T_\tau(u_0,u_1) := (1-\tau)u_0+\tau u_1,
\qquad
\rho_\tau := (T_\tau)_\#\gamma.
\]
Thus $\rho_\tau$ is the law of the random variable $U_\tau=(1-\tau)U_0+\tau U_1$ when
$(U_0,U_1)\sim\gamma$.

\subsection{Barycentric rectified velocity and the continuity equation}

\begin{definition}[Barycentric rectified velocity]
\label{def:app:barycentric}
For each $\tau\in[0,1]$ and $\rho_\tau$-a.e.\ $u$, define
\[
v_\star(u,\tau)
:= \EE\!\bigl[\,U_1-U_0 \,\big|\, U_\tau=u\,\bigr].
\]
Equivalently, $v_\star$ is obtained by disintegrating $\gamma$ along the fibers
$T_\tau^{-1}(u)$, as in equation~\ref{eq:barycentric} in the main text.
\end{definition}

\paragraph{Why this is the “right” velocity.}
If we think of $\rho_\tau$ as mass moving from $\rho_0=\mu_0$ to $\rho_1=\mu_1$ in rectification time,
then the pair $(\rho_\tau,v_\star)$ solves the continuity equation
\[
\partial_\tau \rho_\tau + \nabla\!\cdot(\rho_\tau v_\star(\cdot,\tau)) = 0
\quad\text{(on the finite-dimensional surrogate).}
\]
This is the deterministic analogue of the probability-flow (ODE) viewpoint of diffusion models:
the flow map of $v_\star$ pushes $\rho_0$ to $\rho_1$.

\begin{proposition}[Exact transport under the barycentric velocity]
\label{prop:app:exact-transport}
Assume we work on a finite-dimensional surrogate of $\Hcal$ so that $u\in\R^{d_N}$ and
$\rho_\tau$ admits a density (or, more generally, is a narrowly continuous curve of measures).
Let $X_\tau$ be the (classical) flow map solving
\[
\frac{d}{d\tau} X_\tau(u) = v_\star(X_\tau(u),\tau),
\qquad
X_0(u)=u.
\]
Then $(X_\tau)_\#\mu_0=\rho_\tau$ for all $\tau\in[0,1]$, and in particular
$(X_1)_\#\mu_0=\mu_1$.
\end{proposition}

\begin{proof}
Let $(U_0,U_1)\sim\gamma$ and define $U_\tau=(1-\tau)U_0+\tau U_1$ so that $\Law(U_\tau)=\rho_\tau$.
Fix $\varphi\in C_c^\infty(\R^{d_N})$.
By differentiating under the expectation,
\[
\frac{d}{d\tau}\EE[\varphi(U_\tau)]
= \EE\bigl[\nabla\varphi(U_\tau)\cdot (U_1-U_0)\bigr].
\]
Condition on $U_\tau$ and use Definition~\ref{def:app:barycentric}:
\[
\EE\bigl[\nabla\varphi(U_\tau)\cdot (U_1-U_0)\bigr]
= \EE\bigl[\nabla\varphi(U_\tau)\cdot \EE[U_1-U_0\mid U_\tau]\bigr]
= \EE\bigl[\nabla\varphi(U_\tau)\cdot v_\star(U_\tau,\tau)\bigr].
\]
Thus $\rho_\tau$ satisfies the weak continuity equation
\[
\frac{d}{d\tau}\int \varphi(u)\,\rho_\tau(du)
=
\int \nabla\varphi(u)\cdot v_\star(u,\tau)\,\rho_\tau(du).
\]
On the other hand, the pushforward curve $\tilde\rho_\tau:=(X_\tau)_\#\rho_0$ satisfies the same
weak equation by standard characteristics. Uniqueness of solutions to the continuity equation on
$\R^{d_N}$ (under the regularity ensuring existence of $X_\tau$) yields $\rho_\tau=\tilde\rho_\tau$,
hence $(X_1)_\#\mu_0=\mu_1$.
\end{proof}

\paragraph{Empirical tie.}
Correctness in Proposition~\ref{prop:app:exact-transport} is a law-level statement: it says that if a
learned velocity $v_\theta$ matches $v_\star$, then integrating the ODE recovers the target law.
This is what is tested empirically through mean/spread/Wasserstein statistics (Table~\ref{tab:results})
and through the evolution of multiscale spectra along $\tau$ (Figure~\ref{fig:spec_s1_E}).

\section{Continuous-time error decomposition: fit vs.\ straightness}
\label{app:theory:err-decomp}

\subsection{The training objective targets the barycentric velocity}

In the (unconditional) rectified-flow construction, minimizing
\[
\int_0^1 \EE\bigl[\| (U_1-U_0) - v(U_\tau,\tau)\|^2 \bigr]\;d\tau
\]
over measurable $v$ yields the conditional expectation
$v_\star(u,\tau)=\EE[U_1-U_0\mid U_\tau=u]$.
In the conditional PDE setting we use the same principle but with $(U_0,U_1)$ sampled from the
conditional pairing induced by the dataset and conditioning variables (initial/boundary data).
This explains why the regression loss in Algorithm~\ref{alg:rectified-train} is the natural analogue
of score matching: it is an $L^2$ projection onto the optimal transport direction at each $\tau$.

\subsection{Terminal error bound with explicit decomposition}

Let $u_\tau$ solve the ideal rectified ODE $\dot u_\tau=v_\star(u_\tau,\tau)$ and let $\hat u_\tau$
solve the learned ODE $\dot{\hat u}_\tau=v_\theta(\hat u_\tau,\tau)$ with the same initial data
$u_0=\hat u_0$.

Define the time-average velocity
\[
\bar v_\theta(u) := \int_0^1 v_\theta(u,\tau)\,d\tau.
\]
The average isolates the “straight-line” component of the learned field, while deviations from it
quantify curvature in rectification time.

\begin{theorem}[Terminal error decomposition (fit + straightness)]
\label{thm:app:terminal}
Assume $v_\theta(\cdot,\tau)$ is $L$-Lipschitz in $u$ uniformly in $\tau$ and that $u_\tau$ remains in
a region where the quantities below are finite. Define
\[
\varepsilon_{\mathrm{fit}}^2
:=\int_0^1 \EE\bigl\|\bar v_\theta(u_\tau)-v_\star(u_\tau,\tau)\bigr\|^2\,d\tau,
\qquad
\varepsilon_{\mathrm{curv}}^2
:=\int_0^1 \EE\bigl\|v_\theta(u_\tau,\tau)-\bar v_\theta(u_\tau)\bigr\|^2\,d\tau.
\]
Then
\begin{equation}
\label{eq:app:terminal-L1}
\EE\| \hat u_1-u_1\|
\;\le\;
e^{L}\,\bigl(\varepsilon_{\mathrm{fit}}+\varepsilon_{\mathrm{curv}}\bigr).
\end{equation}
Moreover, the same argument in $L^2$ yields
\begin{equation}
\label{eq:app:terminal-L2}
\EE\| \hat u_1-u_1\|^2
\;\le\;
C(L)\,\bigl(\varepsilon_{\mathrm{fit}}^2+\varepsilon_{\mathrm{curv}}^2\bigr),
\qquad C(L)\sim e^{2L}.
\end{equation}
\end{theorem}

\begin{proof}
Set $e_\tau:=\hat u_\tau-u_\tau$. Then
\[
\dot e_\tau
= v_\theta(\hat u_\tau,\tau)-v_\star(u_\tau,\tau)
= \bigl[v_\theta(\hat u_\tau,\tau)-v_\theta(u_\tau,\tau)\bigr]
  + \bigl[v_\theta(u_\tau,\tau)-v_\star(u_\tau,\tau)\bigr].
\]
Take norms and use Lipschitz continuity:
\[
\frac{d}{d\tau}\|e_\tau\|
\le L\|e_\tau\| + \|v_\theta(u_\tau,\tau)-v_\star(u_\tau,\tau)\|.
\]
Integrate from $0$ to $1$ and apply Gr\"onwall:
\[
\|e_1\|
\le
e^{L}\int_0^1 \|v_\theta(u_\tau,\tau)-v_\star(u_\tau,\tau)\|\,d\tau.
\]
Now add and subtract $\bar v_\theta(u_\tau)$ and use triangle inequality:
\[
\|v_\theta(u_\tau,\tau)-v_\star(u_\tau,\tau)\|
\le
\|v_\theta(u_\tau,\tau)-\bar v_\theta(u_\tau)\|
+
\|\bar v_\theta(u_\tau)-v_\star(u_\tau,\tau)\|.
\]
Take expectations and apply Cauchy--Schwarz in $\tau$ to obtain \eqref{eq:app:terminal-L1}.
For \eqref{eq:app:terminal-L2}, repeat with $\|e_\tau\|^2$:
\[
\frac{d}{d\tau}\|e_\tau\|^2
= 2 e_\tau\cdot\dot e_\tau
\le 2L\|e_\tau\|^2 + 2\|e_\tau\|\,\|v_\theta(u_\tau,\tau)-v_\star(u_\tau,\tau)\|,
\]
and apply Young’s inequality plus Gr\"onwall. The resulting constant is $C(L)\sim e^{2L}$.
\end{proof}

\paragraph{Empirical tie.}
Theorem~\ref{thm:app:terminal} explains the two central practical observations:
(i) if the learned velocity is nearly \emph{constant in $\tau$} (small straightness error), then the
continuous-time mismatch is small, and
(ii) straighter trajectories are easier to integrate, which translates into fewer solver steps at
inference (Section~\ref{app:theory:discretization}).
This is exactly the mechanism behind “8 steps vs.\ 128 steps” in the main text and the sharp spectral
tracking in Figure~\ref{fig:spec_s1_E}: early steps already reconstruct high wavenumbers because the
trajectory is not spending many steps “turning” in rectification time.

\section{Discretization: why curvature controls step count, and why EMA helps}
\label{app:theory:discretization}

\subsection{Local truncation error and the curvature driver}

We start from a clean Taylor expansion for the ODE $\dot u=v(u,\tau)$.

\begin{lemma}[Local truncation error for explicit Euler]
\label{lem:app:lte}
Assume $v\in C^1$ in $(u,\tau)$ on the region traversed and that
$\|\partial_\tau v(u,\tau)\|\le M_\tau$, $\|J_u v(u,\tau)\|\le L$, and $\|v(u,\tau)\|\le M$.
Then one explicit-Euler step from $(u,\tau)$ with step size $\Delta\tau$ has local truncation error
\begin{align}
\mathrm{LTE}(\tau;\Delta\tau)
&:= \bigl\|u(\tau+\Delta\tau)-\bigl(u(\tau)+\Delta\tau\,v(u(\tau),\tau)\bigr)\bigr\| \nonumber\\
&= \frac{(\Delta\tau)^2}{2}\,
\bigl\|\partial_\tau v(u,\tau)+J_u v(u,\tau)\,v(u,\tau)\bigr\|
\;+\; \cO\!\bigl((\Delta\tau)^3\bigr) \label{eq:app:lte-expansion}\\
&\le \frac{(\Delta\tau)^2}{2}\,(M_\tau+LM)
\;+\; \cO\!\bigl((\Delta\tau)^3\bigr). \nonumber
\end{align}
\end{lemma}

\begin{proof}
Differentiate $\dot u=v(u,\tau)$ along the trajectory:
\[
\ddot u(\tau)=\partial_\tau v(u(\tau),\tau)+J_u v(u(\tau),\tau)\,\dot u(\tau)
=\partial_\tau v(u(\tau),\tau)+J_u v(u(\tau),\tau)\,v(u(\tau),\tau).
\]
Taylor expand:
\[
u(\tau+\Delta\tau)=u(\tau)+\Delta\tau\,v(u(\tau),\tau)+\frac{(\Delta\tau)^2}{2}\ddot u(\tau)+\cO((\Delta\tau)^3).
\]
Subtract the Euler update and take norms to obtain \eqref{eq:app:lte-expansion}. The bound follows by
triangle inequality and the assumed sup bounds.
\end{proof}

\paragraph{Interpretation.}
Lemma~\ref{lem:app:lte} identifies two drivers of discretization error:
\[
\|\partial_\tau v\| \quad\text{(time curvature)}\qquad\text{and}\qquad \|J_uv\,v\| \quad\text{(spatial nonlinearity).}
\]
Rectified flows are designed so that trajectories are \emph{nearly straight in $\tau$}, i.e.\
$\partial_\tau v$ is small along typical paths. This is the fundamental reason they can use large
steps without losing fidelity.

\subsection{Global error: straightness reduces the required number of steps}

\begin{corollary}[Global Euler error controlled by time-curvature]
\label{cor:app:global-euler}
Assume the hypotheses of Lemma~\ref{lem:app:lte} and that $v(\cdot,\tau)$ is $L$-Lipschitz in $u$
uniformly in $\tau$ on the region traversed.
Let $u^\Delta$ be explicit Euler with uniform step $\Delta\tau=1/N$.
Then there exists $C=C(L)$ such that
\begin{equation}
\label{eq:app:global-euler}
\sup_{\tau\in[0,1]}\|u^\Delta_\tau-u_\tau\|
\;\le\;
C\,\Delta\tau\,
\int_0^1
\Big(\|\partial_\tau v(u_\tau,\tau)\|
     +\|J_u v(u_\tau,\tau)\,v(u_\tau,\tau)\|\Big)\,d\tau,
\end{equation}
up to higher-order terms.
\end{corollary}

\begin{proof}
Write the one-step error recursion
$e_{n+1} \le (1+L\Delta\tau)e_n + \mathrm{LTE}_n$ and iterate it over $n=0,\dots,N-1$.
Use $\prod_{k=0}^{N-1}(1+L\Delta\tau)\le e^{L}$ and sum the LTE bounds from
Lemma~\ref{lem:app:lte}. The integral form \eqref{eq:app:global-euler} is obtained by viewing the
sum as a Riemann approximation of $\int_0^1(\cdots)d\tau$.
\end{proof}

\paragraph{Empirical tie.}
Equation \eqref{eq:app:global-euler} explains the step-count gap we observe:
if $\int_0^1\|\partial_\tau v\|$ is small (straight trajectories), then the $N$ required for a fixed
tolerance is small. Diffusion-class samplers typically integrate a much more curved reverse-time
dynamics (in addition to stochasticity), which translates into much larger effective curvature and
hence many more steps. This is also consistent with the qualitative “trajectory inpainting”
comparison in Figure~\ref{fig:traj_rho_s1_main}.

\subsection{EMA curvature proxy and blending as a Tikhonov correction}

\paragraph{Discrete-time EMA as a low-pass filter.}
During sampling we evaluate $v_t:=v_\theta(u_t,\tau_t)$ at a discrete set of times $\tau_t$ and
maintain an exponential moving average (EMA)
\[
v^{\mathrm{ema}}_t=\lambda\,v^{\mathrm{ema}}_{t-1}+(1-\lambda)\,v_t,\qquad \lambda\in(0,1).
\]
If $v_t$ varies slowly with $\tau_t$, then $v^{\mathrm{ema}}_t$ tracks it closely; if $v_t$ changes
quickly, the EMA lags and the deviation
\[
s_t:=\|v_t-v^{\mathrm{ema}}_t\|
\]
spikes. This motivates $s_t$ as a computable curvature/straightness proxy.

A simple bound shows $s_t$ controls a discrete time-variation scale.

\begin{lemma}[EMA deviation controls incremental variation]
\label{lem:app:ema}
Let $\{v_t\}_{t\ge0}$ be any sequence in a normed space and define
$v_t^{\mathrm{ema}}=\lambda v_{t-1}^{\mathrm{ema}}+(1-\lambda)v_t$ with $v_0^{\mathrm{ema}}=v_0$.
Then
\[
\|v_t-v_t^{\mathrm{ema}}\|
\le
(1-\lambda)\sum_{j=0}^{t-1}\lambda^{j}\,\|v_{t-j}-v_{t-j-1}\|.
\]
In particular, if $\|v_{k}-v_{k-1}\|\le \delta$ for all $k$, then
$\|v_t-v_t^{\mathrm{ema}}\|\le \delta$ for all $t$.
\end{lemma}

\begin{proof}
Unroll the recursion:
$v_t^{\mathrm{ema}}=(1-\lambda)\sum_{j=0}^{t}\lambda^{j}v_{t-j}$.
Subtract from $v_t$ and telescope:
\[
v_t-v_t^{\mathrm{ema}}=(1-\lambda)\sum_{j=0}^{t-1}\lambda^j\,(v_t-v_{t-j-1}),
\]
then use triangle inequality and rewrite $v_t-v_{t-j-1}$ as a sum of increments.
\end{proof}

\paragraph{Blending as a quadratic minimizer.}
We use the blended update
\[
\tilde v_t=(1-\alpha_t)v_t+\alpha_t v_t^{\mathrm{ema}},
\qquad \alpha_t\in[0,1].
\]
This is the unique minimizer of a Tikhonov-regularized quadratic:
\[
\tilde v_t
=
\argmin_{w}
\Big\{
\|w-v_t\|^2 + \beta_t\|w-v_t^{\mathrm{ema}}\|^2
\Big\},
\qquad
\alpha_t=\frac{\beta_t}{1+\beta_t}.
\]
Thus blending is not an ad-hoc smoothing step: it is the closed-form solution of a principled
stabilization objective trading fidelity to the instantaneous network prediction against adherence to
a locally averaged trend.

\paragraph{Step-size rule from LTE control.}
Lemma~\ref{lem:app:lte} suggests choosing $\Delta\tau_t$ so that
\[
(\Delta\tau_t)^2\Big(\|\partial_\tau v\| + \|J_u v\,v\|\Big)\approx \text{const}.
\]
Since $\|\partial_\tau v\|$ is not directly available, we replace it by a multiple of the proxy
$s_t$ (Lemma~\ref{lem:app:ema}), and we upper-bound $\|J_u v\,v\|\lesssim L\|v_t\|$ using Lipschitz
control. This yields the square-root-type rule in the main text:
\[
\Delta\tau_t \propto \frac{1}{\sqrt{\kappa_1 s_t+\kappa_2}},
\qquad \kappa_2\approx L\|v_t\|.
\]
This is exactly the mechanism behind the “curvature-aware integration” contribution: when the
trajectory bends (large $s_t$), we both (i) regularize the velocity via blending and (ii) reduce the
step size to control the discretization term in \eqref{eq:app:global-euler}.

\section{Multiscale coverage: structure functions control unavoidable law tails}
\label{app:theory:coverage}

\subsection{Why spectra/structure functions appear in a law-level bound}

\paragraph{Finite resolution is a hard constraint.}
All models operate on a fixed grid (e.g.\ $128^2$ for training/evaluation). This enforces an
effective Fourier cutoff $K_{\max}\simeq N$ (Nyquist), and architectural choices may impose an even
lower effective cutoff $K_c$. Therefore, even a perfect generative model cannot represent energy
beyond that cutoff. The best it can do is match the \emph{projected} law $(P_{\le K_c})_\#\mu$.

\paragraph{Structure functions quantify tail energy.}
In turbulence, small-scale content is captured by structure functions and energy spectra. This is
exactly why our empirical analysis emphasizes spectral evolution along $\tau$
(Figure~\ref{fig:spec_s1_E}). The theory below formalizes the link:
a structure-function modulus bound implies a Fourier-tail bound, which becomes a coverage term in
$\Wtwo$.

\subsection{Bandlimited fields: capacity bounds and why missing an annulus matters}

We record two standard Fourier facts to make the “capacity” language explicit.

\begin{lemma}[Bernstein upper bound for bandlimited fields]
\label{lem:app:bernstein}
Let $u:\T^d\to\R^m$ satisfy $u=P_{\le K}u$ for some integer $K\ge1$.
Then there exists $C_d>0$ such that
\[
\|\nabla_x u\|_{L^\infty(\T^d)}
\le
C_d\,K^{1+\frac d2}\,\|u\|_{L^2(\T^d)}.
\]
\end{lemma}

\begin{proof}
Expand $u(x)=\sum_{|k|\le K}\hat u(k)e^{ik\cdot x}$, write $\nabla u$ as a Fourier series, and apply
Cauchy--Schwarz using that $\#\{|k|\le K\}\lesssim K^d$ and Parseval.
\end{proof}

\begin{lemma}[Annulus lower bound]
\label{lem:app:annulus}
Let $f=P_{[K,2K]}u$ and $K\ge1$. Then there exists $c_d>0$ such that
\[
\|\nabla_x f\|_{L^\infty}
\ge
c_d\,K^{1-\frac d2}\,\|f\|_{L^2}.
\]
\end{lemma}

\begin{proof}
On the annulus $K\le|k|\le 2K$, the Fourier symbol of $\nabla$ has magnitude $\simeq K$.
A reverse Bernstein estimate on an annulus yields $\|f\|_{L^\infty}\gtrsim K^{-d/2}\|f\|_{L^2}$,
and $\|\nabla f\|_{L^\infty}\gtrsim K\|f\|_{L^\infty}$, giving the claim.
\end{proof}

\paragraph{Empirical tie.}
Lemma~\ref{lem:app:annulus} formalizes why “spectral collapse” is fatal for turbulent statistics:
if a surrogate misses energy in a whole annulus, it underestimates strain and small-scale variability.
This is exactly what appears visually when deterministic baselines blur fine scales and also what is
quantified by the spread errors in Table~\ref{tab:results}.

\subsection{Structure functions imply Fourier-tail control (law-level)}

For $u:\T^d\to\R^m$ define the second-order structure function
\[
S_r^2(u)
:= \frac{1}{|B_r|}\int_{|h|\le r}
\|u(\cdot+h)-u(\cdot)\|_{L^2(\T^d)}^2\,dh,
\qquad
S_r^2(\mu):=\int S_r^2(u)\,d\mu(u).
\]
Following the statistical-solution framework of \cite{lanthaler2021statistical}, assume there exists
a modulus $\omega(r)\downarrow 0$ as $r\downarrow0$ such that
\begin{equation}
\label{eq:app:sf-modulus}
S_r^2(\mu_t)\le \omega(r)\qquad\text{for all }r\in(0,1],\ t\in[0,T].
\end{equation}
In practice, $\omega$ can be calibrated from empirical spectra (Figure~\ref{fig:spec_s1_E} and
supplementary plots).

\begin{lemma}[Structure function controls Fourier tail]
\label{lem:app:sf-tail}
Let $\mu$ be a law on $L^2(\T^d)$ satisfying $S_r^2(\mu)\le\omega(r)$ for all $r\in(0,1]$.
Then there exist $c_d,C_d>0$ such that for all integers $K\ge1$,
\[
\int \|P_{>K}u\|_{L^2}^2\,d\mu(u)
\le
C_d\,\omega\!\bigl(c_d/K\bigr).
\]
\end{lemma}

\begin{proof}
For fixed $u$, expand
$\|u(\cdot+h)-u(\cdot)\|_{L^2}^2=\sum_k |\hat u(k)|^2|e^{ik\cdot h}-1|^2$.
Average over $|h|\le r$ to obtain
$S_r^2(u)=\sum_k |\hat u(k)|^2\Psi_r(k)$ where
$\Psi_r(k)=\frac1{|B_r|}\int_{|h|\le r}|e^{ik\cdot h}-1|^2\,dh$.
A standard oscillation estimate yields $\Psi_r(k)\ge c\min\{(|k|r)^2,1\}$.
Hence $\Psi_r(k)\ge c_0$ for all $|k|\ge c'/r$, and therefore
$S_r^2(u)\ge c_0\sum_{|k|\ge c'/r}|\hat u(k)|^2=c_0\|P_{>c'/r}u\|_{L^2}^2$.
Integrate in $u$ and set $r=c_d/K$.
\end{proof}

\begin{lemma}[Bandlimited projection in $\Wtwo$]
\label{lem:app:proj-W2}
For any law $\mu$ on $L^2(\T^d)$ and cutoff $K_c\ge1$,
\[
\Wtwo\bigl(\mu,(P_{\le K_c})_\#\mu\bigr)
\le
\Bigl(\int \|P_{>K_c}u\|_{L^2}^2\,d\mu(u)\Bigr)^{1/2}.
\]
\end{lemma}

\begin{proof}
Couple $u\sim\mu$ with $\tilde u=P_{\le K_c}u$. Then $\tilde u\sim (P_{\le K_c})_\#\mu$ and
$\Wtwo^2\le \EE\|u-\tilde u\|_{L^2}^2=\EE\|P_{>K_c}u\|_{L^2}^2$.
\end{proof}

\begin{corollary}[Coverage term from structure-function modulus]
\label{cor:app:coverage}
Under \eqref{eq:app:sf-modulus}, for any cutoff $K_c\ge1$,
\[
\Wtwo\bigl(\mu_t,(P_{\le K_c})_\#\mu_t\bigr)
\lesssim
\omega(c_d/K_c)^{1/2}.
\]
\end{corollary}

\paragraph{Empirical tie.}
Corollary~\ref{cor:app:coverage} is the clean mathematical version of what the spectra show:
at fixed resolution, there is an unavoidable tail beyond the cutoff. Both ReFlow and GenCFD operate
at the same grid resolution (hence comparable $K_c$), so neither can beat the coverage floor.
The difference is that ReFlow reaches the best attainable projected law with far fewer steps
(Figure~\ref{fig:spec_s1_E}, Figure~\ref{fig:traj_rho_s1_main}).

\section{One-step law error for PDE pushforwards and training fit}
\label{app:theory:one-step}

Let $S_{\Delta t}:\Hcal\to\Hcal$ denote the PDE solution operator over a physical step $\Delta t$ and
$\mu_{t+\Delta t}=(S_{\Delta t})_\#\mu_t$. Let $\Tcal^\theta_{\Delta t}$ be the learned one-step
surrogate and $\hat\mu_{t+\Delta t}=(\Tcal^\theta_{\Delta t})_\#\mu_t$.

\paragraph{Why we insert a bandlimit.}
Since the model outputs live at finite resolution, it is natural to compare to the \emph{projected}
PDE output $P_{\le K_c}S_{\Delta t}u$. This isolates (i) an irreducible coverage term and (ii) a
trainable fit term.

Define the mean-square training error at time $t$:
\begin{equation}
\label{eq:app:eps-train}
\varepsilon_{\mathrm{train}}^2(t)
:=
\EE_{u\sim\mu_t}\Bigl[\,
\bigl\|P_{\le K_c}(S_{\Delta t}u)-\widehat u\bigr\|_{L^2}^2
\Bigr],
\end{equation}
where $\widehat u$ is the model output given input $u$ (i.e.\ after integrating the learned RF ODE).

\begin{proposition}[One-step law error = coverage + fit]
\label{prop:app:onestep}
Assume $\mu_{t+\Delta t}$ satisfies the structure-function modulus \eqref{eq:app:sf-modulus} (hence
Corollary~\ref{cor:app:coverage}).
Then
\[
\Wtwo\bigl(\mu_{t+\Delta t},\hat\mu_{t+\Delta t}\bigr)
\;\le\;
\underbrace{\Wtwo\bigl(\mu_{t+\Delta t},(P_{\le K_c})_\#\mu_{t+\Delta t}\bigr)}_{\text{coverage}}
\;+\;
\underbrace{\bigl(\varepsilon_{\mathrm{train}}^2(t)\bigr)^{1/2}}_{\text{fit}},
\]
and therefore
\[
\Wtwo\bigl(\mu_{t+\Delta t},\hat\mu_{t+\Delta t}\bigr)
\;\lesssim\;
\omega(c_d/K_c)^{1/2}+\varepsilon_{\mathrm{train}}(t).
\]
\end{proposition}

\begin{proof}
Use the triangle inequality in $\Wtwo$:
\[
\Wtwo(\mu_{t+\Delta t},\hat\mu_{t+\Delta t})
\le
\Wtwo\bigl(\mu_{t+\Delta t},(P_{\le K_c})_\#\mu_{t+\Delta t}\bigr)
+
\Wtwo\bigl((P_{\le K_c})_\#\mu_{t+\Delta t},\hat\mu_{t+\Delta t}\bigr).
\]
For the second term, couple by sampling $u\sim\mu_t$ and using the same input to produce
$P_{\le K_c}S_{\Delta t}u$ and $\widehat u$. Then
\[
\Wtwo^2\bigl((P_{\le K_c})_\#\mu_{t+\Delta t},\hat\mu_{t+\Delta t}\bigr)
\le
\EE\|P_{\le K_c}S_{\Delta t}u-\widehat u\|_{L^2}^2
=\varepsilon_{\mathrm{train}}^2(t).
\]
The second display follows from Corollary~\ref{cor:app:coverage}.
\end{proof}

\paragraph{Empirical tie.}
Proposition~\ref{prop:app:onestep} explains why training curves saturate at a resolution-dependent
floor and why spectra are the correct diagnostic for that floor.
It also makes explicit that “better training loss” means smaller law error \emph{only up to} the
coverage limit. This aligns with the observation that increasing diffusion steps improves spectral
reconstruction but cannot beat finite-resolution constraints, while ReFlow reaches that regime with
far fewer steps.

\section{Master inequality: coverage + fit + straightness + discretization}
\label{app:theory:master}

We now combine the law-level coverage/fit decomposition with the continuous-time RF mismatch and the
discretization control.

\begin{theorem}[Master law-level bound for learned rectified flows]
\label{thm:app:master}
Fix an effective cutoff $K_c$ and assume the structure-function modulus \eqref{eq:app:sf-modulus} for
the target PDE laws.
Let $\hat\mu_1^{(N)}$ be the law obtained by integrating the learned rectified ODE
$\dot u_\tau=v_\theta(u_\tau,\tau)$ from $\tau=0$ to $\tau=1$ using $N$ explicit-Euler steps (uniform
step $1/N$). Assume $v_\theta(\cdot,\tau)$ is $L$-Lipschitz in $u$ uniformly in $\tau$.

Then, for constants depending only on $L$,
\begin{equation}
\label{eq:app:master}
\Wtwo(\mu_1,\hat\mu_1^{(N)})
\;\lesssim\;
\underbrace{\omega(c_d/K_c)^{1/2}}_{\text{coverage}}
\;+\;
\underbrace{\varepsilon_{\mathrm{fit}}+\varepsilon_{\mathrm{curv}}}_{\text{fit + straightness}}
\;+\;
\underbrace{\frac{1}{N}\int_0^1\|\partial_\tau v_\theta(u_\tau,\tau)\|\,d\tau}_{\text{discretization}},
\end{equation}
where $\varepsilon_{\mathrm{fit}},\varepsilon_{\mathrm{curv}}$ are the continuous-time terms from
Theorem~\ref{thm:app:terminal} evaluated along the ideal trajectory.
\end{theorem}

\begin{proof}
(1) \emph{Coverage.} By Corollary~\ref{cor:app:coverage},
$\Wtwo(\mu_1,(P_{\le K_c})_\#\mu_1)\lesssim\omega(c_d/K_c)^{1/2}$.

(2) \emph{Continuous-time learned flow error.}
Let $\hat\mu_1$ denote the law produced by the \emph{exact} (continuous-time) flow of $v_\theta$.
Then Theorem~\ref{thm:app:terminal} implies a bound of the form
$\Wtwo((P_{\le K_c})_\#\mu_1,\hat\mu_1)\lesssim \varepsilon_{\mathrm{fit}}+\varepsilon_{\mathrm{curv}}$
(up to the standard coupling-by-same-input argument on the finite-dimensional surrogate).

(3) \emph{Discretization.} Let $\hat\mu_1^{(N)}$ be the Euler-discretized law.
Corollary~\ref{cor:app:global-euler} bounds the trajectory error by a term proportional to
$\frac1N\int_0^1\|\partial_\tau v_\theta\|\,d\tau$ (plus the spatial term, absorbed in constants under
the Lipschitz assumption and bounded $\|v_\theta\|$ on the region traversed).

(4) Combine with triangle inequalities:
\[
\Wtwo(\mu_1,\hat\mu_1^{(N)})
\le
\Wtwo(\mu_1,(P_{\le K_c})_\#\mu_1)
+
\Wtwo((P_{\le K_c})_\#\mu_1,\hat\mu_1)
+
\Wtwo(\hat\mu_1,\hat\mu_1^{(N)}),
\]
and insert the three bounds.
\end{proof}

\paragraph{What Theorem~\ref{thm:app:master} explains in the paper.}
\begin{itemize}
\item \textbf{Why rectification accelerates sampling.}
The discretization term depends on the integral of time-curvature.
Straightness reduces that integral, so fewer steps suffice (8--10 in our experiments).

\item \textbf{Why curvature-aware integration helps under shift.}
Under distribution shift, the learned velocity may bend more in $\tau$, increasing the last term.
The EMA proxy and blending/step control reduce the effective curvature and hence stabilize sampling,
which is exactly how curvature is used in our method (control, not OOD detection).

\item \textbf{Why spectra matter.}
The coverage term is dictated by small-scale physics (structure functions / spectrum).
This is why Figure~\ref{fig:spec_s1_E} is the right diagnostic: it visualizes how quickly the method
reconstructs high-frequency energy up to the grid cutoff, and it clarifies what errors are
unavoidable at fixed resolution.

\item \textbf{Why deterministic baselines struggle.}
Deterministic predictors trained by pointwise losses effectively target conditional means; in
multimodal chaotic regimes this collapses variability (as emphasized by \cite{molinaro2024generative}).
In the master inequality language, they do not produce a good fit to the law itself, which is seen
in their larger spread and Wasserstein errors in Table~\ref{tab:results}.
\end{itemize}

\section{From law-level bounds to reported pointwise metrics}
\label{app:theory:metrics-bridge}

\paragraph{From $\Wtwo$ to marginal discrepancies.}
While the paper reports one-point $\Wone$ (averaged over space) for computational tractability, the
mechanisms above are consistent with those marginal tests:
if two laws on fields are close in $\Wtwo$ and have bounded second moments, then many bounded-Lipschitz
observables of the fields (including smoothed one-point marginals) are close as well.
This is the conceptual bridge between the law-level theory and the reported statistics.

\paragraph{Why we emphasize spectra during sampling.}
The spectral evolution plots (Figure~\ref{fig:spec_s1_E} and supplementary variants) act as a
\emph{scale-resolved} diagnostic: they visualize the progressive recovery of high frequencies along
$\tau$. This connects directly to the structure-function/coverage viewpoint (Section~\ref{app:theory:coverage})
and empirically demonstrates that ReFlow reaches the best attainable high-frequency reconstruction
regime in far fewer steps than diffusion-based baselines.

\section{Architecture of the ReFlow Model}
\label{sec:rectifiedflow-design}

\begin{figure}[h]
  \centering
  \includegraphics[width=1.0\linewidth]{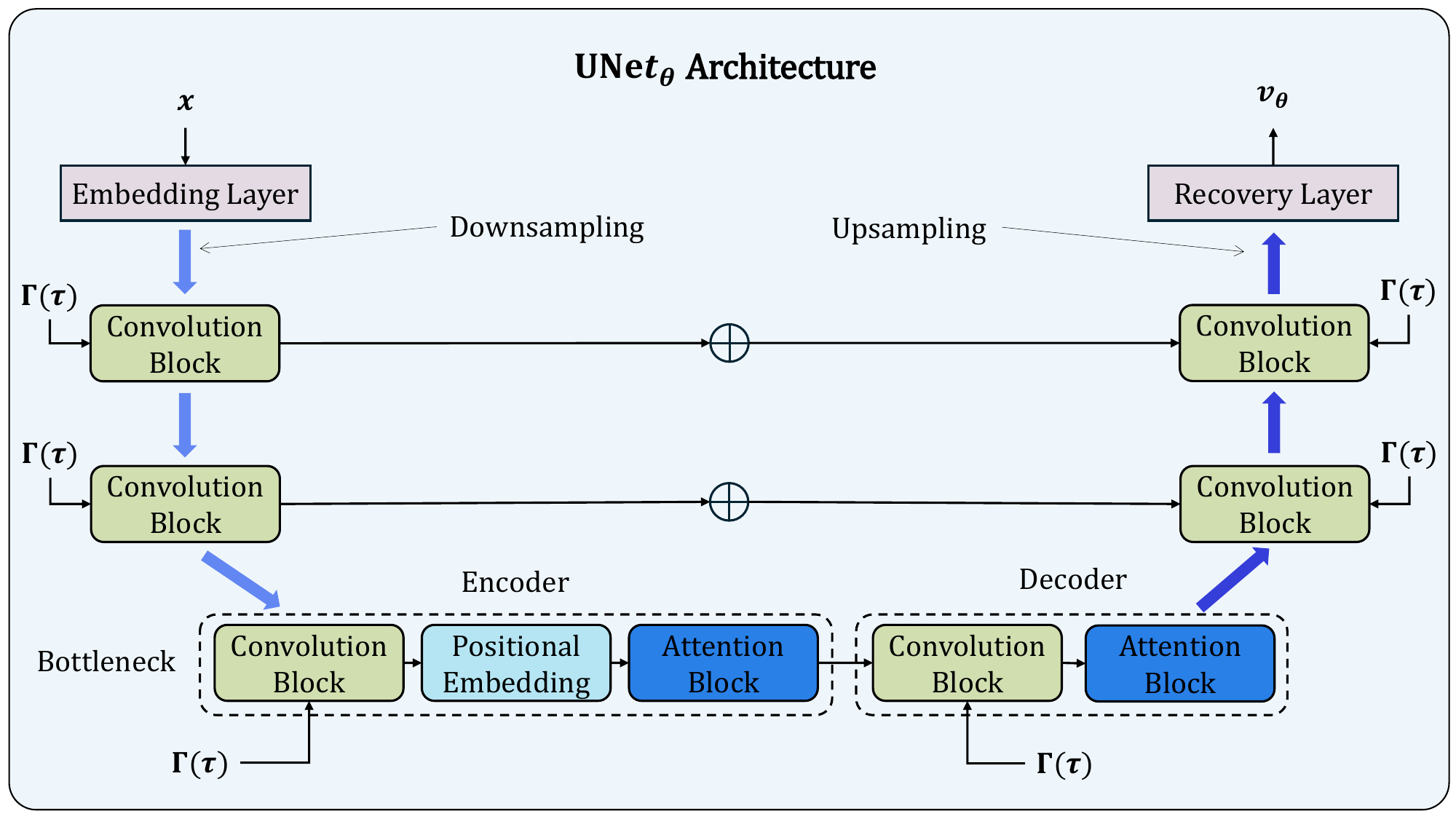}
  \caption{\textbf{UNet Backbone Architecture Used in ReFlow.} This schematic illustrates the core UNet-based architecture used within the ReFlow framework, structured across three resolution levels. For clarity, the number of blocks per level is set to one in this illustration. In the actual GenCFD and ReFlow configurations, each block is repeated four times per level. The bottleneck shows an asymmetry between the encoder and decoder sides: the block on the encoder side includes a Convolution Block, Positional Embedding, and Attention Block, while the corresponding block on the decoder side omits the Positional Embedding.}
  \label{fig:RFSchematics}
\end{figure}

\subsection{Architectural Components}
\label{sec:unet-architecture}

\textbf{Motivation.}
Multi-scale PDE data contain wide-ranging spatial scales, so a \emph{multi-resolution} encoder--decoder is natural. We incorporate time embeddings to condition each layer on the interpolation fraction \(t\). 

\textbf{Architecture Layout.}
We employ a UNet-based architecture augmented with attentional layers, typically termed as \textbf{UViT} by \cite{saharia2022photorealistic}. In addition to the MLP-based time embedding \(\Gamma(\tau)\), the input signal is lifted into a higher-dimensional embedding space and later projected back into physical space through convolutional layers. The data flows from high to low resolution across a down-sampling stack composed of convolution blocks into a bottleneck layer that encodes the fine-scale features of the inputs across its channel dimension and mixes them via a multi-head attention mechanism. 
\newline
On the encoder side, exclusively at the bottleneck level, each attention block preceded by a simple linear positional embedding. This positional encoding helps preserve spatial locality before global mixing through attention. The decoder side does not apply any positional embedding. 
\newline
The UNet is conditioned at every level through normalized Feature-wise Linear Modulation (FiLM) techniques embedded within the convolution blocks. The conditioning signal modulates the normalization layers via learned scale and shift parameters. This conditioning mechanism is consistently applied across all encoder and decoder stages.
\newline
The output is symmetrically reconstructed via an up-sampling pipeline. Downsampling is performed using standard convolutional layers with stride, while upsampling can be implemented either via transposed convolutions or by applying a standard convolution followed by a non-learnable pixel shuffle operation, which rearranges elements of the tensor spatially according to a fixed upsampling factor.
\newline
Importantly, the architecture described here resembles the one presented in \cite{molinaro2024generative}, ensuring consistency with prior state-of-the-art practices in high-fidelity generative modeling for scientific computing.

\subsubsection{Convolution Block}
Each UNet level employs dedicated convolution blocks that process features while enabling conditioning on the diffusion time $\tau$. These blocks consist of two convolutional layers interleaved with group normalization and Swish activations. Between the convolutions, a normalized FiLM layer adaptively scales features using the conditioning embeddings. A residual connection combines the processed features with a projected skip connection, ensuring stable gradient flow. Crucially, these convolution blocks form the fundamental building units across all encoder and decoder stages, not just the bottleneck, with consistent application of temporal conditioning through the FiLM mechanism. 

\subsubsection{Normalized FiLM Layer}
Each convolution block uses a normalized FiLM layer and applies a convolution layer interleaved with normalization and an activation, then adds a residual skip:
\begin{equation}
\label{eq:res-block}
\begin{aligned}
    \mathbf{z} \;&=\; \mathrm{Conv}\bigl(\mathrm{Norm}(\mathbf{x})\bigr),
    \qquad 
    \mathbf{z} \;=\; \sigma(\mathbf{z} + \alpha \,\gamma(t)), \\
    \mathbf{x}_{\mathrm{out}} 
    \;&=\; 
    \mathbf{x} \;+\; \mathrm{Conv}\!\bigl(\mathbf{z}\bigr),
\end{aligned}
\end{equation}
where \(\mathrm{Norm}\) is often \(\texttt{RMSNorm}\), GroupNorm or LayerNorm, \(\sigma\) is a nonlinearity (e.g.\ \(\mathrm{SiLU}\)), and \(\gamma(t)\) represents the time embedding (an MLP on \(\mathrm{Sinusoid}(t)\)) that modulates scale/shift. This temporal conditioning is crucial for guiding the generative process through rectified diffusion trajectories.

\subsubsection{Attention Block}
\label{sec:attention}
The architecture strategically employs multi-head attention exclusively at the bottleneck layer to balance global interaction modeling with computational efficiency. Each attention block processes normalized features through a spatial-channel reformatting, where features are reshaped by a flattening operation. This  preserves the channel structure while collapsing the spatial grid into a single flattened dimension, allowing attention to operate across spatial locations. At certain scales, a multi-head attention (MHA) mechanism captures distant flow interactions.  We define
\begin{equation}
\label{eq:attention}
    \mathbf{Q} \;=\; W_Q\,\mathbf{h}, \quad
    \mathbf{K} \;=\; W_K\,\mathbf{h}, \quad
    \mathbf{V} \;=\; W_V\,\mathbf{h},
    \quad
    \mathcal{A}(\mathbf{h})
    \;=\;
    \mathrm{softmax}\!\Bigl(\tfrac{\mathbf{Q}\mathbf{K}^\top}{\sqrt{d}}\Bigr)\,\mathbf{V},
\end{equation}
where \(\mathbf{h} \in \mathbb{R}^{(H \cdot W) \times d}\) denotes the flattened input sequence, with each of the \(H \cdot W\) spatial positions represented as a \(d\)-dimensional feature vector. In practice, attention is split into multiple heads for better representational power, then recombined. This global mixing is key for PDE flows dominated by far-field couplings (e.g.\ vortex merging). The MHA operator is defined as
\begin{equation}
\label{eq:attention}
    \mathrm{MHA}(\mathbf{h})
    \;=\;
    W^O \; \mathrm{Concat}\left(\mathcal{A}_1(\mathbf{h}), \dots,  \mathcal{A}_n(\mathbf{h})\right),
\end{equation}
where $n \in \mathbb{N}$ denotes the number of attention heads and $W^O \in \mathbb{R}^{C \times C}$ is a learned output projection matrix.
The attention block applies MHA to a normalized input, then adds a residual connection to preserve the original signal and stabilize training while enabling global context to influence local features:
\begin{equation}
\label{eq:attention-block}
\begin{aligned}
\mathbf{z} \;&=\; \mathrm{MHA}\bigl(\mathrm{Norm}(\mathbf{x})\bigr), \\
\mathbf{x}_{\mathrm{out}}
\;&=\;
\mathbf{x} \;+\; \mathbf{z}.
\end{aligned}
\end{equation}

\subsection{Training Pipeline Details}
\label{sec:hyperparams}
The training pipeline follows the algorithm described in the main text faithfully.

\paragraph{Trainer Workflow.}
A high-level \(\texttt{Trainer}\) class manages:
\begin{enumerate}
    \item Batching PDE pairs \((u_0,u_1)\). These normalized pixel-wise with statistics extracted per each entire dataset.
    \item Sampling \(\tau \in [0,1]\) and random noise \(\xi\) according to a given noise scheduler $\sigma(\cdot)$.
    \item Partial interpolation + noise addition: \(\tilde{u}_\tau := \tau\,u_1 + (1-\tau)\,\xi\) (or the flow-based variant). We have experimented with non-linear interpolation methods, but have not noticed meaningful oscillations in performance.
    \item Minimizing the squared deviation between model output and target displacement/noise, optionally with a consistency/EMA update.
    \item Throughout training, we monitor validation performance on a separate holdout dataset across all main metrics. We explicitly note that different validation splits have been tested out to ensure there is no data leakage that would bias performance metrics.
\end{enumerate}

\FloatBarrier
\begin{table}[t]
  \centering
  \caption{Representative hyperparameters for 2D fluid-flow tasks. See main text for justification of these choices.}
  \label{tab:rf_params}  
  \vspace{5pt}

  \begin{tabular}{lcc}
    \toprule
    \textbf{Parameter}    & \textbf{Value (Optimal empirical performance)} & \textbf{Notes} \\
    \midrule
    Channels per level    & (128, 256, 256)  &  Number of channels per level. \\
    Downsample Ratios     & (2, 2, 2)        & Downsampling ratio per level. \\
    Attention Heads       & 8                & Each attention block. \\
    Dropout Rate          & 0.1              & In ConvBlock for generalization. \\
    Noise Schedule        & \texttt{uniform}, \texttt{log-uniform}, \texttt{cos-map} & Helps with multi-scale noise. \\
    ODE Steps (Sampling)  & 4--8             & Often $>15\times$ fewer vs.\ diffusion. \\
    Batch Size            & 16               & Tied to GPU capacity. \\
    EMA Decay             & 0.9999           & Teacher model for consistency. \\
    Learning Rate         & $3\times10^{-4}$ & Cosine decay in code. \\
    \bottomrule
  \end{tabular}
\end{table}
\FloatBarrier

In practice we noticed that the most robust denoising schedule was just the \texttt{uniform} one. For details on the \texttt{cos-map} schedule, we direct the reader to its description in \cite{esser2024scaling}.

\subsection{Summary and Key Takeaways}
The \(\texttt{RectifiedFlow}\) class, combined with a time-aware \(\texttt{UNet2D}\) architecture, builds a deterministic transport model well-suited to multi-scale PDE tasks. By interpolating between noise and data along nearly-straight paths, it reduces sampling overhead compared to conventional diffusion PDE solvers while retaining the flexibility to model complex turbulent phenomena. 
Further experimental and theoretical results appear in \textbf{SM} Section~\ref{app:results}.
Instead of the multi-step \emph{stochastic} integration in diffusion-based models, \texttt{RectifiedFlow} fits a \emph{deterministic} ODE from noise to data. This leverages straighter trajectories in frequency space, enabling fewer network evaluations at inference time.

\paragraph{ODE Integration for Sampling.}
At inference:
\[
  \frac{d \tilde{u}_\tau}{d\tau}
  \;=\;
  v_\theta\bigl(\tilde{u}_\tau,\,u_0,\,\tau\bigr),
  \quad \tau \in [0,1],
  \quad 
  \tilde{u}_{\tau=0} \,=\, \text{random noise}.
\]
We solve this with \(\texttt{odeint}\). Few steps (\(8\!-\!10\)) typically suffice, contrasting with the tens or hundreds required by full diffusion samplers.

\section{Datasets}
\subsection{Multi-scale Flow Datasets} 

Following the dataset generation procedure of Herde et al.\ \cite{herde2024poseidon}, the Richtmyer–Meshkov ensemble is created by imposing a randomized sinusoidal perturbation on a two‐fluid interface and driving it with a planar shock via prescribed pressure jumps.  For the CloudShock and ShearFlow benchmarks, we follow the GenCFD mesh‐generation and solver setup of Molinaro et al.\ \cite{molinaro2024generative}, initializing CloudShock with concentric density perturbations and ShearFlow with orthogonal shear jets on adaptive Cartesian grids, and employing the same high‐order finite‐volume scheme and dissipation settings to ensure consistent numerical fidelity across both datasets.  \\

\noindent\textbf{Overview.}  
We experiment on three paired‐field benchmarks \((u_0,u_1)\):

\begin{itemize}
  \item \textbf{Richtmyer–Meshkov (RM)}: 64\,000 samples  
  \item \textbf{Cloud-Shock (CS)}: 40\,000 samples  
  \item \textbf{Shear‐Layer (SL)}: 79\,200 samples  
\end{itemize}

All datasets track the $u_0 \to S^t(u_0) \vert_{t=1}$ solution mapping from inputs to the evolved outputs at $t=1$. The first two datasets monitor the evolution of density ($\rho$), momentum components $(m_x, m_y)$, and energy respectively ($E$). The shear-flow dataset monitors the velocity components of the flow $(u_x, u_y)$. 

\medskip\noindent\textbf{Train/Validation Split.}  
For each dataset, we reserve 80\% of the samples for training and 20\% for validation.

\medskip\noindent\textbf{Macro–Micro Ensemble Evaluation (CS \& SL).}  
To probe out‐of‐distribution generalization under small perturbations, we adopt a two‐stage ensemble protocol:

\begin{enumerate}
  \item \emph{Macro‐sampling:} select \(M_{\mathrm{macro}}=10\) distinct base initial conditions.  
  \item \emph{Micro‐perturbation:} for each base, generate \(M_{\mathrm{micro}}=1{,}000\) perturbed copies within a small radius.  
  \item \emph{Metrics:} compute all performance measures (mean‐field error \(e_\mu\), standard‐deviation error \(e_\sigma\), average \(W_1\), spectral agreement) by first averaging over each micro‐ensemble, then averaging across the \(M_{\mathrm{macro}}\) cases.
\end{enumerate}

\medskip\noindent\textbf{In‐Distribution Evaluation (RM).}  
For the Richtmyer–Meshkov dataset, we additionally evaluate on a held‐out test set drawn from the same distribution (no macro–micro perturbations).  This allows us to study how deterministic and probabilistic models scale with training set size under standard, in‐distribution conditions.

\subsection{Preprocessing and Data Organization} 

\subsubsection*{Data Processing} \label{app:data_processing}

Before training, we compute and store global, channel‐wise statistics over the entire training dataset.  Concretely, for each physical variable \(c\) we calculate
\[
\begin{aligned}
  \mu_c 
  &= \frac{1}{N_{\rm train}\,H\,W\,[D]}
     \sum_{i=1}^{N_{\rm train}}
     \sum_{x,y,[z]}
       u^{(i)}_c\bigl(x,y,[z],t_0\bigr), \\[0.5em]
  \sigma_c 
  &= \sqrt{ 
       \frac{1}{N_{\rm train}\,H\,W\,[D]}
       \sum_{i=1}^{N_{\rm train}}
       \sum_{x,y,[z]}
         \Bigl(u^{(i)}_c\bigl(x,y,[z],t_0\bigr) - \mu_c\Bigr)^2
     }.
\end{aligned}
\]

where \(N_{\rm train}\) is the number of training samples, \(H\times W\,[\times D]\) the spatial grid size, and \(u_c(x,y,[z],t_0)\) the initial field for channel \(c\).  We average over all spatial locations to yield a single mean \(\mu_c\) and standard deviation \(\sigma_c\) per channel.  The same statistics are also computed over the final‐time fields \(u_c(t_f)\), producing \(\mu_c'\) and \(\sigma_c'\) for the output channels.

During both training and evaluation, each input field \(u_c\) is standardized via
\[
  \tilde u_c = \frac{\,u_c \;-\;\mu_c\,}{\sigma_c + 10^{-16}},
  \qquad
  \tilde v_c = \frac{\,v_c \;-\;\mu_c'\,}{\sigma_c' + 10^{-16}},
\]
where \(u_c\) and \(v_c\) denote the initial and target fields, respectively.  We then concatenate \(\{\tilde u_c\}\) and \(\{\tilde v_c\}\) along the channel axis before feeding them to the network.  Because we always apply the same pre‐computed statistics at test time, no information from the evaluation set ever influences these normalization parameters.

\paragraph{Test‐time ensemble perturbation (evaluation only).} We adopt the same testing setup proposed in \emph{GenCFD} \cite{molinaro2024generative}. To evaluate how well our model captures the \emph{distribution} of solutions $u(t)$ arising from a fixed, chaotic initial field $\bar u$, we generate a small perturbed ensemble around $\bar u$ and evolve each member with a high‐fidelity PDE solver—\emph{but only at evaluation time}. Specifically:

\begin{enumerate}
  \item \textbf{Micro‐ensemble generation.}  Around the test initial condition $\bar u$, draw $M_{\mathrm{micro}}$ perturbed copies $\{\bar u_j\}_{j=1}^{M_{\mathrm{micro}}}$ by sampling uniformly in a ball of radius $\varepsilon$ centered at $\bar u$.
  \item \textbf{Reference propagation.}  Integrate each perturbed field $\bar u_j$ forward to time $t$ using the reference solver, producing end‐states $\{u_j(t)\}$.
  \item \textbf{Empirical conditional law.}  Form the empirical measure
  \[
    \hat P_t(\,\cdot\mid \bar u)
    \;=\;
    \frac{1}{M_{\mathrm{micro}}}
    \sum_{j=1}^{M_{\mathrm{micro}}}
    \delta_{u_j(t)},
  \]
  which approximates the true chaotic conditional law $P_t(\,\cdot\mid \bar u)$.
\end{enumerate}

When the input distribution itself is non‐degenerate, one would first draw $M_{\mathrm{macro}}$ base fields from that distribution and then apply the micro‐ensemble procedure to each.  For our Dirac initial‐condition tests we set $M_{\mathrm{macro}}=1$.

\paragraph{Training vs.\ evaluation.}  
During training the model only ever sees independent pairs $(u_0,u_1)$—one target per input—and never observes any ensemble.  The above ensemble perturbation is used \emph{only} to construct a ground‐truth distribution at test time.

\section{Models and baselines} \label{app:models}
We tested our method against $4$ other baselines, out of which $2$ (namely GenCFD and its FNO-conditioned variant) are diffusion-based algorithms of the same kind. The other two baselines rely on traditional operator learning setup, with the UViT baseline satisfying also the purpose of an ablation study that quantifies the gain obtained by adding diffusion on top. \\

\textbf{Shared Backbone}. We stress that for GenCFD, GenCFD $\circ$ FNO and UViT we use the exact same UViT backbone architecture outlined in the earlier parameterization of our rectified flow as in Table~\ref{tab:rf_params} for the sake of consistency. The architecture is further illustrated schematically in Figure~\ref{fig:RFSchematics}, which shows the UNet-based design in detail. 

\subsection*{GenCFD}
GenCFD~\cite{molinaro2024generative} is an end-to-end conditional score‐based diffusion model that directly learns the mapping from an input flow field \(u_0\) to the target field \(u_1\).  It uses the same UViT backbone described in Table~\ref{tab:rf_params}, with noise levels $\sigma_\tau$ corresponding to the standard deviation of the added Gaussian perturbation, in contrast to our ReFlow model which evolves over a rectified diffusion time $\tau\in[0,1]$. Common choices for noise schedulers include exponential and tangent formulations. In our case, we consistently adopted an exponential noise schedule, as it yielded the best results across all datasets and metrics reported. For score-based diffusion models, either variance-exploding (VE) or variance-preserving (VP) formulations are typically used. In all GenCFD models, we opted for the VE setting throughout. During training we minimize the denoising score‐matching loss over \((u_0,u_1)\) pairs sampled uniformly in \(\tau\in[0,1]\), using Adam with initial learning rate \(3\times10^{-4}\) (cosine decay), batch size 16, and EMA decay 0.9999. As is common in score-based diffusion models, we applied denoiser preconditioning and used uniform weighting across noise levels. At inference we solve the corresponding SDE (via an Euler-Maruyama scheme with discretization $128$ steps) and incorporate reconstruction guidance to condition on \(u_0\) so that we recover the conditional posterior measure $p(u_1 \vert u_0)$.

\subsection*{GenCFD \(\boldsymbol{\circ}\) FNO}
In this hybrid approach, a Fourier Neural Operator (FNO)~\cite{FNO} is first trained under an \(\ell_2\) loss to predict \(u_1\) from \(u_0\).  Its output \(\hat u_1^{\rm FNO}\) is concatenated channel‐wise with \(u_0\) and fed into the GenCFD score network at both train and test time.  All other architectural and optimization settings (backbone, scheduler, SDE sampler) remain identical to GenCFD, allowing the FNO to provide fast low‐frequency priors while the diffusion stage refines fine‐scale turbulent features. 

\textbf{Custom implementation.}This schematic closely parallels the pipeline of \cite{oommen2024integrating}, but with two key distinctions. First, we fully replace both the diffusion backbone and its denoising stages with GenCFD. Second, and most important, we condition not only on the FNO’s coarse solution but also on the original high-resolution initial condition. This ensures that our model retains fine-scale structures that would otherwise be lost if we relied solely on the FNO predictions.

\subsection*{FNO}
Our FNO baseline follows Li \emph{et al.}~\cite{FNO} and is implemented in PyTorch.  First, each spatial location of the input field \(u_0\) is lifted from \(C_{\rm in}\) to 256 channels via a two‐layer MLP with a SiLU activation, then projected down to a 128‐channel hidden representation.  This is followed by four \texttt{FnoResBlock}s: in each block, the hidden features undergo a Conditional LayerNorm (time embedding size 128), SiLU, and two spectral convolutions (\texttt{SpectralConv}) with \(num\_modes=(24,24)\), combined through a learnable “soft‐gate” residual skip.  Finally, a two‐layer projection MLP (128 → 256 → \(C_{\rm out}\), with SiLU in between) produces the output field \(\hat u_1\).  We train for 100k iterations minimizing \(\|u_1 - \hat u_1\|_2^2\) with Adam (learning rate \(10^{-3}\), batch size 16), and perform inference in one forward pass—no diffusion or autoregression is used.

\subsection*{UViT}
Our UViT surrogate~\cite{saharia2022photorealistic} uses the identical UNet2D backbone of Table~\ref{tab:rf_params}, injecting time embeddings and employing 8‐head multi‐head attention at the bottleneck.  It is trained deterministically with MSE loss on \((u_0,u_1)\), using Adam at \(3\times10^{-4}\), batch size 16, for 100 k iterations (EMA 0.9999).  For multi‐step forecasting we roll out autoregressively, feeding each prediction back as the next input.  This ablates out the diffusion component while matching all other design choices.

\subsection*{Baseline Model Summary and Implementation Setup}

Table~\ref{tab:model_specs_summary} summarizes key architectural settings for all baseline models considered in this work. These configurations were selected to reflect the strongest performance achievable within a shared training budget of up to 120{,}000 iterations, as used in the quantitative comparison reported in Table ~\ref{tab:results}.
\paragraph{Padding and boundary conditions.} Each dataset required padding choices tailored to its physical boundary conditions. SL and RM, which exhibit periodic boundaries, use circular padding in all convolutional layers. This significantly improved the performance of the diffusion-based models. In contrast, CS features non-periodic boundaries and is thus trained with standard zero padding. These padding choices apply to all UNet-based architectures except ReFlow, which uniformly uses zero padding across all datasets regardless of boundary conditions.

\begin{table}[h]
    \centering
        \caption{Summary of architectural depth and approximated parameter count for each model evaluated. All UNet-based models (ReFlow, GenCFD, GenCFD $\circ$ FNO, and UViT) share a consistent backbone design to ensure consistency across diffusion and non-diffusion baselines, while the FNO baseline uses a structurally different architecture. These configurations reflect the best-performing settings under a training budget of up to 120{,}000 iterations, as used in the statistical comparisons reported in Table~\ref{tab:results}.}
        \label{tab:model_specs_summary}
        \begin{tabular}{l c c c c c}
        \toprule
        \textbf{Model} & \textbf{\# Levels} & \textbf{\# Blocks} & \textbf{\# Params} & \textbf{Noise Schedule} & \textbf{Diffusion Scheme} \\
        \midrule\midrule
        \multirow{2}{*}{ReFlow} & \multirow{2}{*}{3} & \multirow{2}{*}{4} & \multirow{2}{*}{22M} & \multirow{2}{*}{uniform} & rectified continuous \\
         &  &  &  &  & time variable $t \in [0,1]$ \\
        \midrule
        GenCFD & 2 & 4 & 5M & exponential & variance-exploding \\
        \midrule
        GenCFD $\circ$ FNO & 2 & 4 & 5M & exponential & variance-exploding \\
        \midrule
        UViT & 2 & 4 & 5M & \textemdash & \textemdash \\
        \midrule
        FNO & \textemdash & 4 & 11M & \textemdash & \textemdash \\
        \bottomrule
    \end{tabular}
\end{table}

\section{Results} \label{app:results}
\subsection{Performance Metrics}

To compare our generated ensembles against the reference Monte Carlo samples, we employ three complementary measures:

\begin{itemize}
  \item \textbf{Mean‐field error.}  Let $\mu_{\rm ref}(x)$ and $\mu_{\rm model}(x)$ denote the pointwise spatial means of the reference and model ensembles, respectively.  We measure their discrepancy by the $L^2$‐norm
  \[
    e_{\mu}
    \;=\;
    \bigl\|\,
      \mu_{\rm ref} \;-\;\mu_{\rm model}
    \bigr\|_{2}\,.
  \]

  \item \textbf{Standard‐deviation error.}  Similarly, let $\sigma_{\rm ref}(x)$ and $\sigma_{\rm model}(x)$ be the pointwise standard deviations.  We therefore compute
  \[
    e_{\sigma}
    \;=\;
    \bigl\|\,
      \sigma_{\rm ref} \;-\;\sigma_{\rm model}
    \bigr\|_{2}\,,
  \]

Both metrics are reported after normalization by the $L^2$ norm of the ground truth.

 \item \textbf{Average 1-Wasserstein distance.}  
  For a fixed initial condition \(u_0\), let
  \(\{u^{(j)}_{\rm ref}(x)\}_{j=1}^N\) and
  \(\{u^{(j)}_{\rm model}(x)\}_{j=1}^N\)
  be the ensembles of final‐time values produced by the reference solver and our model, respectively.  Since these fields are discretized on \(M\) spatial points \(\{x_i\}_{i=1}^M\), we obtain two empirical 1D distributions at each \(x_i\).  Denoting their inverse CDFs by
  \(F^{-1}_{\rm ref}(\cdot\,;\,x_i)\) and \(F^{-1}_{\rm model}(\cdot\,;\,x_i)\), the pointwise Wasserstein‐1 distance is
  \[
    W_1\bigl(p_{\rm ref}(\cdot\mid x_i),\,p_{\rm model}(\cdot\mid x_i)\bigr)
    \;=\;
    \int_{0}^{1}
      \bigl|\,F^{-1}_{\rm ref}(q; x_i)\;-\;F^{-1}_{\rm model}(q; x_i)\bigr|
    \,\mathrm{d}q.
  \]
  We then report the spatial average
  \[
    \overline W_1
    \;=\;
    \frac{1}{M}
    \sum_{i=1}^{M}
    W_1\bigl(p_{\rm ref}(\cdot\mid x_i),\,p_{\rm model}(\cdot\mid x_i)\bigr).
  \]

\end{itemize}

\medskip
\noindent\textbf{Energy spectra.}  Finally, to assess how well different scales are captured, we compute the discrete energy spectrum of each velocity field.  Given a sample $u(x)$ on a $d$‐dimensional periodic grid with spacing $\Delta$, let $\widehat u_k$ be its Fourier coefficient at integer wavevector $k\in\mathbb Z^d$.  We bin modes by their $\ell_1$ radius $|k|_1=k_1+\dots+k_d$, defining
\[
  E_{r}
  \;=\;
  \frac{\Delta^d}{2}\,
  \sum_{\substack{k\in\mathbb Z^d\\|k|_1=r}}
    \bigl\|\widehat u_{k}\bigr\|^2,
  \quad
  r = 0,1,2,\dots
\]
and compare the ensemble‐average spectra of reference and model.  Our plots show $E_r$ versus $r$ on log–log axes to highlight scale‐by‐scale agreement.

\subsection{Training Efficiency Observations} \label{app:training_efficiency}
\begin{figure}[!htbp]
  \centering
  \includegraphics[width=\textwidth]{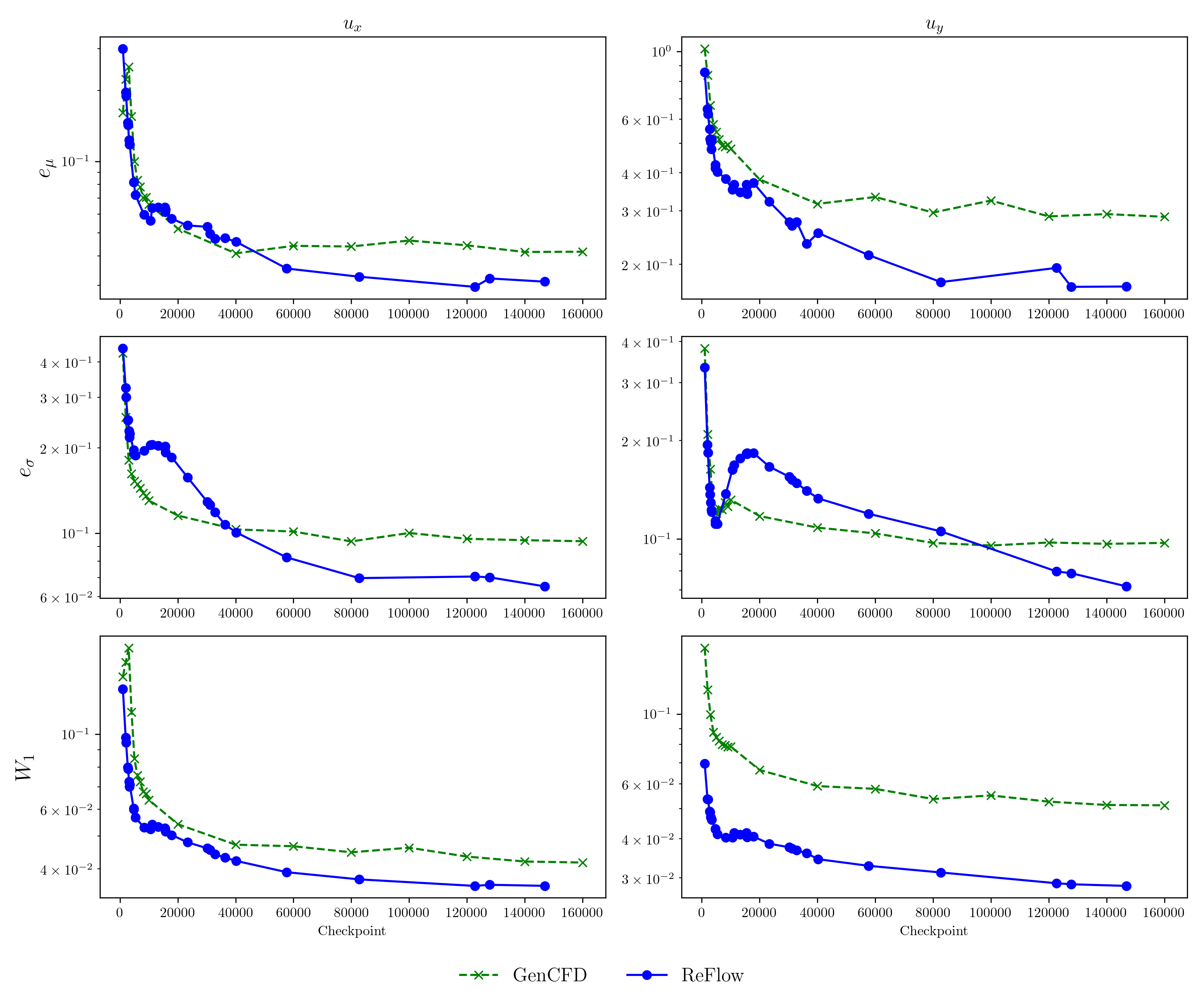}
  \caption{\textbf{Training efficiency comparison.} Performance of ReFlow and GenCFD on the Macro-Micro SL ensemble dataset evaluated at multiple training checkpoints. Metrics match those used in the main text: relative $L^2$ error of the mean ($e_\mu$) and standard deviation ($e_\sigma$), and the average 1-point Wasserstein-1 distance (W1), computed for both velocity components ($u_x$, $u_y$). The $y$-axis uses a logarithmic scale, while the $x$-axis (training iteration) is linear.}
  \label{fig:training_efficiency}
\end{figure}

To assess training efficiency, we compare ReFlow with GenCFD. For fairness, we use a GenCFD variant with the same setup and parameter count (22M), matching the three-level UViT architecture of ReFlow instead of its standard lightweight configuration (5M). Both models were trained for up to 160{,}000 iterations with a batch size of 16.
\newline
Figure~\ref{fig:training_efficiency} summarizes performance over the training trajectory. At approximately 100{,}000 steps, ReFlow consistently outperforms GenCFD across all key metrics: relative $L^2$ error in the mean ($e_\mu$), relative $L^2$ error in the standard deviation ($e_\sigma$), and the average pointwise Wasserstein-1 (W1) distance—evaluated for both $u_x$ and $u_y$ velocity components. Notably, the $y$-axis uses a logarithmic scale to emphasize differences in error magnitude across training steps.
\newline
Across all metrics, ReFlow begins to outperform GenCFD at around 100{,}000 training steps and continues to improve at a sharper rate. While neither model reaches full convergence within 160{,}000 iterations, ReFlow consistently exhibits lower errors and more favorable metric trajectories in later stages of training. These results support the observation that ReFlow is more training-efficient within a fixed compute budget and ultimately achieves stronger generalization performance in this setting.
\newline
To further probe the long-term capabilities of GenCFD, we extended its training to 600{,}000 iterations and checked whether it eventually surpasses ReFlow's performance at its final recorded checkpoint (150{,}000 steps). Table~\ref{tab:training_efficiency} summarizes the earliest training steps at which GenCFD overtakes ReFlow for each evaluation metric. While GenCFD does catch up and exceed ReFlow in several cases, it never surpasses ReFlow on the Wasserstein-1 distance for $u_y$ within the training horizon considered.

\begin{table}[H]
\centering
\caption{Number of training iterations at which GenCFD surpasses ReFlow’s performance at 150{,}000 iterations. If not surpassed, indicated by “—”.}
\label{tab:training_efficiency}
\begin{tabular}{lccc}
\toprule
\textbf{Metric} & \textbf{GenCFD Iteration} & \textbf{GenCFD Value} & \textbf{ReFlow Value @ 150{,}000 iter.} \\
\midrule
$u_x$ Mean Error ($e_\mu$) & 280{,}000 & 3.33\,$\times10^{-2}$ & 3.45\,$\times10^{-2}$ \\
$u_y$ Mean Error ($e_\mu$) & 400{,}000 & 1.70\,$\times10^{-1}$ & 1.88\,$\times10^{-1}$ \\
$u_x$ Std Error ($e_\sigma$)  & 400{,}000 & 6.48\,$\times10^{-2}$ & 7.24\,$\times10^{-2}$ \\
$u_y$ Std Error ($e_\sigma$)  & 340{,}000 & 7.91\,$\times10^{-2}$ & 7.97\,$\times10^{-2}$ \\
$u_x$ Wasserstein-1 ($W_1$)  & 200{,}000 & 3.89\,$\times10^{-2}$ & 3.95\,$\times10^{-2}$ \\
$u_y$ Wasserstein-1 ($W_1$)  & —        & —                         & 4.11\,$\times10^{-2}$ \\
\bottomrule
\end{tabular}
\end{table}

\subsection{Experimental Advantages of Straightness} \label{app:straightness_advantages}
This subsection presents empirical evidence that the “rectified” trajectories generated by our Rectified Flow model (ReFlow) enable faster convergence and yield higher-fidelity reconstructions compared to the baseline GenCFD diffusion approach.

\paragraph{Evaluation settings.}
We use the CloudShock dataset throughout for exposition purposes. However, we stress that the same results hold across datasets.  For each initial condition, both ReFlow and GenCFD are allotted \(T=10\) diffusion steps over normalized \textbf{diffusion time} \(\tau\in[0,1]\).  We record the model outputs at five evenly spaced timesteps \(t=\{0.0,0.25,0.50,0.75,1.0\}\) and analyze:
\begin{itemize}
  \item \emph{Trajectory inpainting} of the density field \(\rho\).
  \item \emph{Latent‐Space Trajectory Visualization via PCA} via PCA.
  \item \emph{Per‐Sample Average Error Evolution}: average MSE across physical channels.
  \item \emph{Evolution of the Energy Spectrum}: radial power‐spectrum of energy \(E\) as inpaintig progresses.
\end{itemize}
All other components (backbone, noise schedule, batch size) are held equal for fair comparison.

\subsection*{Trajectory Inpainting}
\label{sec:traj_inpainting}
We compare Rectified Flow (ReFlow) and GenCFD reconstructions of the density field \(\rho\) over five normalized \textbf{diffusion} timesteps \(t\in\{0.00,0.25,0.50,0.75,1.00\}\). ReFlow inpaints fine-scale features in far fewer steps, while GenCFD remains overly diffusive. The leftmost panel corresponds to the initial condition $u_0$ upon which both model are conditioned, while the following panels showcase the evolution of the model's output as diffusion time progresses. 

\FloatBarrier
\begin{figure}[!htbp]
  \centering
  \includegraphics[width=0.95\linewidth]{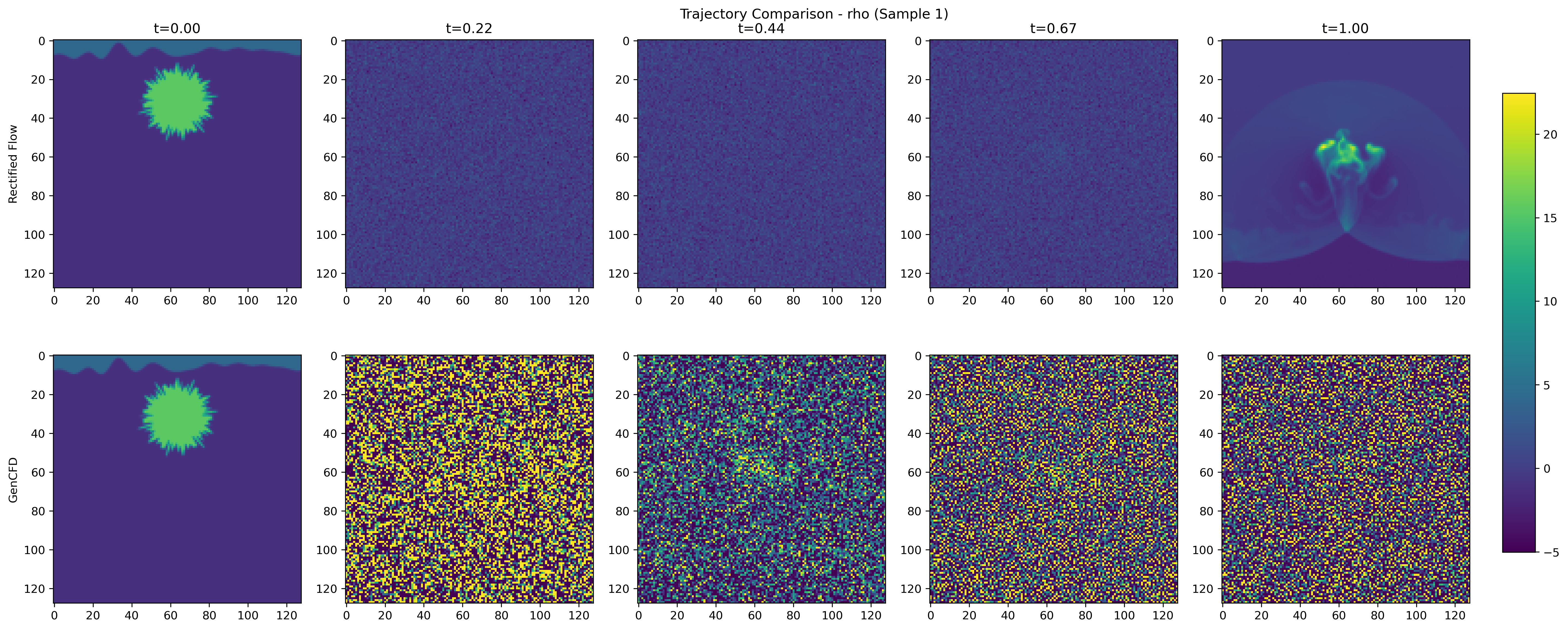}
  \caption{Density trajectories for Sample 1: ReFlow (top) vs.\ GenCFD (bottom). ReFlow already recovers sharp shock fronts by diffusion time \(t=1.0\).}
  \label{fig:traj_rho_s1}
\end{figure}

\begin{figure}[!htbp]
  \centering
  \includegraphics[width=0.95\linewidth]{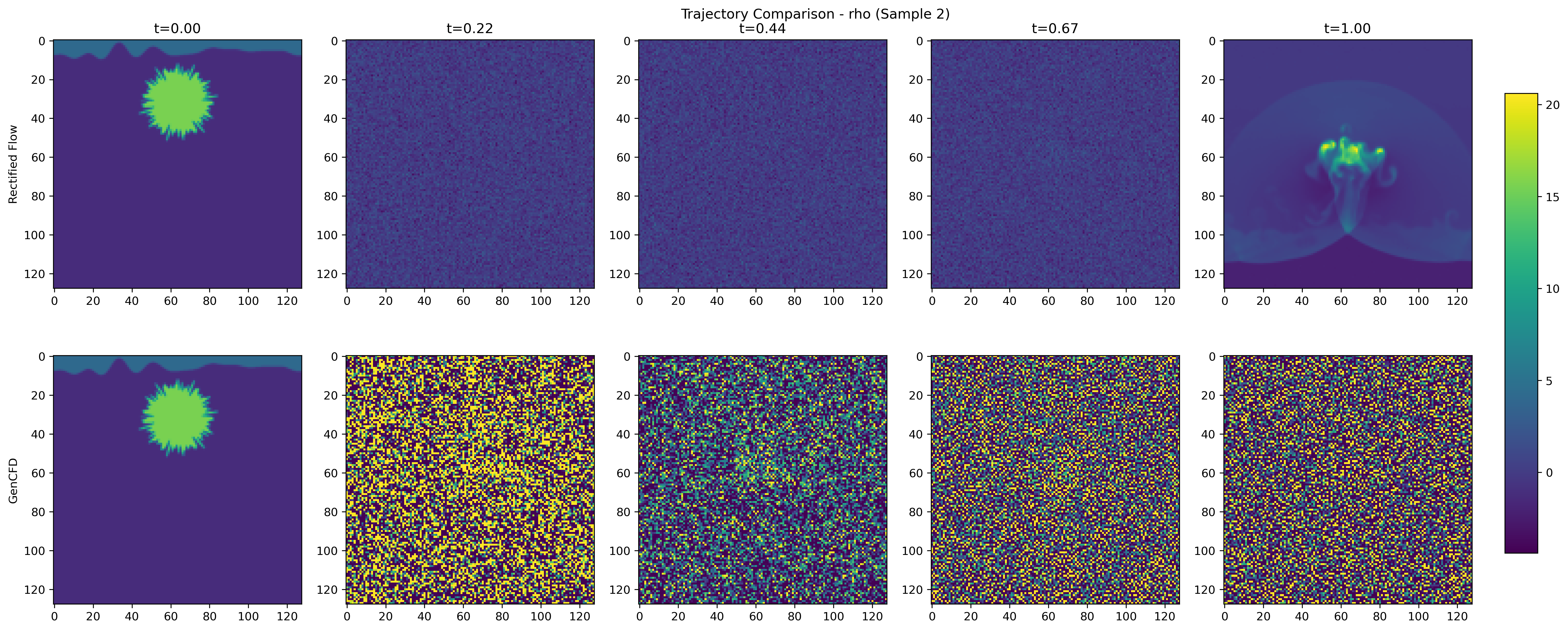}
  \caption{Density trajectories for Sample 2.}
  \label{fig:traj_rho_s2}
\end{figure}

\begin{figure}[!htbp]
  \centering
  \includegraphics[width=0.95\linewidth]{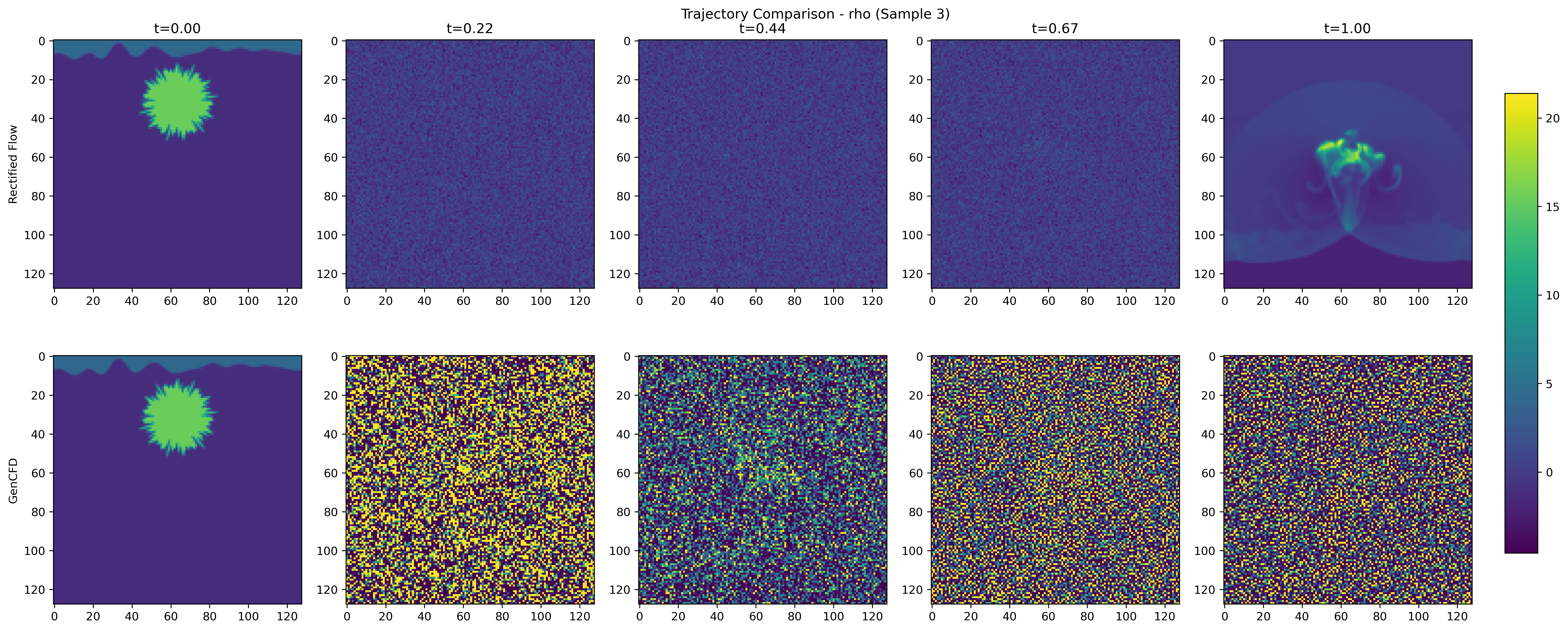}
  \caption{Density trajectories for Sample 3.}
  \label{fig:traj_rho_s3}
\end{figure}
\FloatBarrier

\subsection*{Latent‐Space Trajectory Visualization via PCA}
To quantify how directly each model moves through the data manifold, we flatten the \((H\times W\times C)\) field at each timestep into a single feature vector, stack all Rectified Flow and GenCFD vectors, fit a 3-component PCA, and then project each timestep into PC1/PC2/PC3. A straighter path indicates fewer diffusive detours.

\begin{figure}[!htbp]
  \centering
  \captionsetup[subfigure]{justification=centering,font=small}

  \begin{subfigure}[b]{0.32\columnwidth}
    \includegraphics[width=\linewidth]{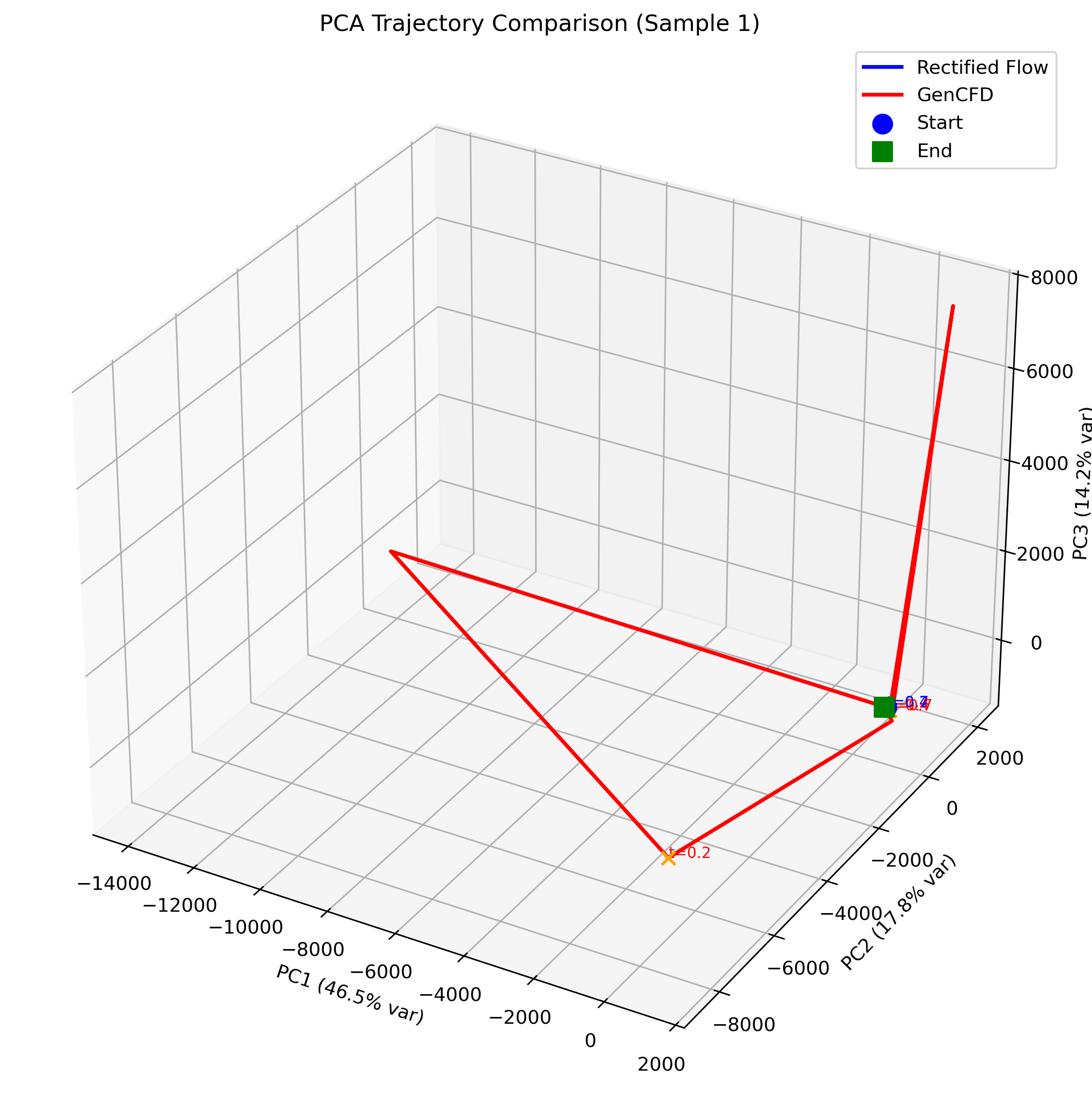}
    \caption{Sample 1}
    \label{fig:pca3d_s1}
  \end{subfigure}\hfill
  \begin{subfigure}[b]{0.32\columnwidth}
    \includegraphics[width=\linewidth]{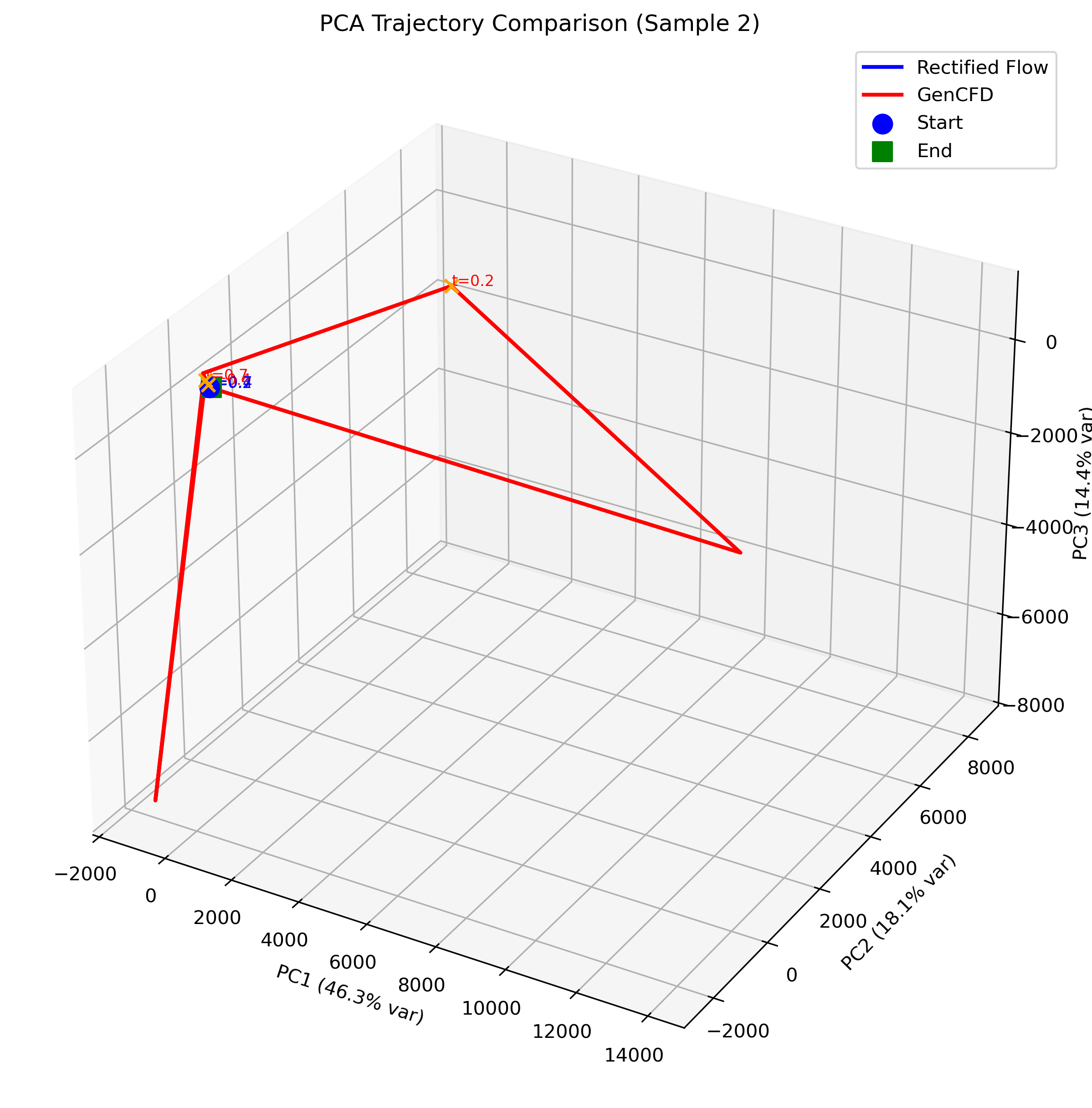}
    \caption{Sample 2}
    \label{fig:pca3d_s2}
  \end{subfigure}\hfill
  \begin{subfigure}[b]{0.32\columnwidth}
    \includegraphics[width=\linewidth]{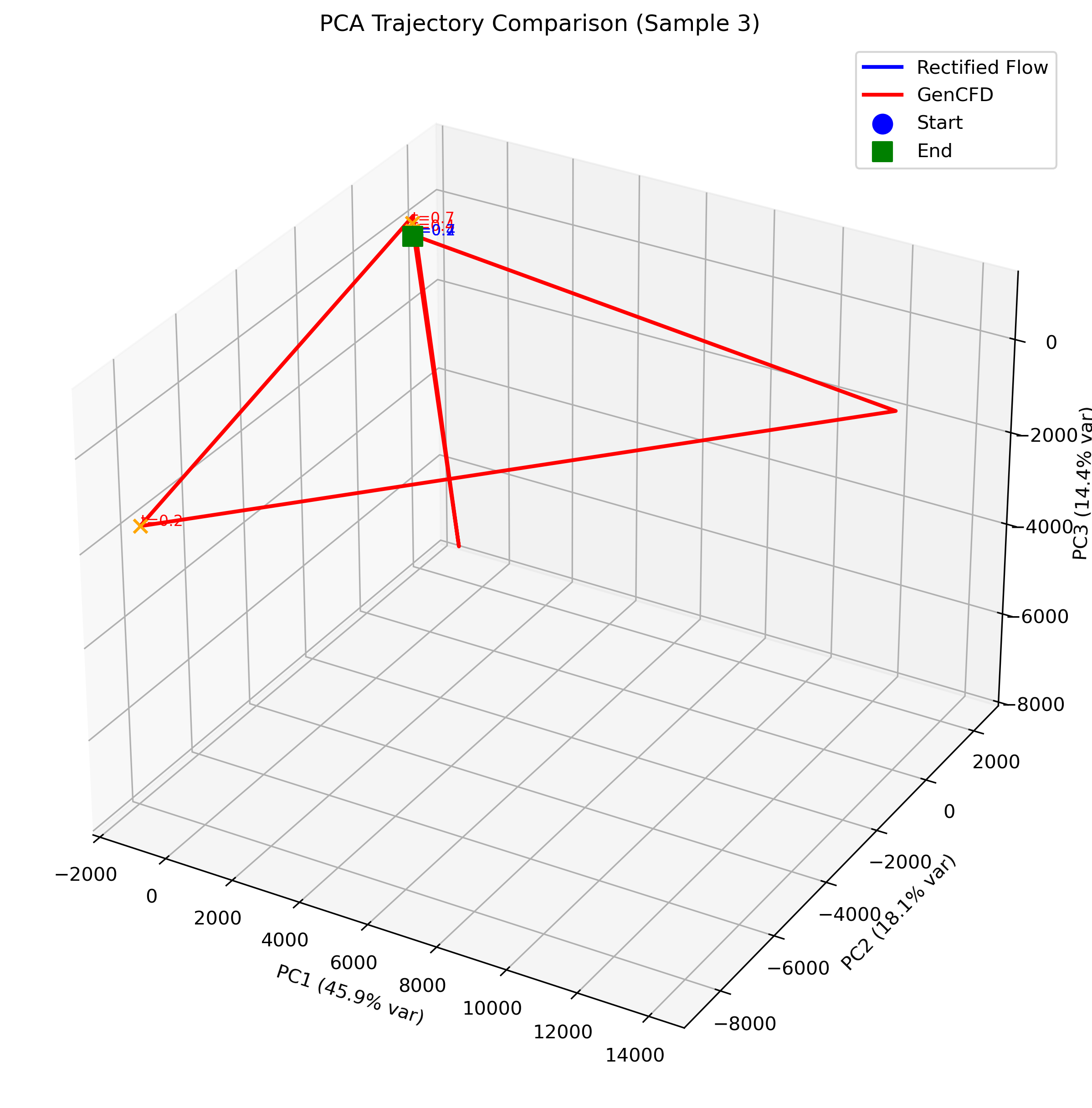}
    \caption{Sample 3}
    \label{fig:pca3d_s3}
  \end{subfigure}

  \vspace{0.8em}

  \begin{subfigure}[b]{0.32\columnwidth}
    \includegraphics[width=\linewidth]{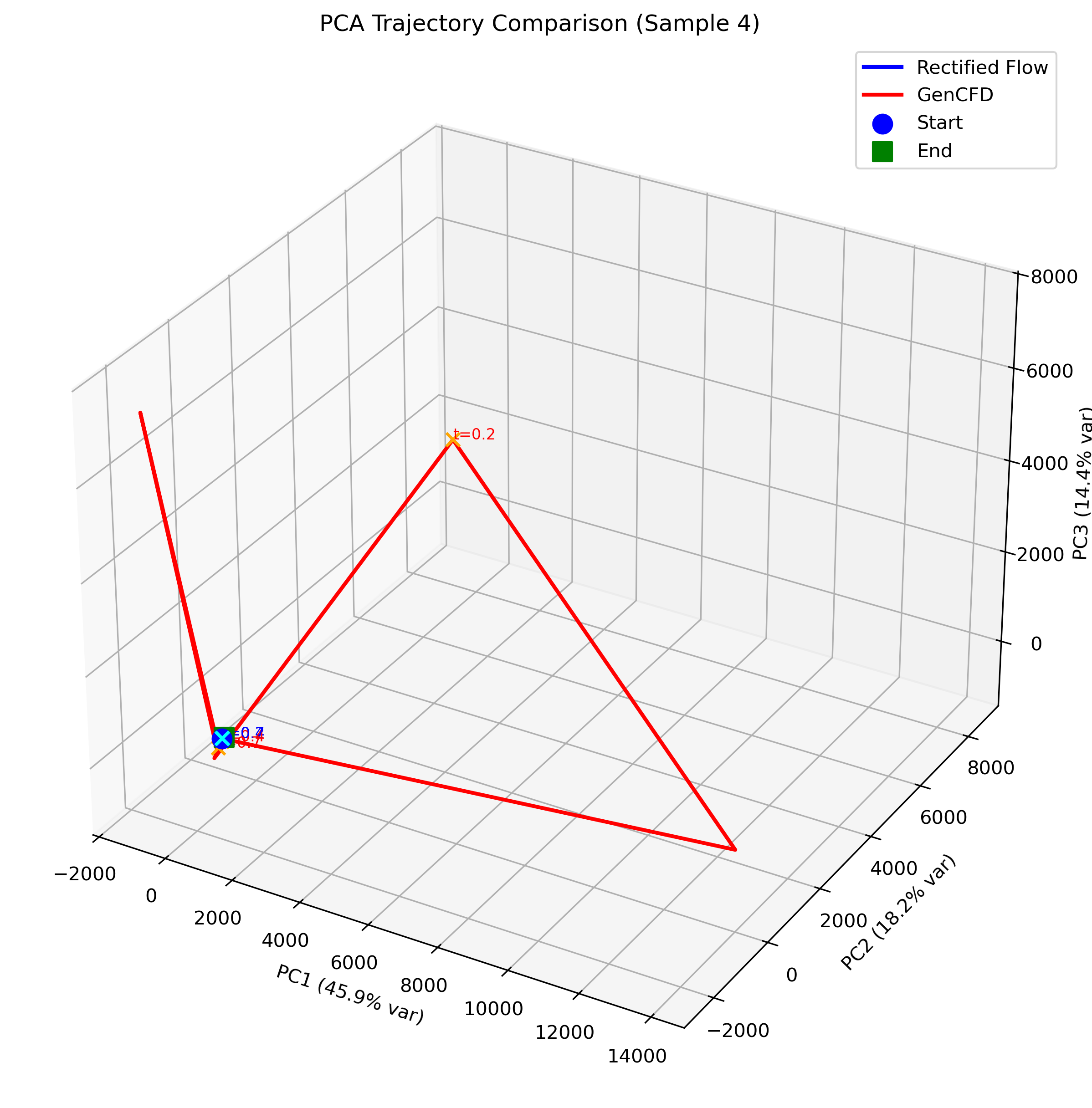}
    \caption{Sample 4}
    \label{fig:pca3d_s4}
  \end{subfigure}\hfill
  \begin{subfigure}[b]{0.32\columnwidth}
    \includegraphics[width=\linewidth]{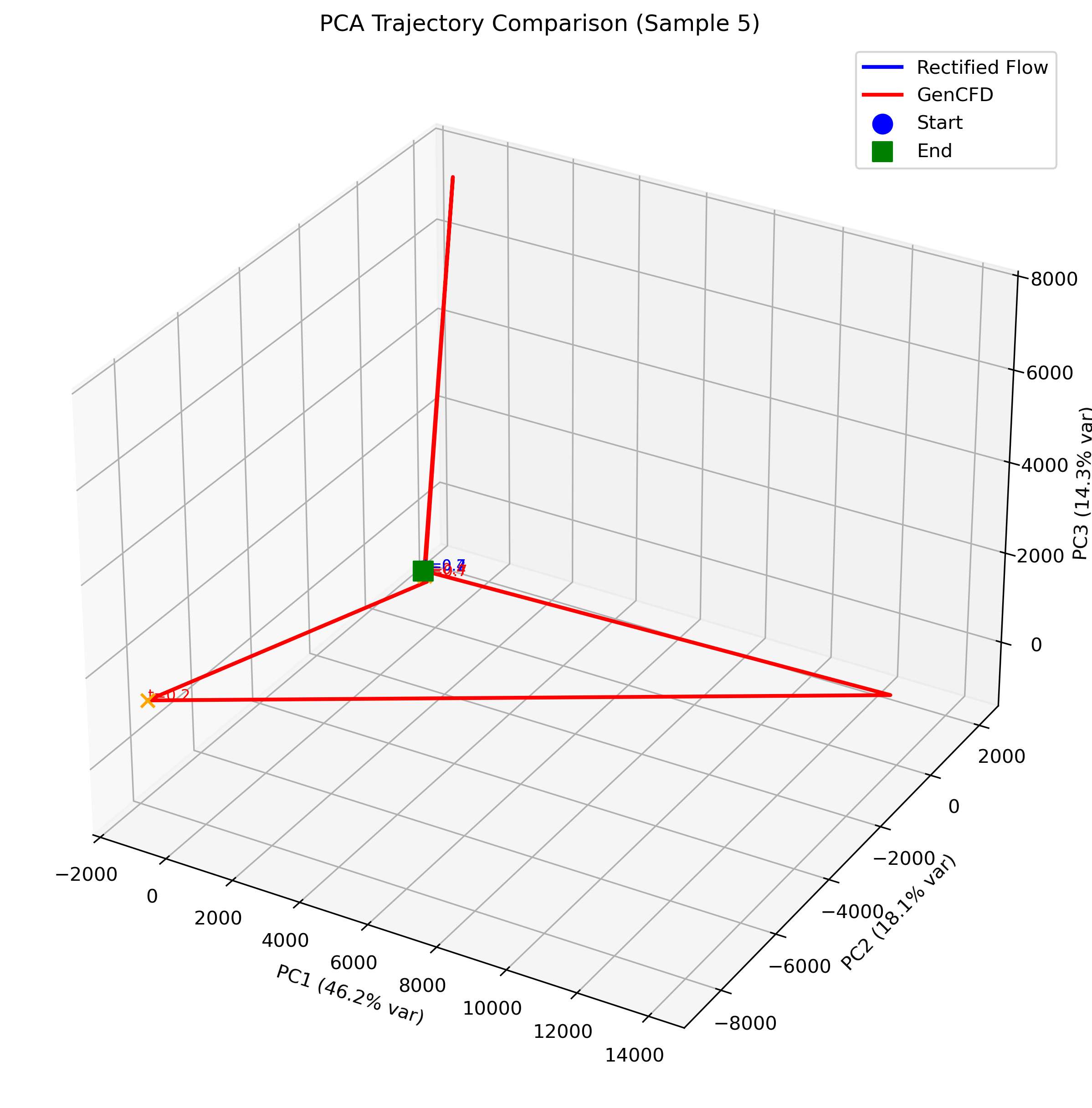}
    \caption{Sample 5}
    \label{fig:pca3d_s5}
  \end{subfigure}\hfill
  \begin{subfigure}[b]{0.32\columnwidth}
    \includegraphics[width=\linewidth]{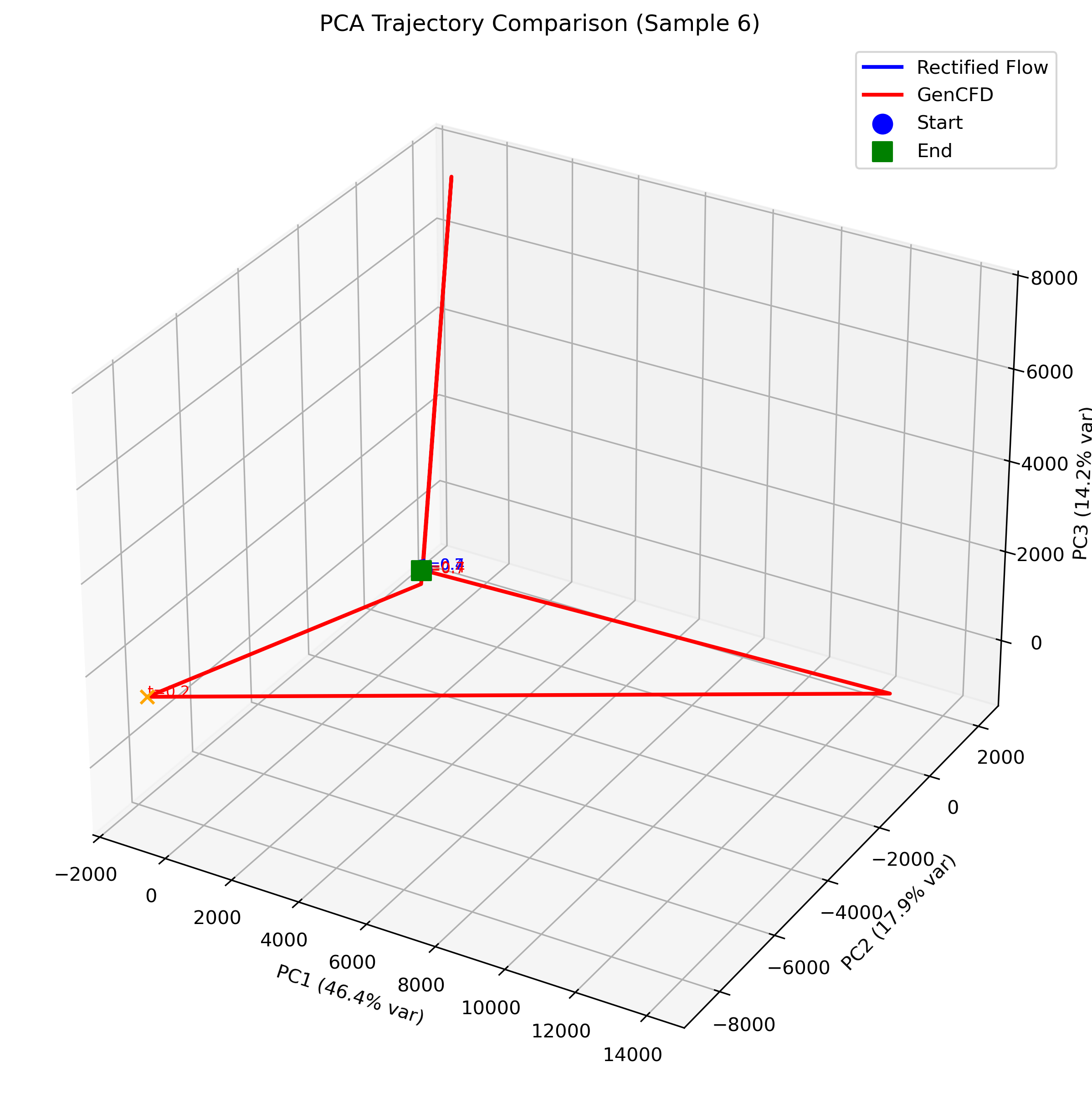}
    \caption{Sample 6}
    \label{fig:pca3d_s6}
  \end{subfigure}

  \caption{%
    \textbf{3D PCA latent-space trajectories (Samples 1–6).}
    We project the flattened snapshots at each of \(T\) timesteps into the top three PCA components. Rectified Flow (blue) consistently traces a straighter path than GenCFD (red), confirming its “rectified” evolution. Circles mark \(t=0\), squares mark \(t=1\), and intermediate timesteps are annotated.}
  \label{fig:pca3d_grid_6}
\end{figure}

\begin{figure}[t]
  \centering
  \begin{minipage}{\textwidth}
    \centering
    \includegraphics[width=\textwidth]{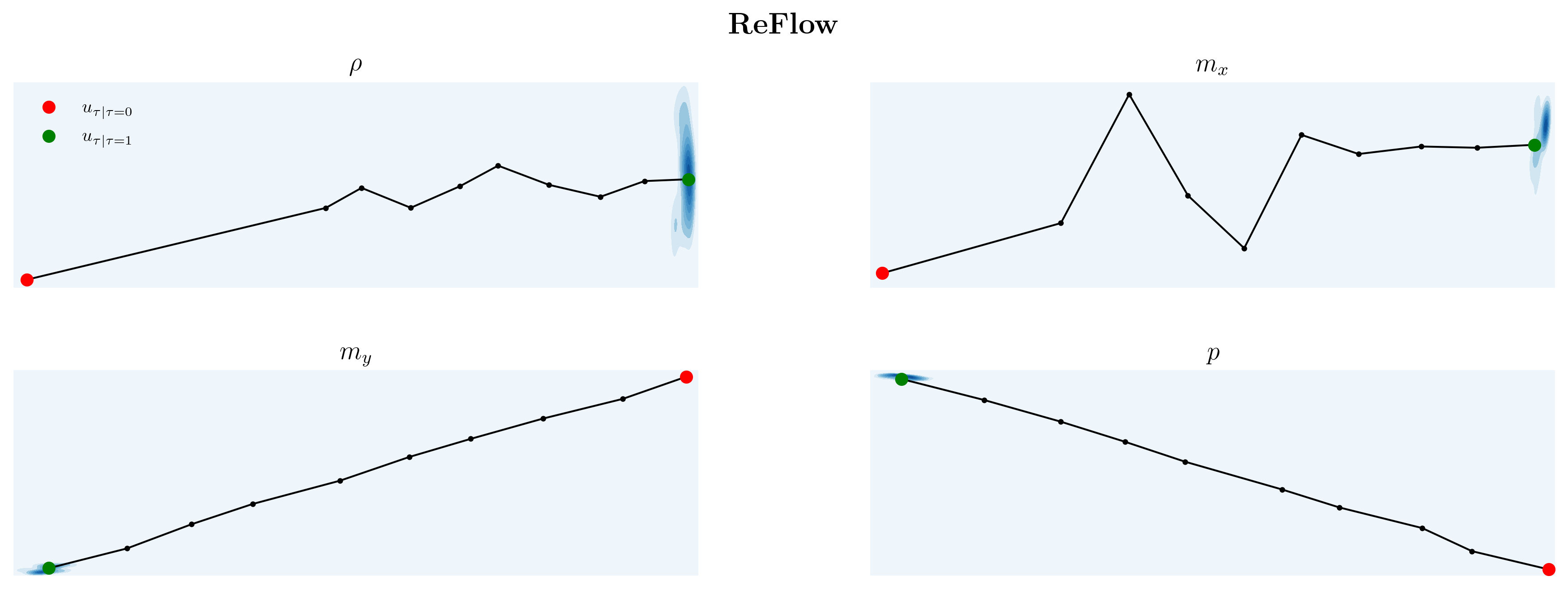}
  \end{minipage}
  
  \vspace{2em}

  \begin{minipage}{\textwidth}
    \centering
    \includegraphics[width=\textwidth]{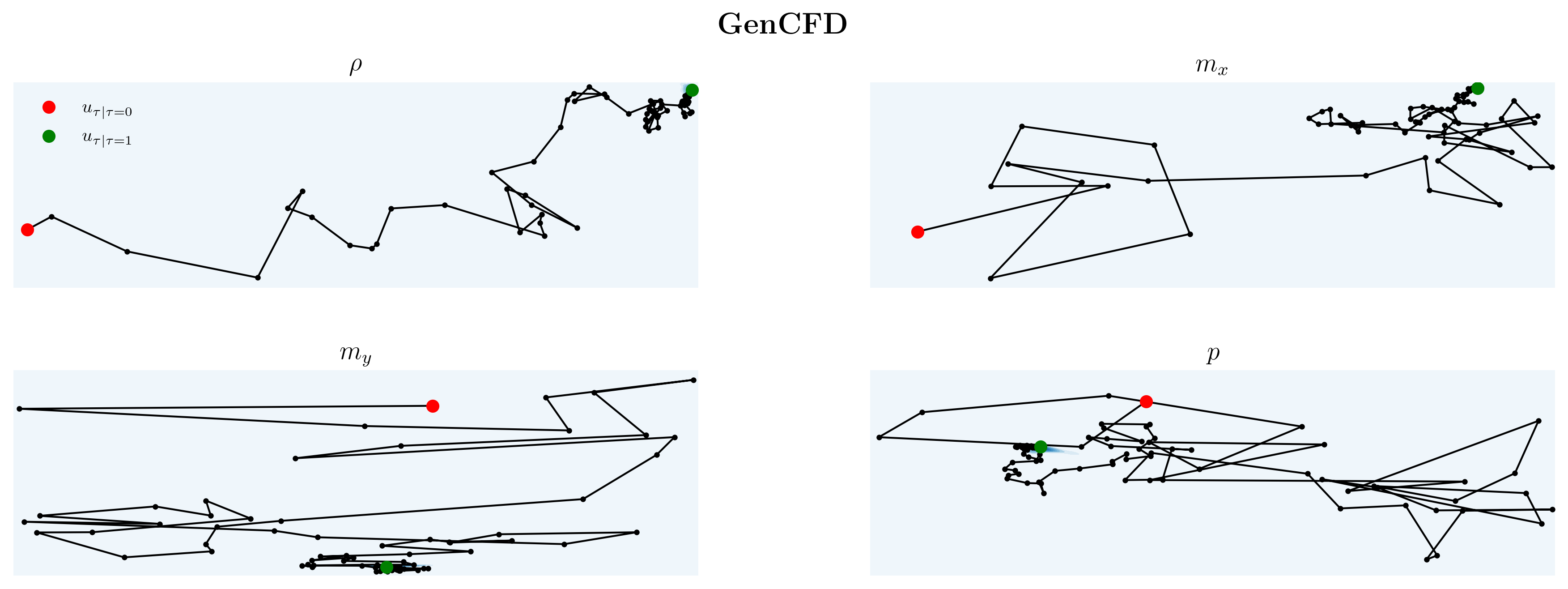}
  \end{minipage}
  
  \caption{\textbf{2D PCA latent-space trajectories.} We project the flattened velocity snapshots from the RM dataset into a 2D PCA space, computed independently for each channel ($\rho$, $m_x$, $m_y$, $p$) across all $T$ time steps. To visualize the underlying distribution of target fields, we apply kernel density estimation (KDE) per channel. The ReFlow model (top) evolves over 10 rectified time steps, producing visibly straighter and more structured trajectories compared to the score-based diffusion model GenCFD (bottom), which uses 128 stochastic steps. Although the projection into two principal components necessarily truncates the high-dimensional dynamics, the qualitative difference remains evident, illustrating the smoother and more directed progression learned by ReFlow.}
  \label{fig:pca2d}
\end{figure}

To further illustrate the nature of each model’s generative dynamics, we visualize single-sample generation trajectories from both models in a 2D PCA space for the RM dataset (Figure~\ref{fig:pca2d}). For each physical channel ($\rho$, $m_x$, $m_y$, $p$), we apply PCA to the full set of ground truth snapshots to define a common projection basis. We then overlay the generative trajectory of a single sample in this space. To contextualize the generated dynamics, we add kernel density estimates (KDEs) showing the empirical data distribution in the same projected space.
\newline
ReFlow (top) again traces a straighter, more compact path over just 10 rectified steps, moving coherently along the data manifold. GenCFD (bottom), by contrast, uses 128 stochastic denoising steps and produces a more diffuse and nonlinear trajectory, often deviating from high-density regions of the ground truth distribution. While the PCA projection inherently reduces dimensional complexity, the consistent difference in shape and alignment supports the conclusion that ReFlow’s rectified flow better preserves manifold structure and sample efficiency.

\newpage
\FloatBarrier
\subsection*{Per‐Sample Average Error Evolution}
\begin{figure}[!htbp]
  \centering
  \captionsetup[subfigure]{justification=centering,font=small}

  \begin{subfigure}[b]{0.45\columnwidth}
    \includegraphics[width=\linewidth]{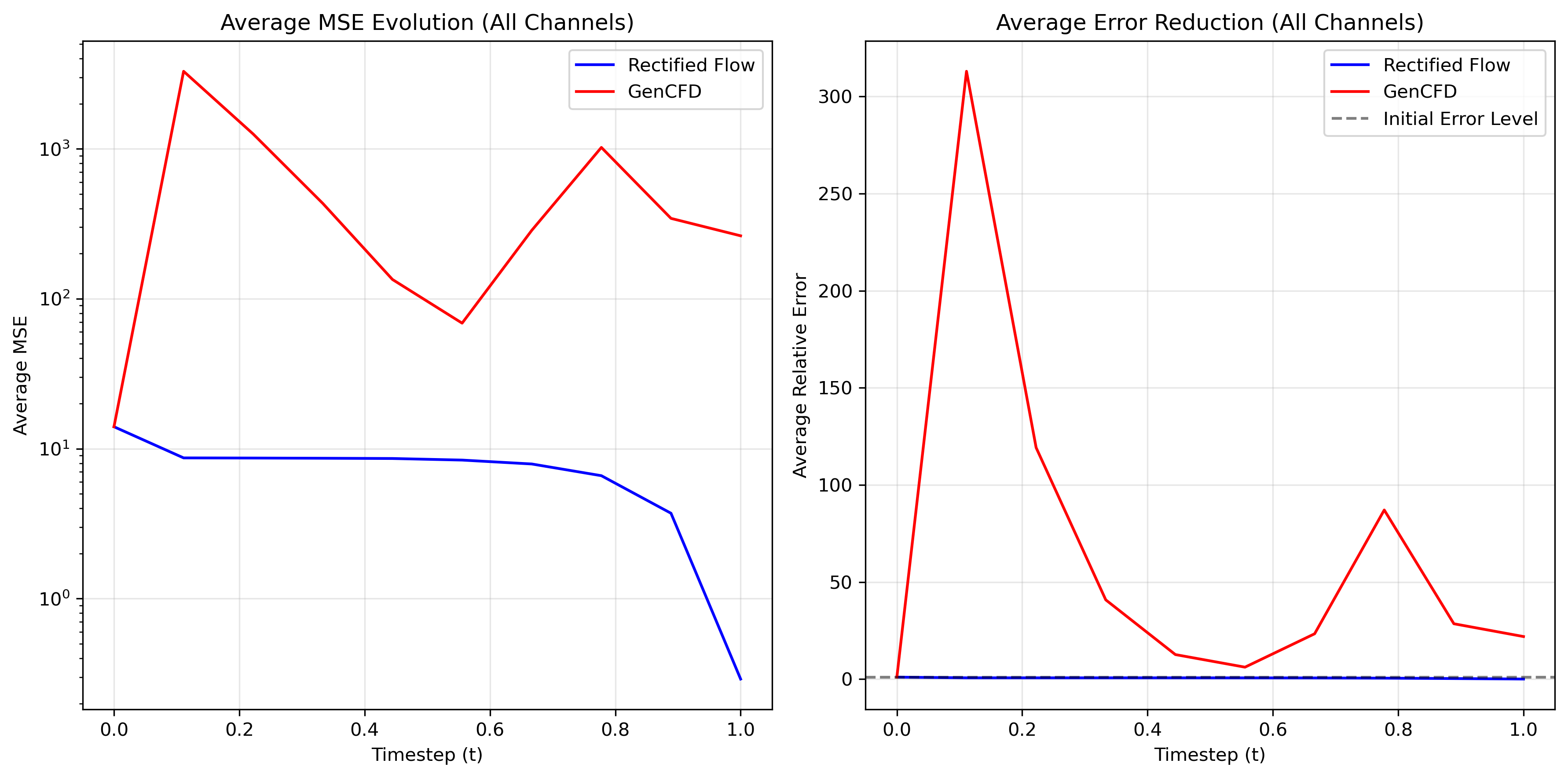}
    \caption{Sample 1}
    \label{fig:err_avg_s1}
  \end{subfigure}\hfill
  \begin{subfigure}[b]{0.45\columnwidth}
    \includegraphics[width=\linewidth]{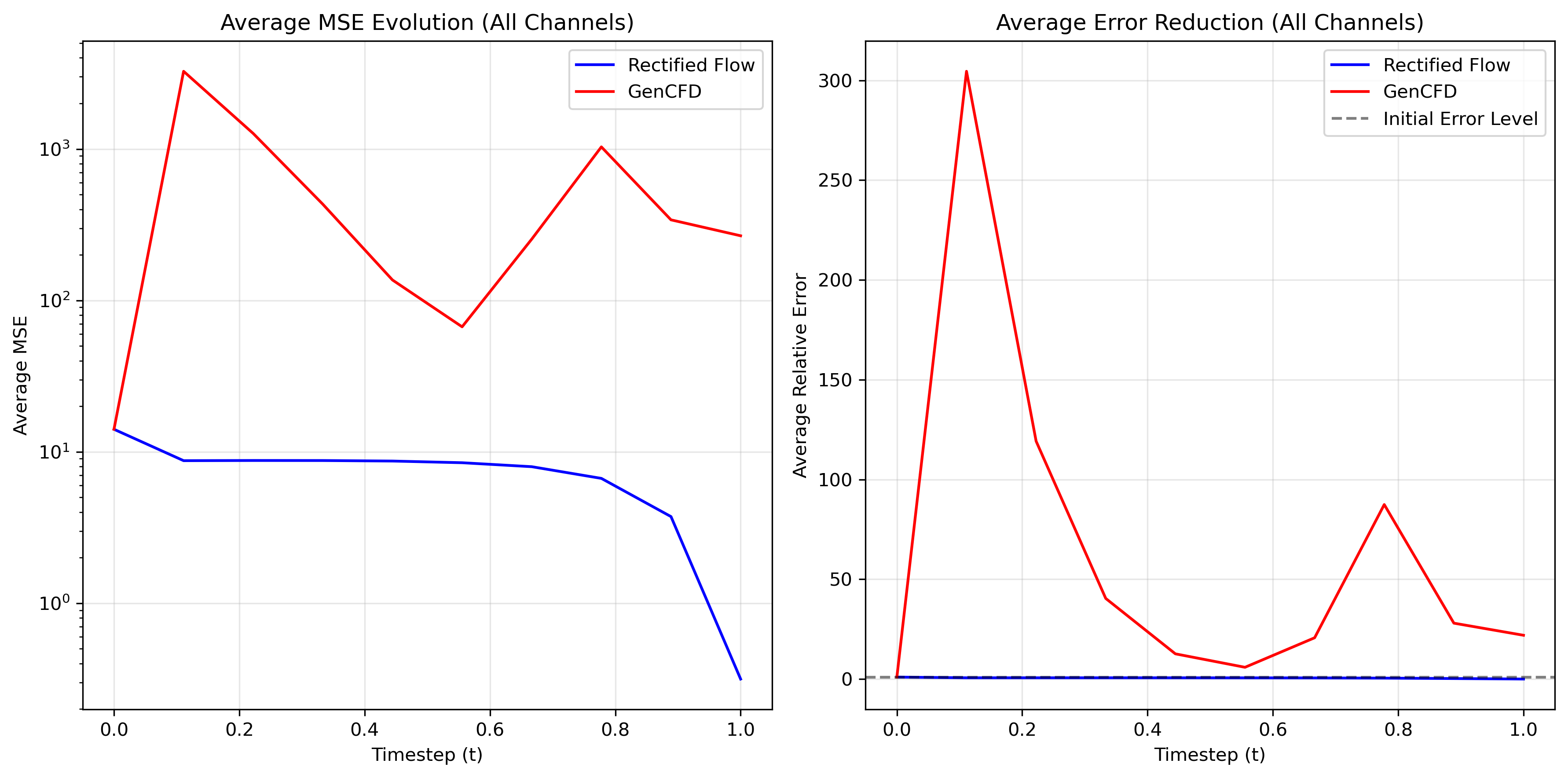}
    \caption{Sample 2}
    \label{fig:err_avg_s2}
  \end{subfigure}

  \vspace{0.8em}

  \begin{subfigure}[b]{0.45\columnwidth}
    \includegraphics[width=\linewidth]{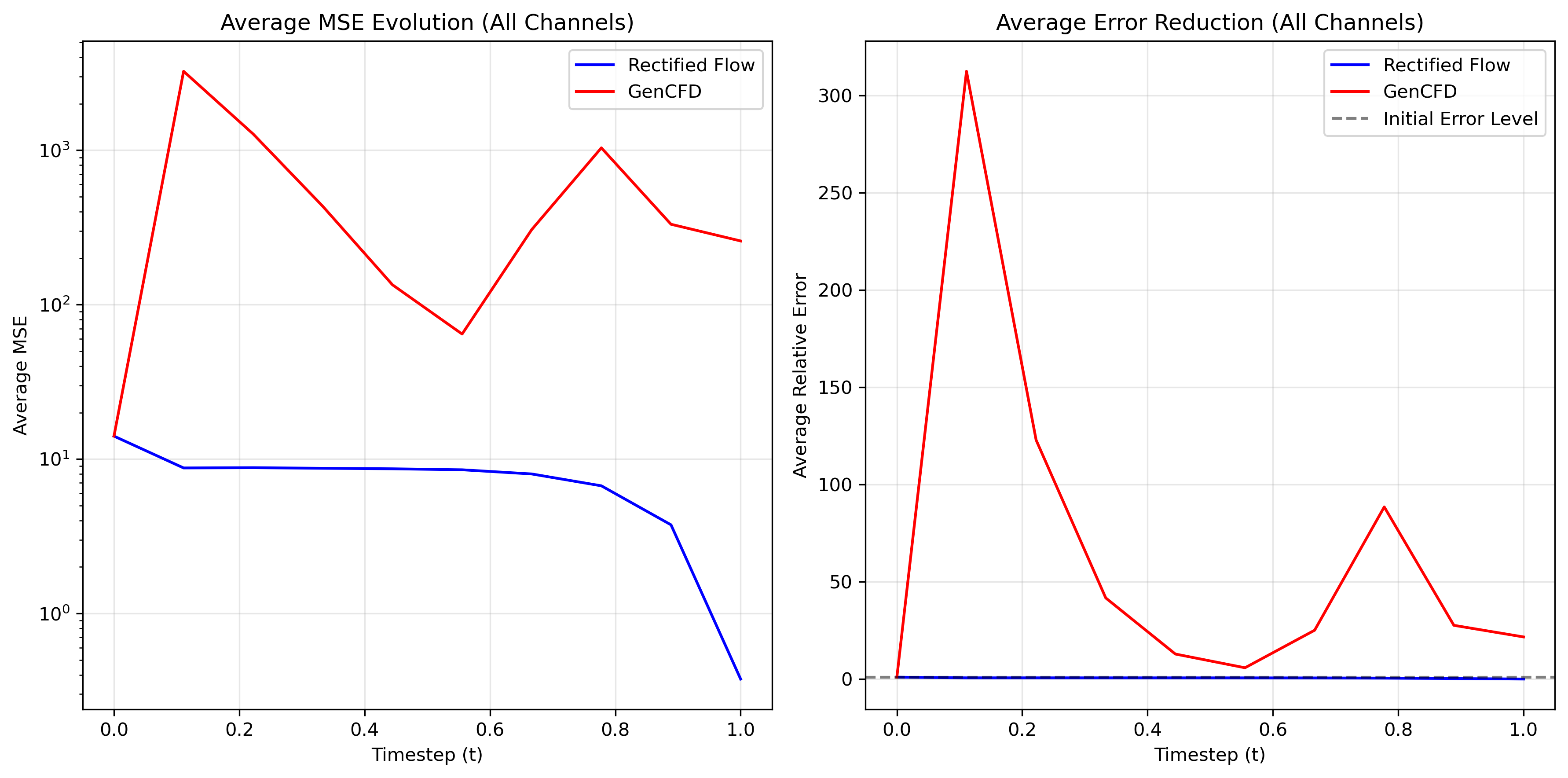}
    \caption{Sample 3}
    \label{fig:err_avg_s3}
  \end{subfigure}\hfill
  \begin{subfigure}[b]{0.45\columnwidth}
    \includegraphics[width=\linewidth]{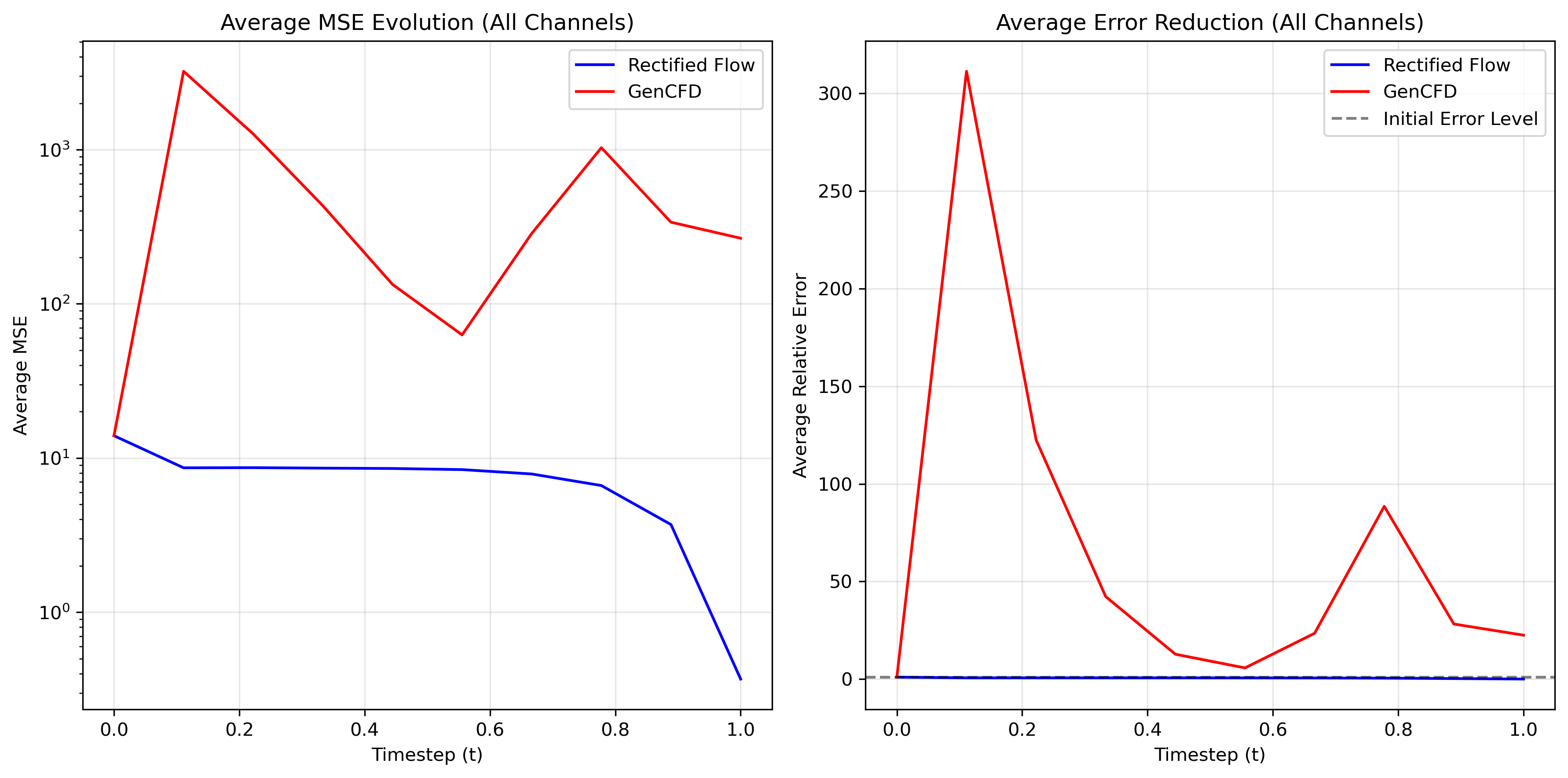}
    \caption{Sample 4}
    \label{fig:err_avg_s4}
  \end{subfigure}

  \vspace{0.8em}

  \begin{subfigure}[b]{0.45\columnwidth}
    \includegraphics[width=\linewidth]{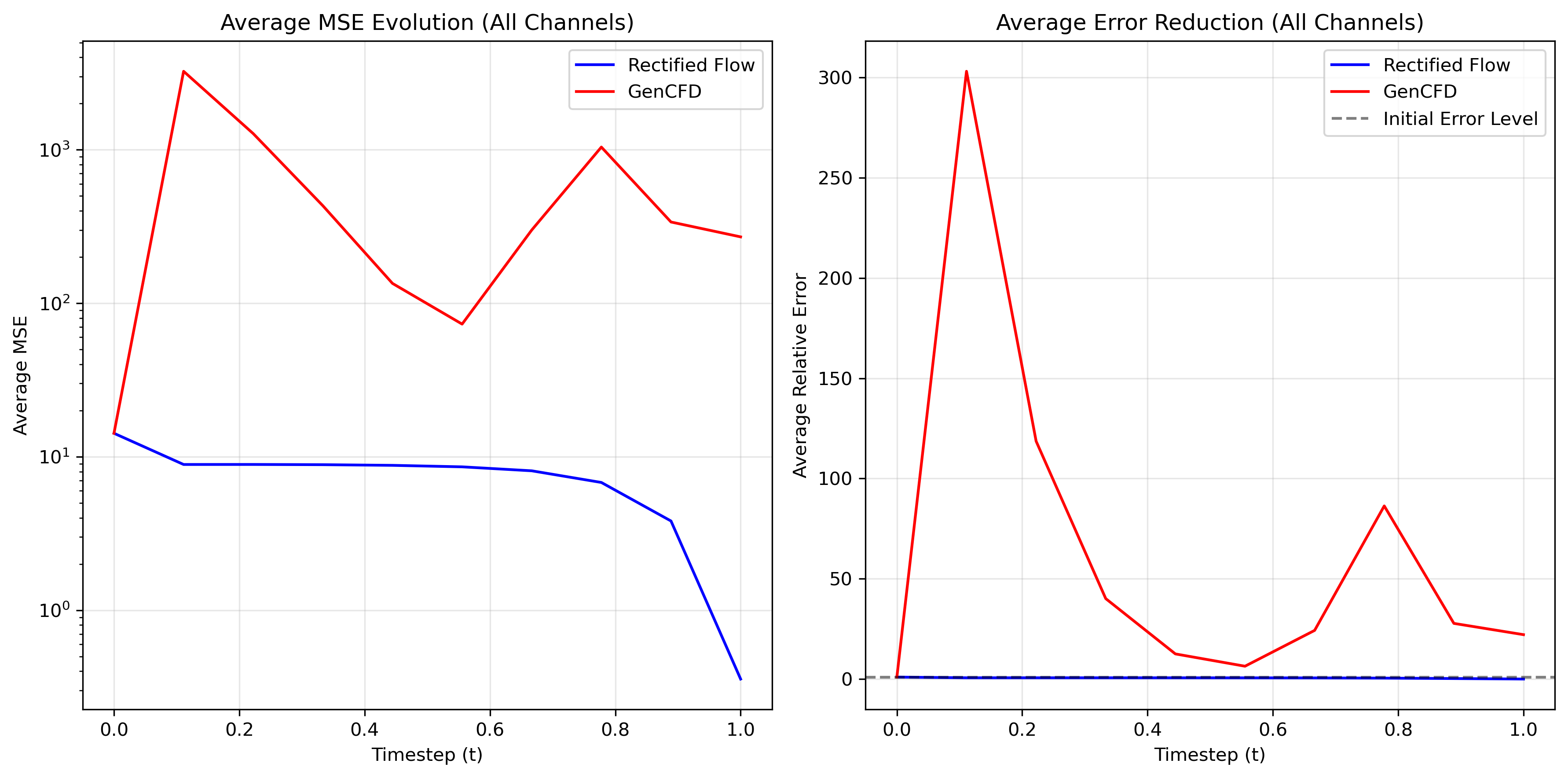}
    \caption{Sample 5}
    \label{fig:err_avg_s5}
  \end{subfigure}\hfill
  \begin{subfigure}[b]{0.45\columnwidth}
    \includegraphics[width=\linewidth]{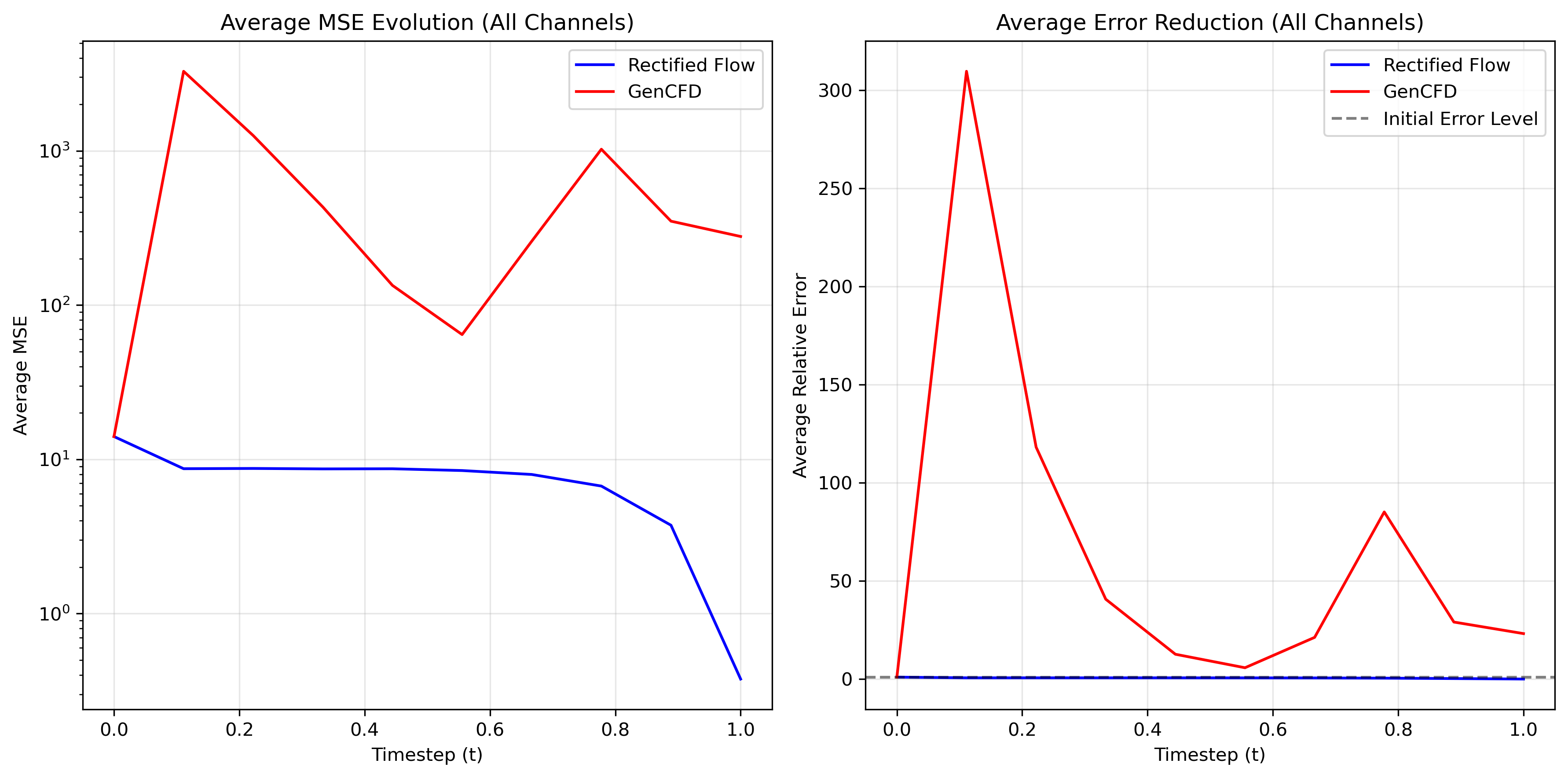}
    \caption{Sample 6}
    \label{fig:err_avg_s6}
  \end{subfigure}

  \caption{%
    \textbf{Per‐sample average MSE evolution.}
    Each subfigure plots, left half: \(\log_{10}\overline{\mathrm{MSE}}(t)\); right half: \(\overline{\mathrm{MSE}}(t)/\overline{\mathrm{MSE}}(0)\).  
    Rectified Flow (blue) rapidly drives error down and keeps it orders of magnitude below GenCFD (red) across all samples.}
  \label{fig:error_evol_grid}
\end{figure}

We compute the mean‐squared error at each timestep \(t\) and average across all \(C\) channels:
\[
  \overline{\mathrm{MSE}}(t)
  = \frac{1}{C}\sum_{c=1}^C \mathrm{MSE}_c(t).
\]
Figure~\ref{fig:error_evol_grid} shows, for Samples 1–6, the log‐scaled average MSE (left) and the normalized error \(\overline{\mathrm{MSE}}(t)/\overline{\mathrm{MSE}}(0)\) (right).  Rectified Flow (blue) consistently reduces error faster and to a lower residual than GenCFD (red).

\FloatBarrier
\FloatBarrier

\subsection*{Evolution of the Energy Spectrum}
We examine the 2D radial power spectrum of the energy field \(E\) at five normalized diffusion times \(\tau\in\{0.00,0.25,0.50,0.75,1.00\}\).  
Each figure below shows log–log power vs.\ wavenumber for Rectified Flow (blue), GenCFD (red), and the ground-truth target (black dashed), with inset log-MSE error annotations. The leftmost snapshot corresponds to the spectrum of the initial data $u_0$ upon which the model is conditioned. Subsequent panels track the spectral evolution of the model predictions, as diffusion time progresses.

\FloatBarrier
\begin{figure}[t]
  \centering
  \includegraphics[width=0.95\linewidth]{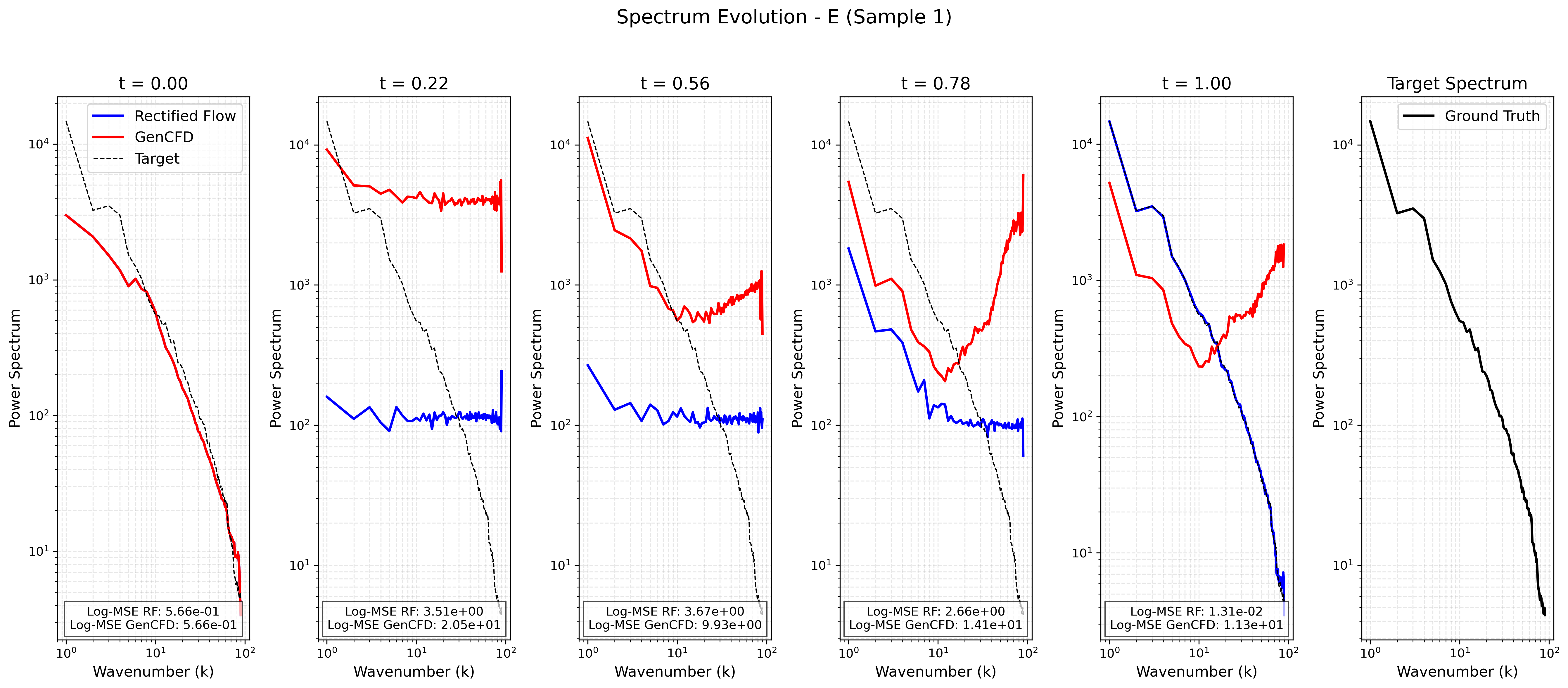}
  \caption{Energy spectrum evolution for Sample 1. ReFlow (blue) outperforms GenCFD (red) in capturing high-wavenumber energy.}
  \label{fig:spec_s1_E}
\end{figure}

\begin{figure}[t]
  \centering
  \includegraphics[width=0.95\linewidth]{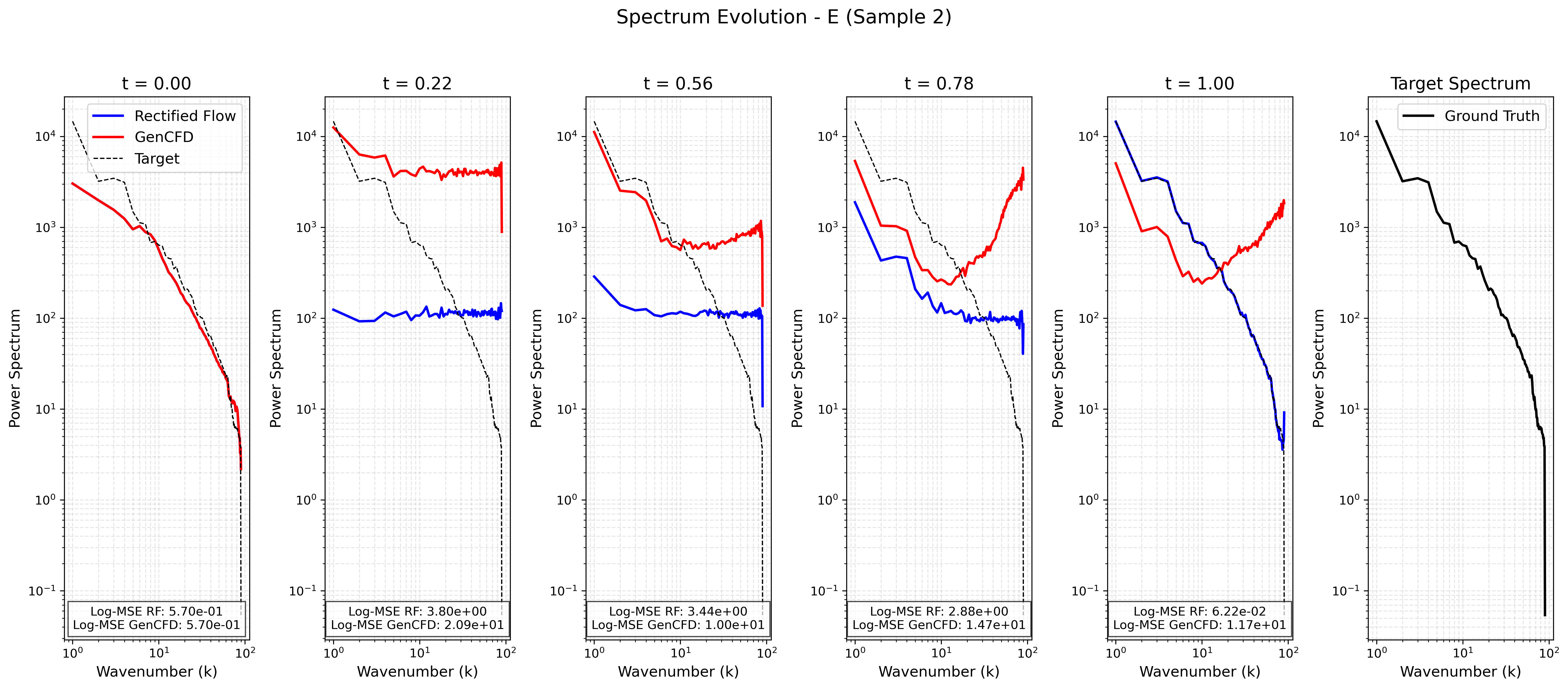}
  \caption{Energy spectrum evolution for Sample 2. ReFlow (blue) remains closer to the ground truth (dashed) than GenCFD (red).}
  \label{fig:spec_s2_E}
\end{figure}

\begin{figure}[t]
  \centering
  \includegraphics[width=0.95\linewidth]{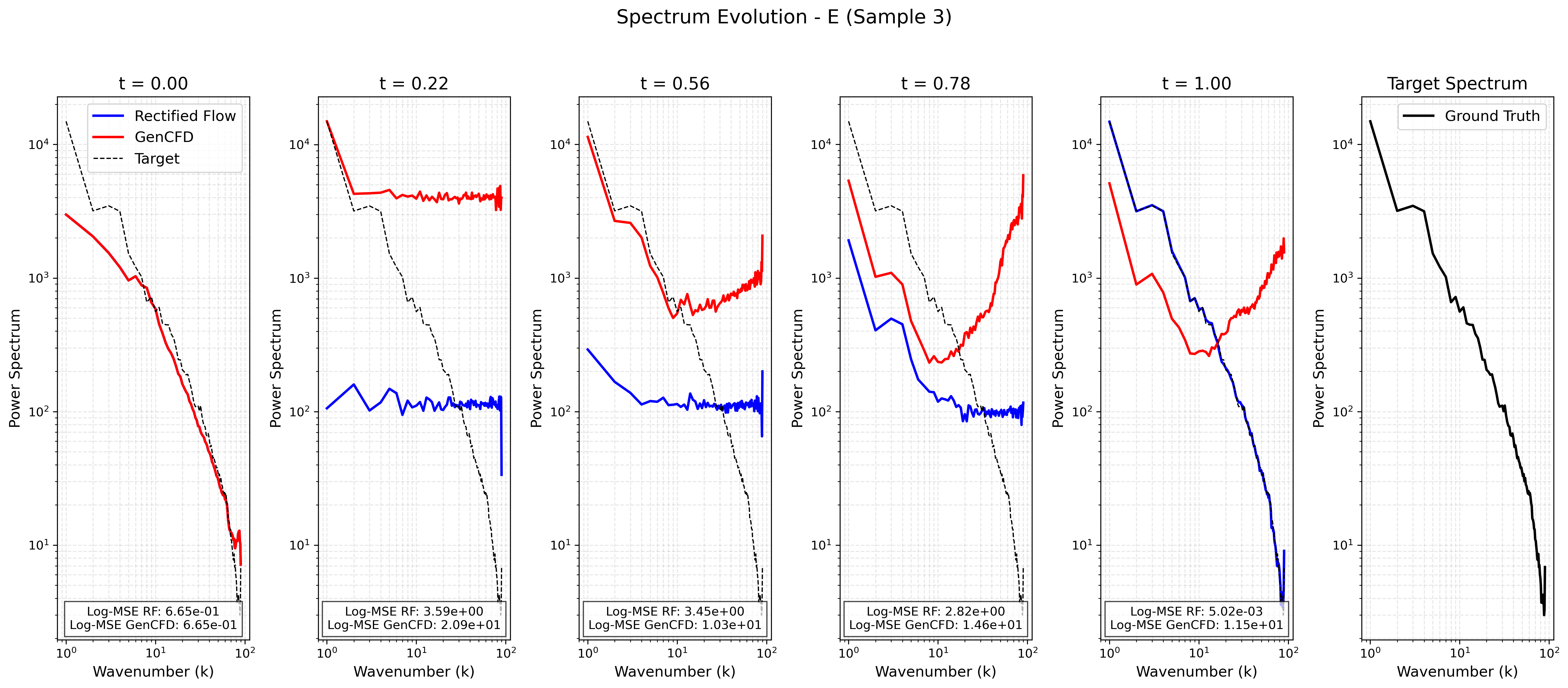}
  \caption{Energy spectrum evolution for Sample 3. The inset log-MSE annotations quantify ReFlow’s advantage at each \(t\).}
  \label{fig:spec_s3_E}
\end{figure}

\begin{figure}[t]
  \centering
  \includegraphics[width=0.95\linewidth]{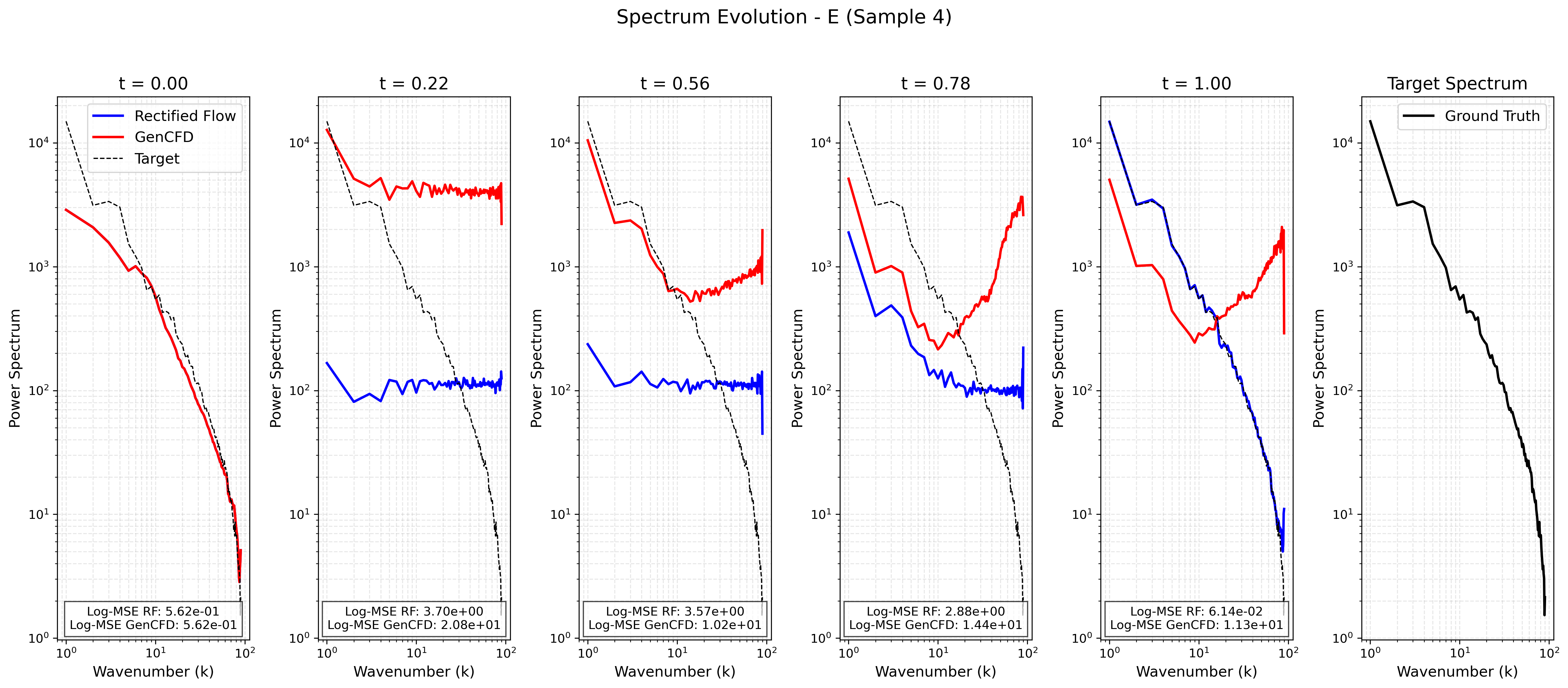}
  \caption{Energy spectrum evolution for Sample 4. ReFlow’s spectrum (blue) gets closer to the target (dashed) as \(\tau \to 1.0\).}
  \label{fig:spec_s4_E}
\end{figure}

\begin{figure}[t]
  \centering
  \includegraphics[width=0.95\linewidth]{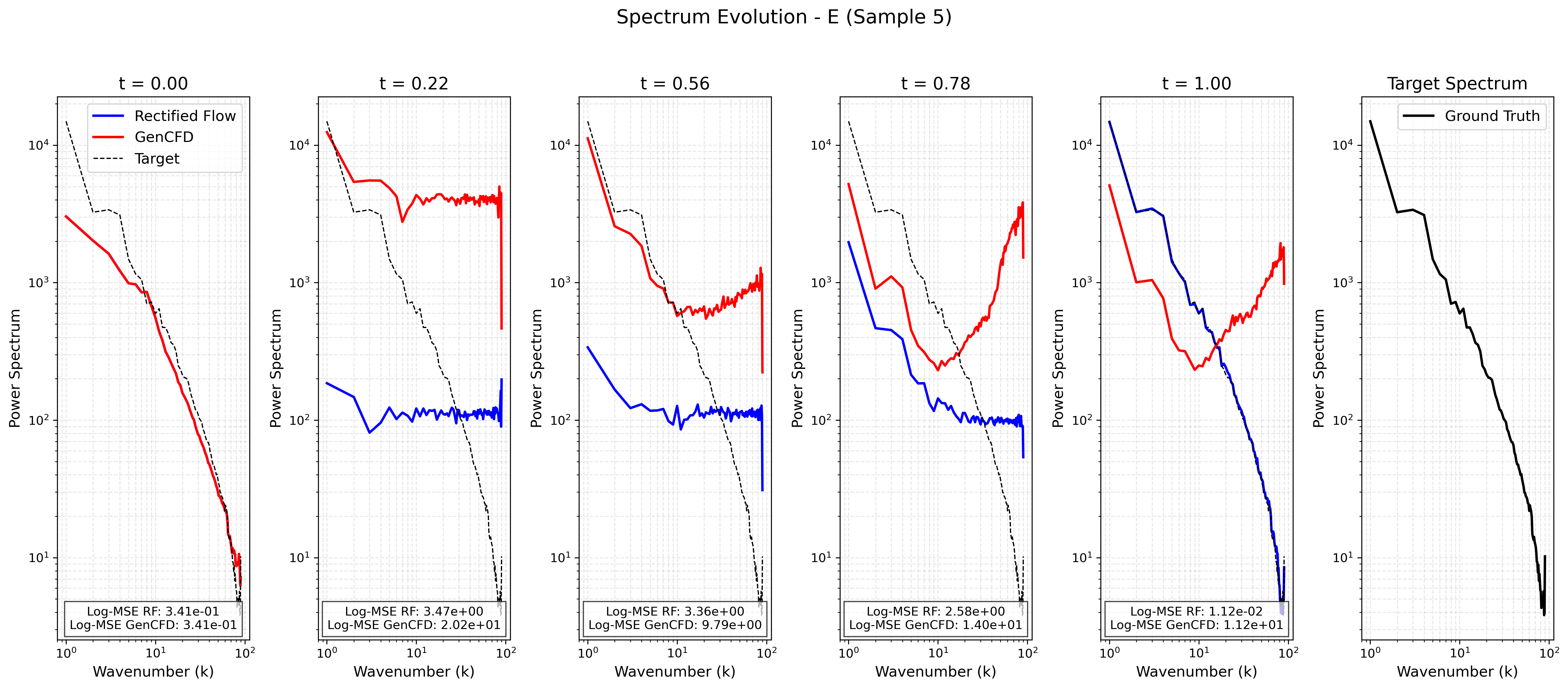}
  \caption{Energy spectrum evolution for Sample 5. ReFlow’s superior high-\(k\) fidelity is evident throughout.}
  \label{fig:spec_s5_E}
\end{figure}

\begin{figure}[t]
  \centering
  \includegraphics[width=0.95\linewidth]{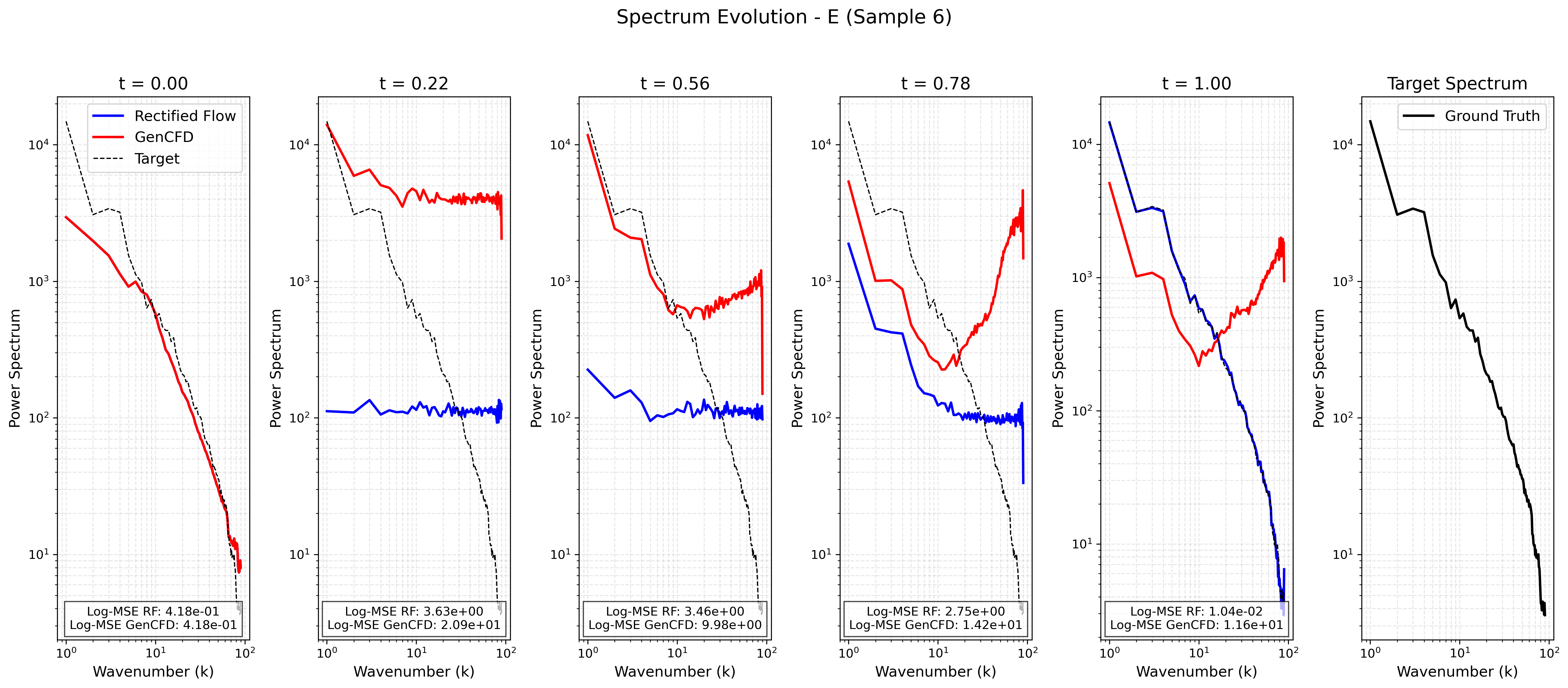}
  \caption{Energy spectrum evolution for Sample 6. ReFlow better captures the ground‐truth spectrum’s slope across all timesteps.}
  \label{fig:spec_s6_E}
\end{figure}
\FloatBarrier

\section{Computational Considerations}

All of our ReFlow models for 2D fluid-flow tasks use the same network configuration summarized in Table~\ref{tab:rf_params} (channels \((128,256,256)\), downsample ratios \((2,2,2)\), 8 attention heads, dropout 0.1, batch size 16, etc.).  Consequently, every model—regardless of dataset—has the same total parameter count (approximately \(8\text{–}10\)M parameters) and identical layer‐wise dimensions.  We ran each training job for 100 000 iterations on a single NVIDIA Quadro RTX 6000 (Driver 570.124.06, CUDA 12.8) in persistence mode.  Resource utilization averaged as follows over the full 100 h window:

\FloatBarrier
\begin{table}[h]
  \centering
  \caption{Aggregate hardware utilization for all “usual‐setup” runs.}
  \label{tab:resource-usage}
  \begin{tabular}{lcc}
    \toprule
    Metric                                & Range                & Typical              \\
    \midrule
    GPU power draw (W)                    & 200–260              & 230                  \\
    GPU utilization (\%)                  & 30–80                & 55                   \\
    GPU memory allocated (GB)             & 12–17                & 15                   \\
    GPU temperature (°C)                  & 40–60                & 50                   \\
    Process GPU memory usage (\%)         & 60–80                & 70                   \\
    Process GPU memory access time (\%)   & 10–50                & 30                   \\
    Process CPU threads in use            & 7                     & 7                    \\
    System CPU utilization (per core, \%) & 60–100               & 80                   \\
    Process memory in use (GB)            & up to 50             & 25                   \\
    Disk I/O read (MB total)              & up to $1.2\times10^{6}$ & $5\times10^{5}$     \\
    Disk utilization (\%)                 & 70–75                & 72                   \\
    Network traffic (bytes total)         & $\approx3\times10^{13}$ & $3\times10^{13}$   \\
    \bottomrule
  \end{tabular}
\end{table}

Since all datasets are paired with models of identical size and the hardware envelope (GPU power, memory, temperature, CPU threads, I/O) is effectively the same across runs, any observed performance differences are attributable to the sampling strategy (noise schedule, ODE steps) rather than to model‐size or hardware variations (the network traffic is due to Wandb usage).

All experiments were run on a single NVIDIA Quadro RTX 6000 GPU (Driver 570.124.06, CUDA 12.8), operating in persistence mode.  

Each model was trained for 100 000 iterations.  We explored architectures ranging from approximately 5 million to 10 million trainable parameters; across this range we observed no significant differences i n final predictive performance or convergence behavior.  All other hyperparameters (learning rate schedules, batch sizes, noise schedules, etc.) were held fixed when evaluated against each other, ensuring that differences in accuracy and runtime reflect only the conditioning and sampling strategies under study.

\section{Code and Data Availability}
Prior to acceptance, we will make available all three datasets, the complete data-processing scripts, and a fully functional training and evaluation pipeline for ReFlow, including a provided model checkpoint that can be loaded and tested. A more comprehensive, user-friendly code release will follow shortly thereafter.

\section{Additional Visualizations} \label{app:additional_error_plots}

\paragraph{Test‐time ensemble perturbation (evaluation only).}  
Following the protocol of \emph{GenCFD} \cite{molinaro2024generative}, we construct small “micro‐ensembles’’ around each test initial condition $\bar u$ by sampling $M_{\mathrm{micro}}$ perturbed fields uniformly in an $\varepsilon$–ball centered at $\bar u$.  Each perturbed member is then advanced to the evaluation time $t$ with a high‐fidelity PDE solver to produce a reference ensemble $\{u^{(j)}_{\rm ref}(t)\}_{j=1}^{M_{\mathrm{micro}}}$.  Our model generates a corresponding ensemble $\{u^{(j)}_{\rm model}(t)\}_{j=1}^{M_{\mathrm{micro}}}$ from the same perturbed inputs, but without ever observing ensembles during training (which only uses independent pairs $(u_0,u_1)$).  We compare these two ensembles via mean‐field error $e_\mu$, standard‐deviation error $e_\sigma$, and spatially averaged 1‐Wasserstein distance $\overline W_1$ (see \textbf{SM} §\ref{app:results}).

\medskip\noindent
\textbf{Out‐of‐distribution testing (cloud–shock and shear‐flow).}  
Because the micro‐perturbations render each initial field slightly OOD, this evaluation probes the model’s ability to generalize under small chaotic perturbations.  In these tasks, our Rectified Flow ensemble closely tracks the reference uncertainty and multiscale statistics, whereas deterministic baselines collapse to a single trajectory and fail to capture the conditional variability.

\medskip\noindent
\textbf{In‐distribution scaling (Richtmyer–Meshkov).}  
For the Richtmyer–Meshkov dataset, we additionally study the effect of training set size using a held‐out test set from the same distribution.  Here, even deterministic approaches improve with more data, but still underperform on the micro‐ensemble test: they require significantly larger training sets to match ReFlow’s accuracy and exhibit poor calibration under small perturbations.  

\begin{figure}[t]
  \centering
  \subfloat[UVit global $E(k)$.]{%
    \includegraphics[width=0.45\textwidth]{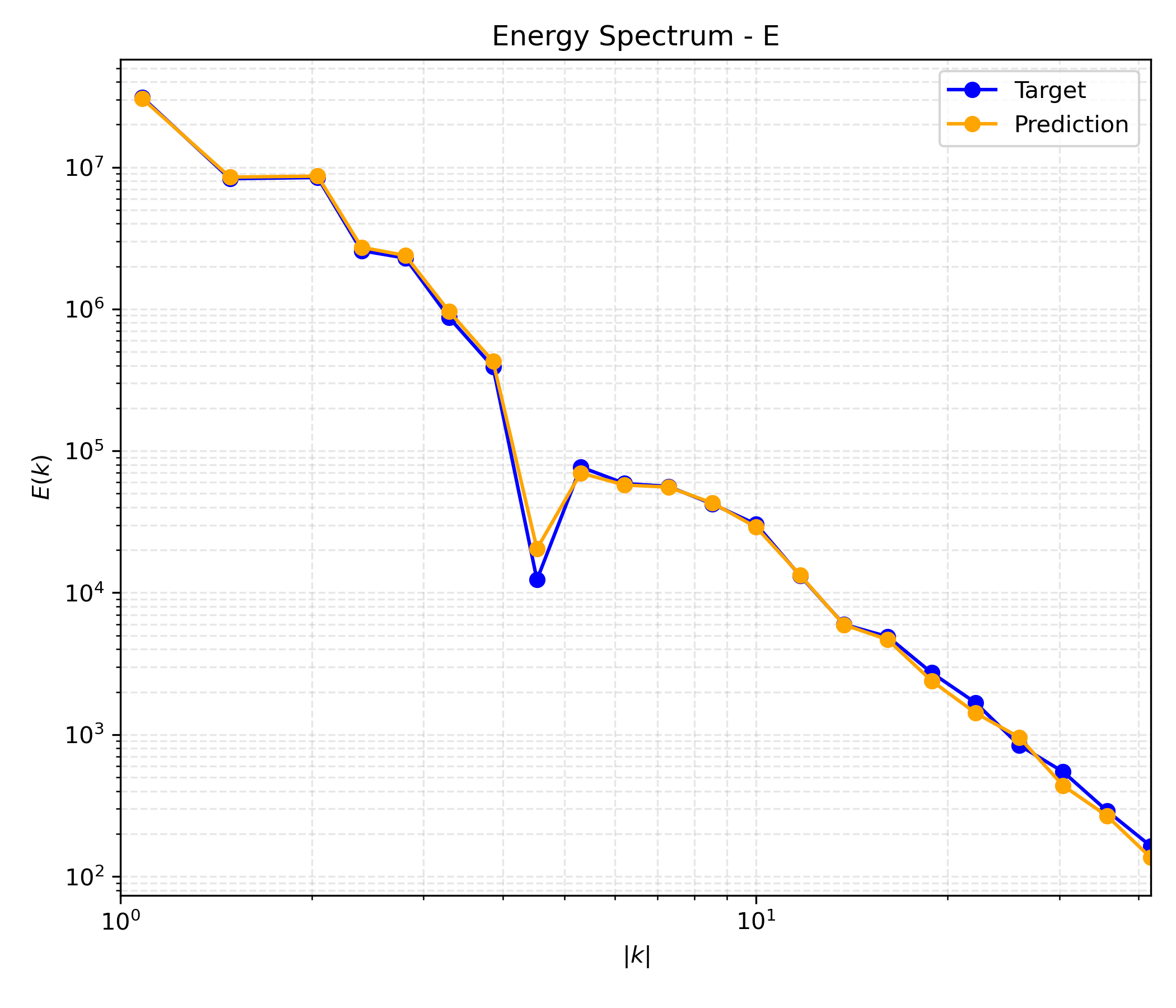}}
  \hfill
  \subfloat[RF global $E(k)$.]{%
    \includegraphics[width=0.45\textwidth]{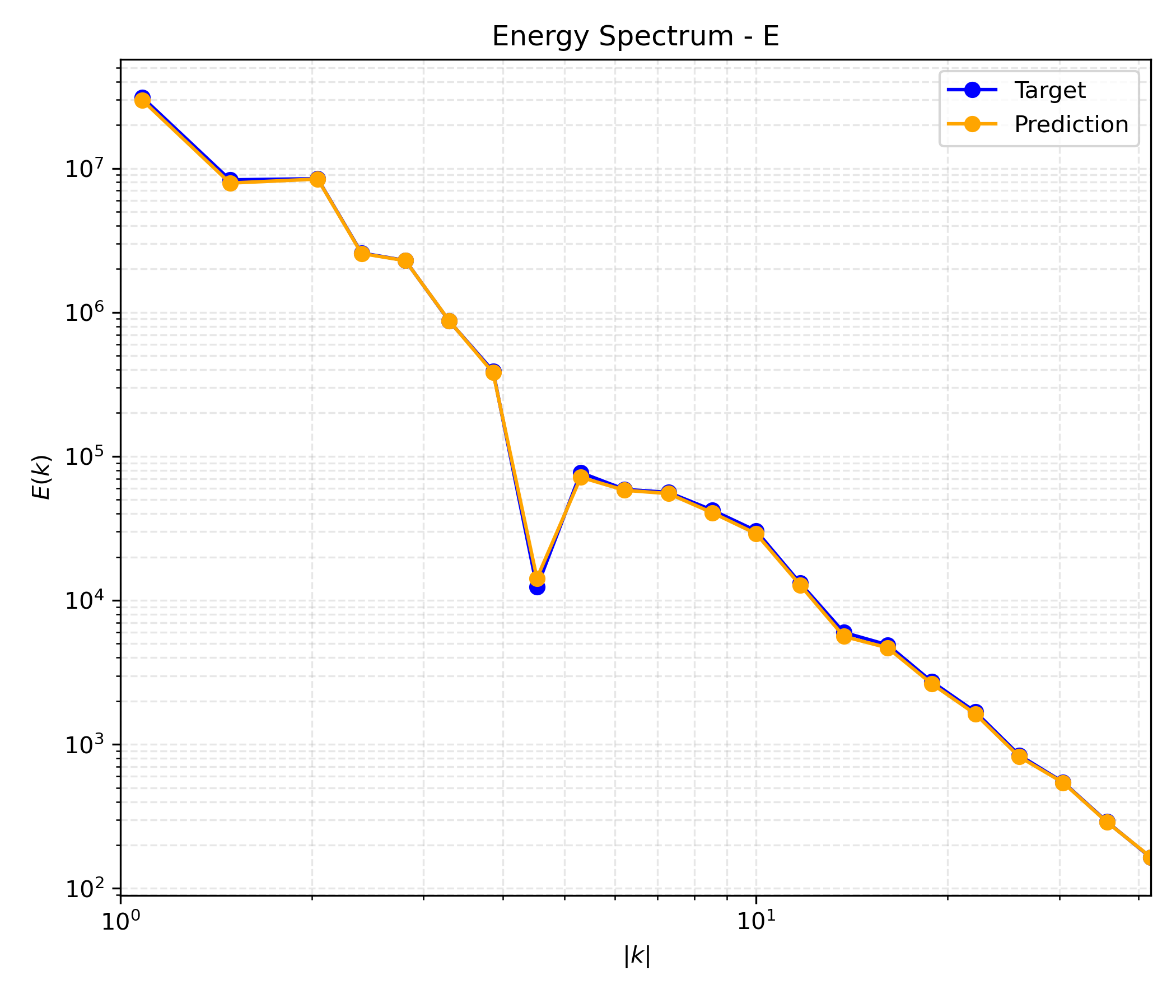}}\\[1ex]
  \subfloat[UVit local at $(120,44)$.]{%
    \includegraphics[width=0.45\textwidth]{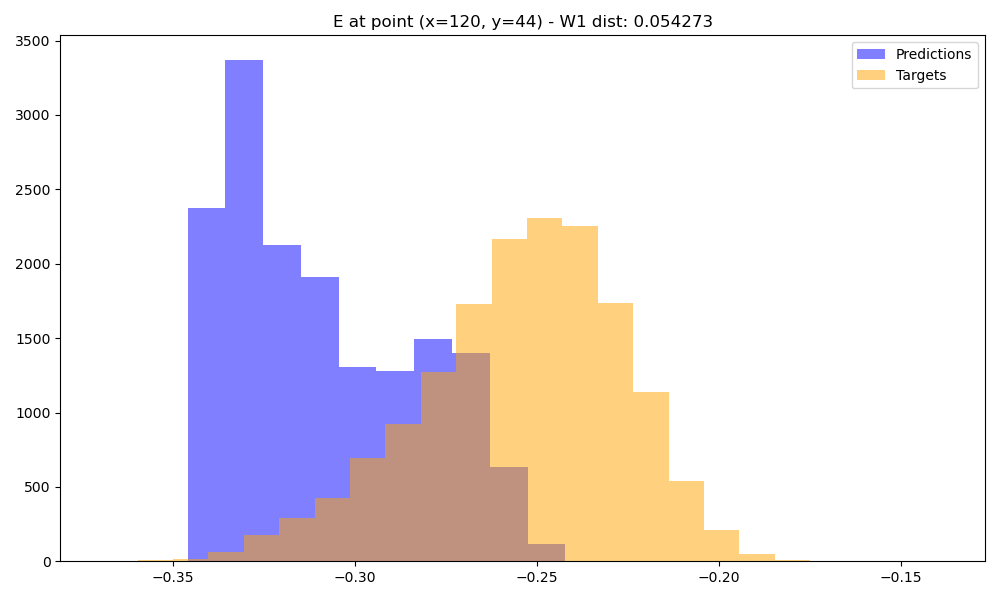}}
  \hfill
  \subfloat[RF local at $(120,44)$.]{%
    \includegraphics[width=0.45\textwidth]{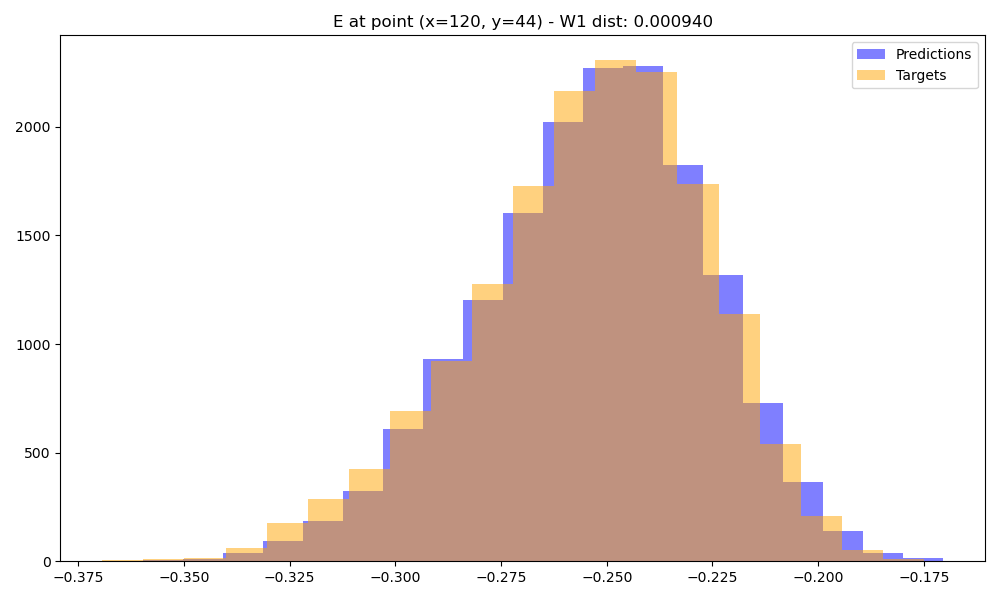}}\\[1ex]
  \subfloat[UVit $\rho$ at $(60,12)$, $W_1=0.0278$.]{%
    \includegraphics[width=0.45\textwidth]{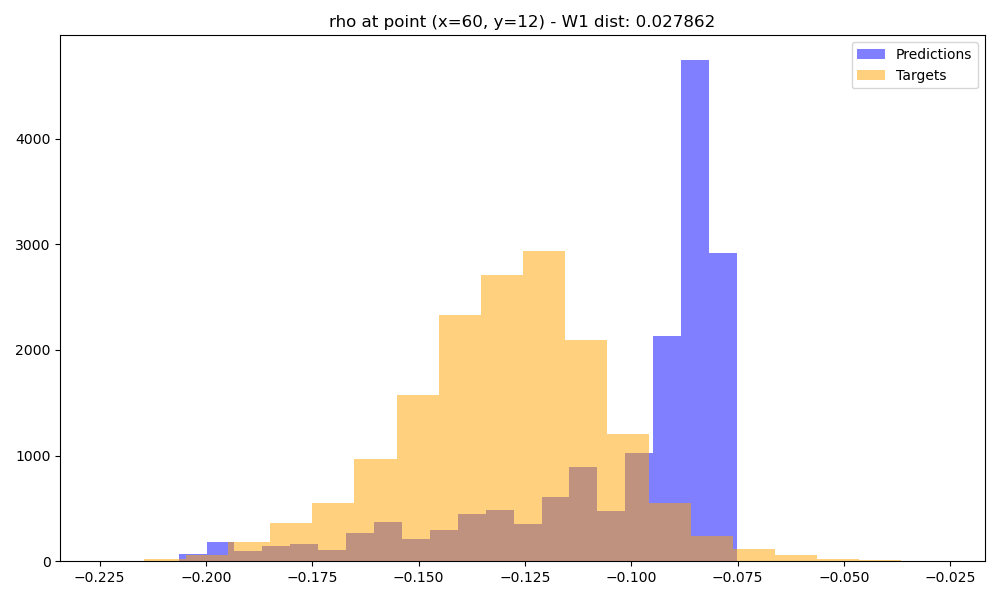}}
  \hfill
  \subfloat[RF $\rho$ at $(60,12)$, $W_1=0.0005$.]{%
    \includegraphics[width=0.45\textwidth]{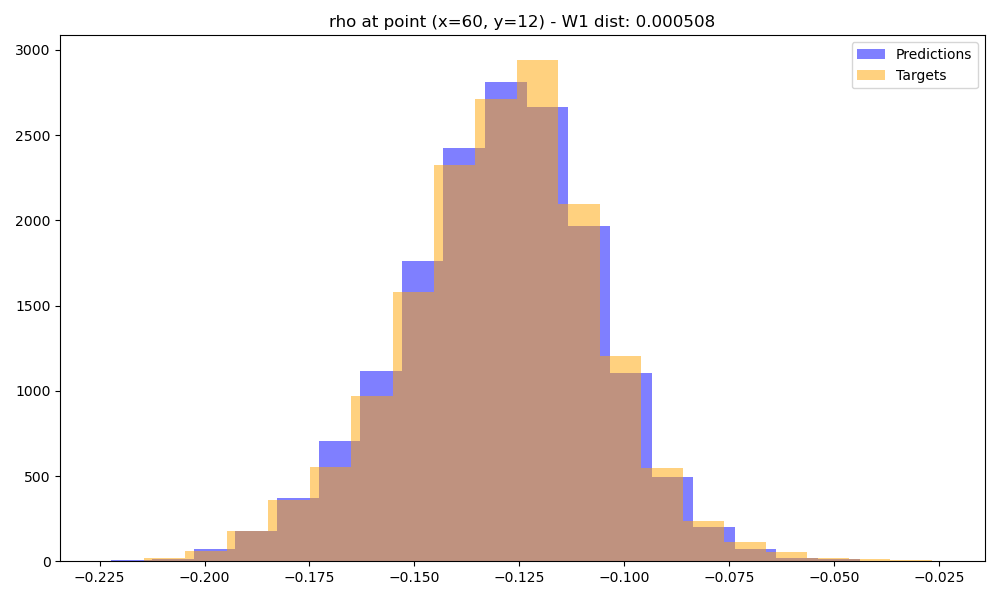}}
  \caption{Richtmyer–Meshkov: spectral and pointwise distribution comparisons between UVit and Rectified Flow.}
  \label{fig:rm_spectra_histograms}
\end{figure}

\begin{figure}[t]
  \centering
  \subfloat[FNO $E_{u_x}(r)$.]{%
    \includegraphics[width=0.45\textwidth]{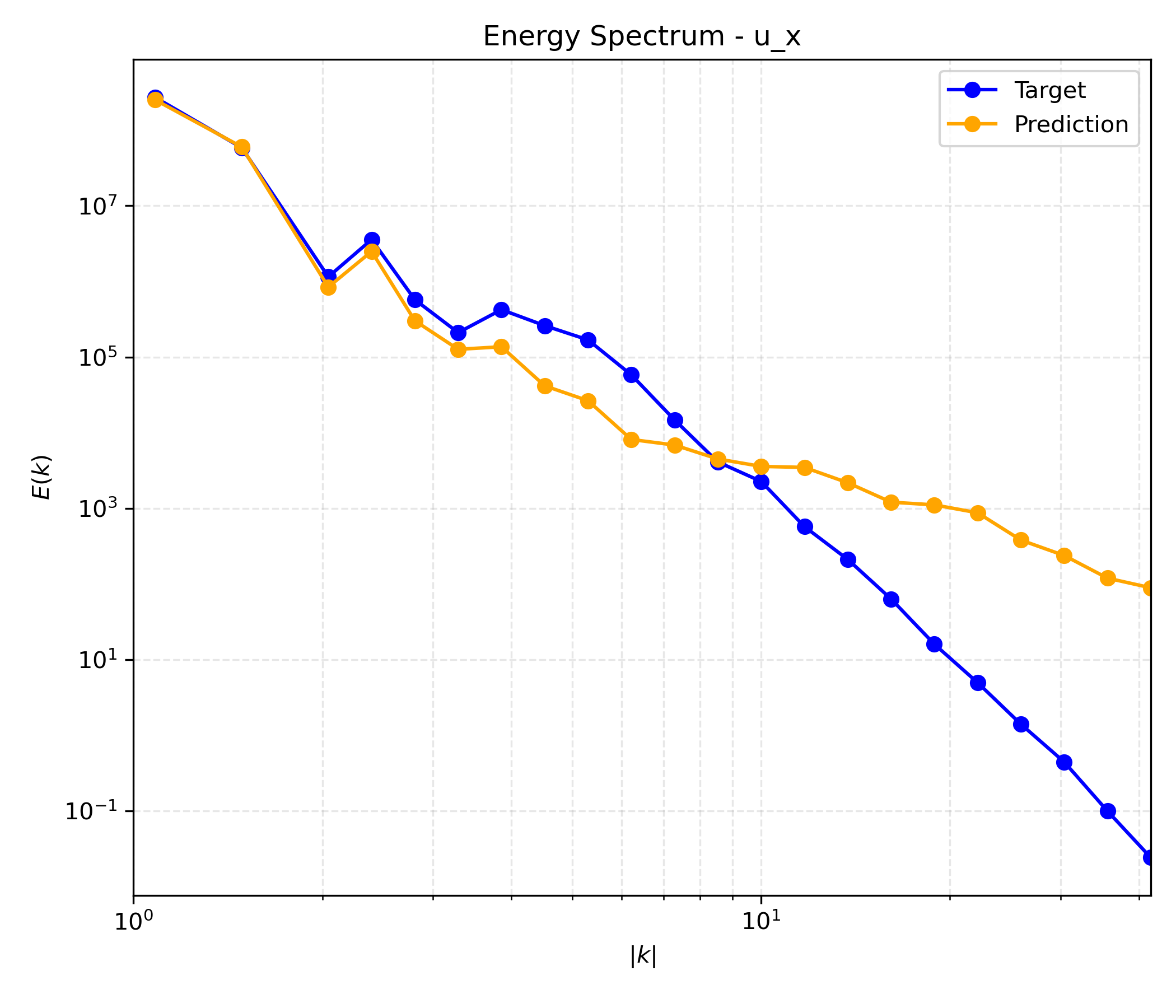}}
  \hfill
  \subfloat[RF $E_{u_x}(r)$.]{%
    \includegraphics[width=0.45\textwidth]{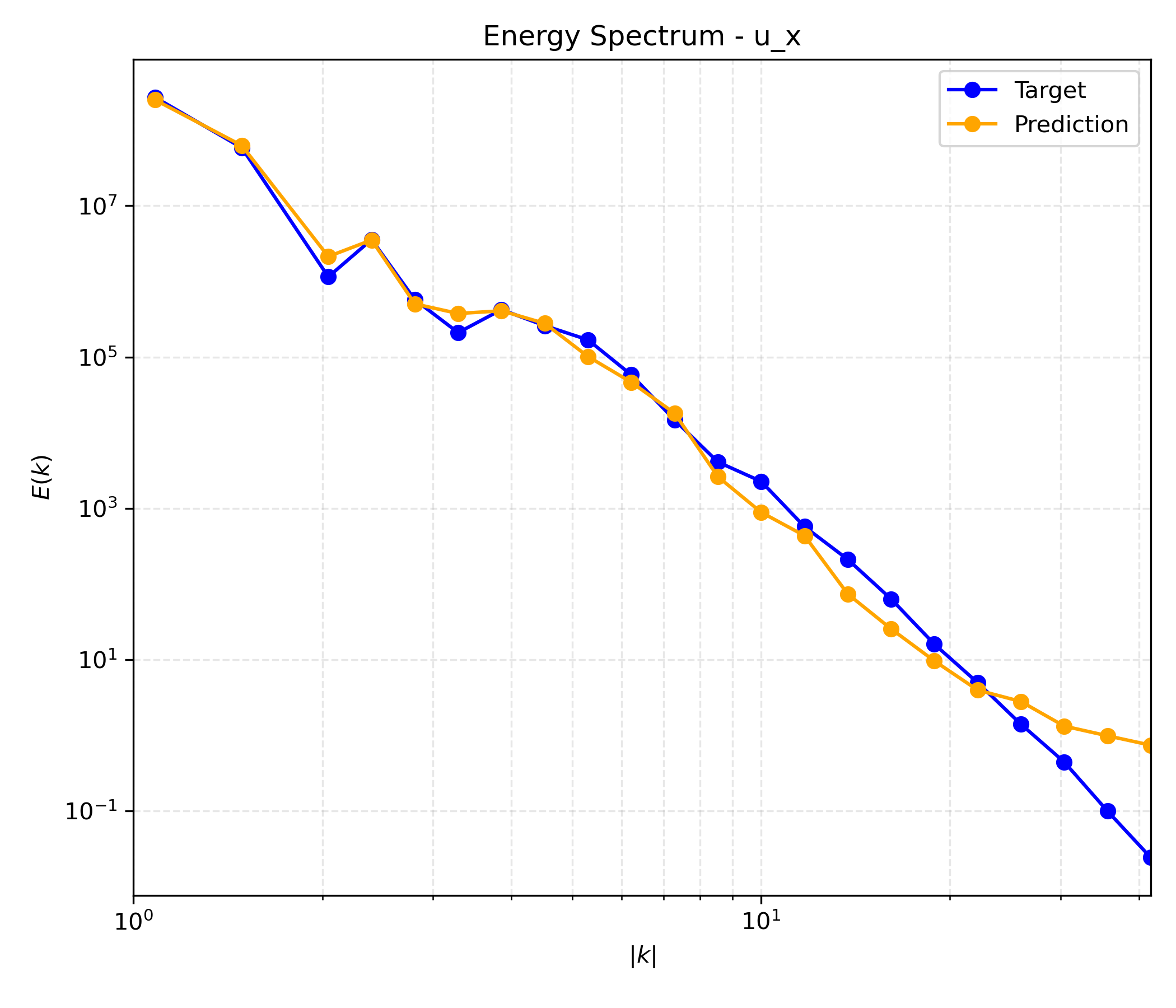}}\\[1ex]
  \subfloat[FNO $u_x$ at $(36,8)$.]{%
    \includegraphics[width=0.45\textwidth]{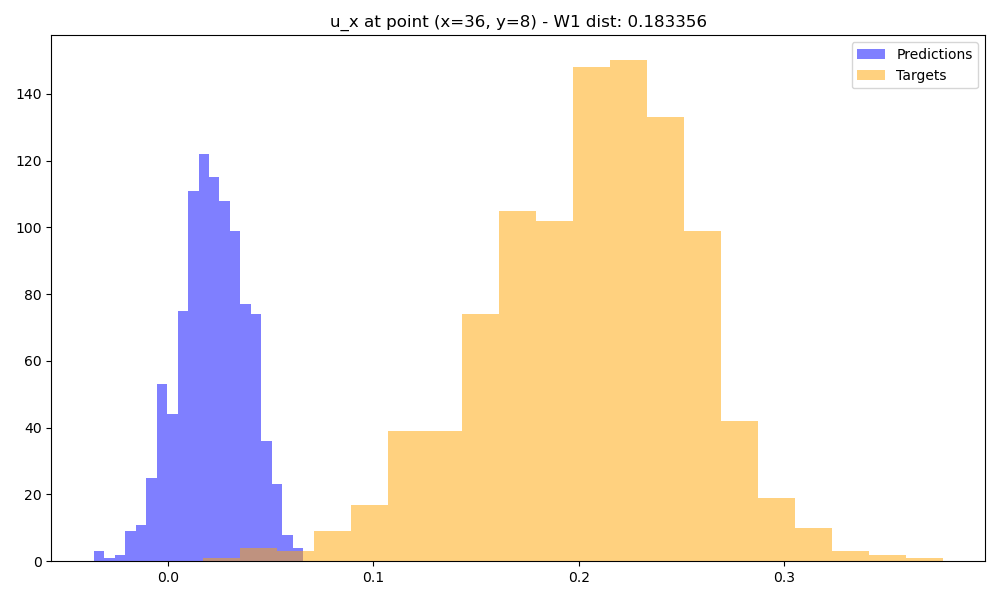}}
  \hfill
  \subfloat[RF $u_x$ at $(36,8)$.]{%
    \includegraphics[width=0.45\textwidth]{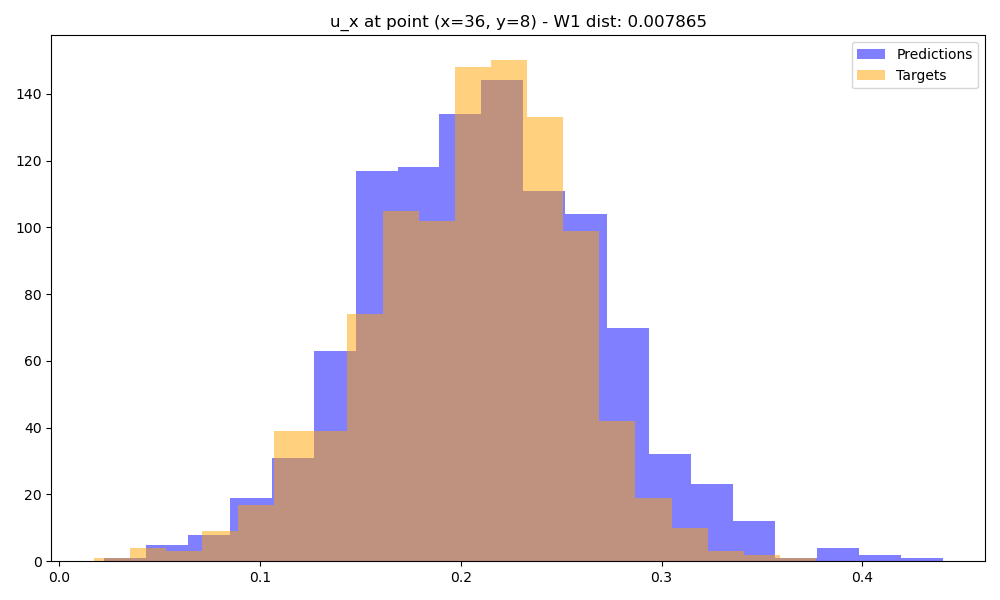}}
  \caption{Shear‐layer: global $u_x$ spectra and pointwise $u_x$ distributions.}
  \label{fig:sl_spectrum_pointwise_ux}
\end{figure}

\begin{figure}[t]
  \centering
  \subfloat[UVit $m_y$ spectrum.]{%
    \includegraphics[width=0.45\textwidth]{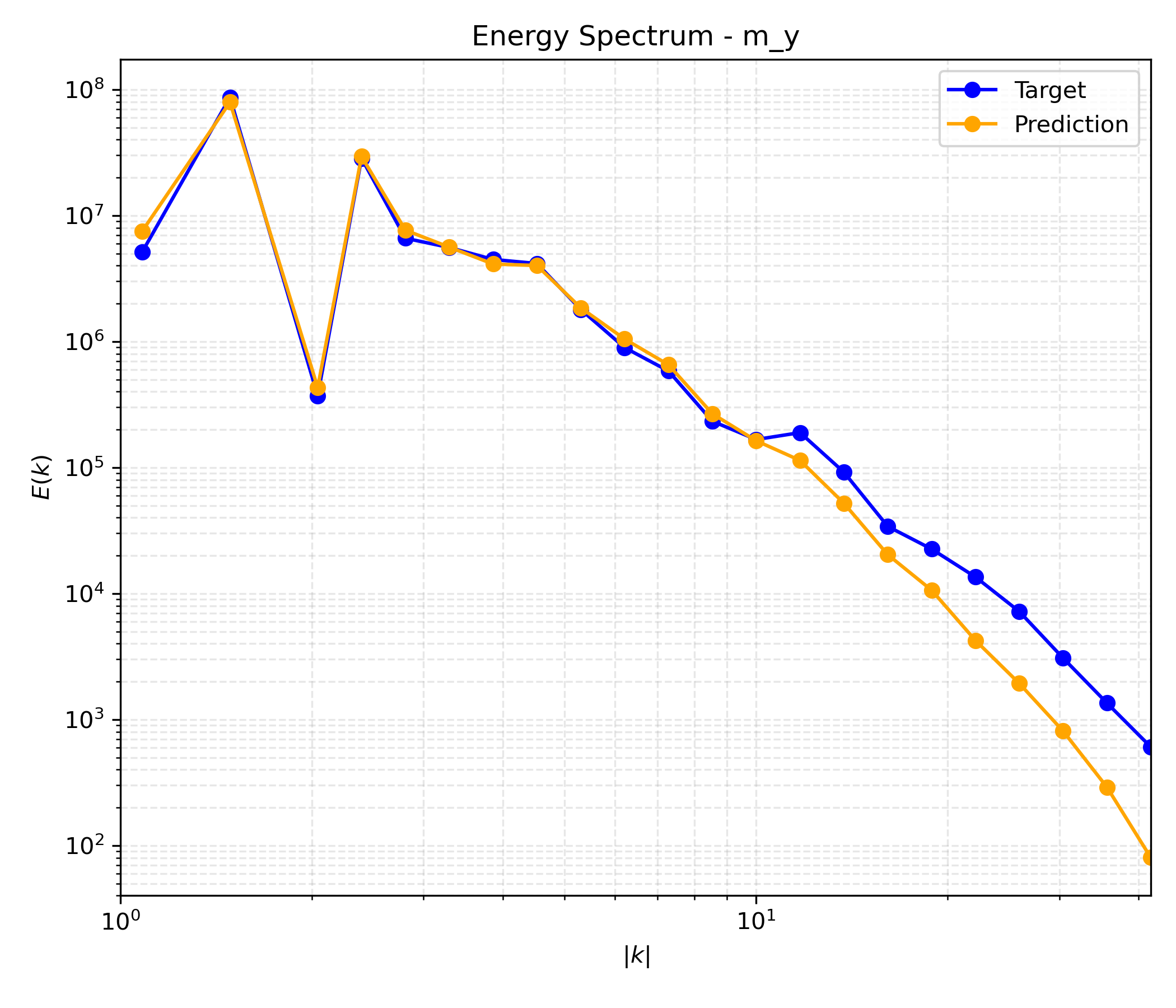}}
  \hfill
  \subfloat[RF $m_y$ spectrum.]{%
    \includegraphics[width=0.45\textwidth]{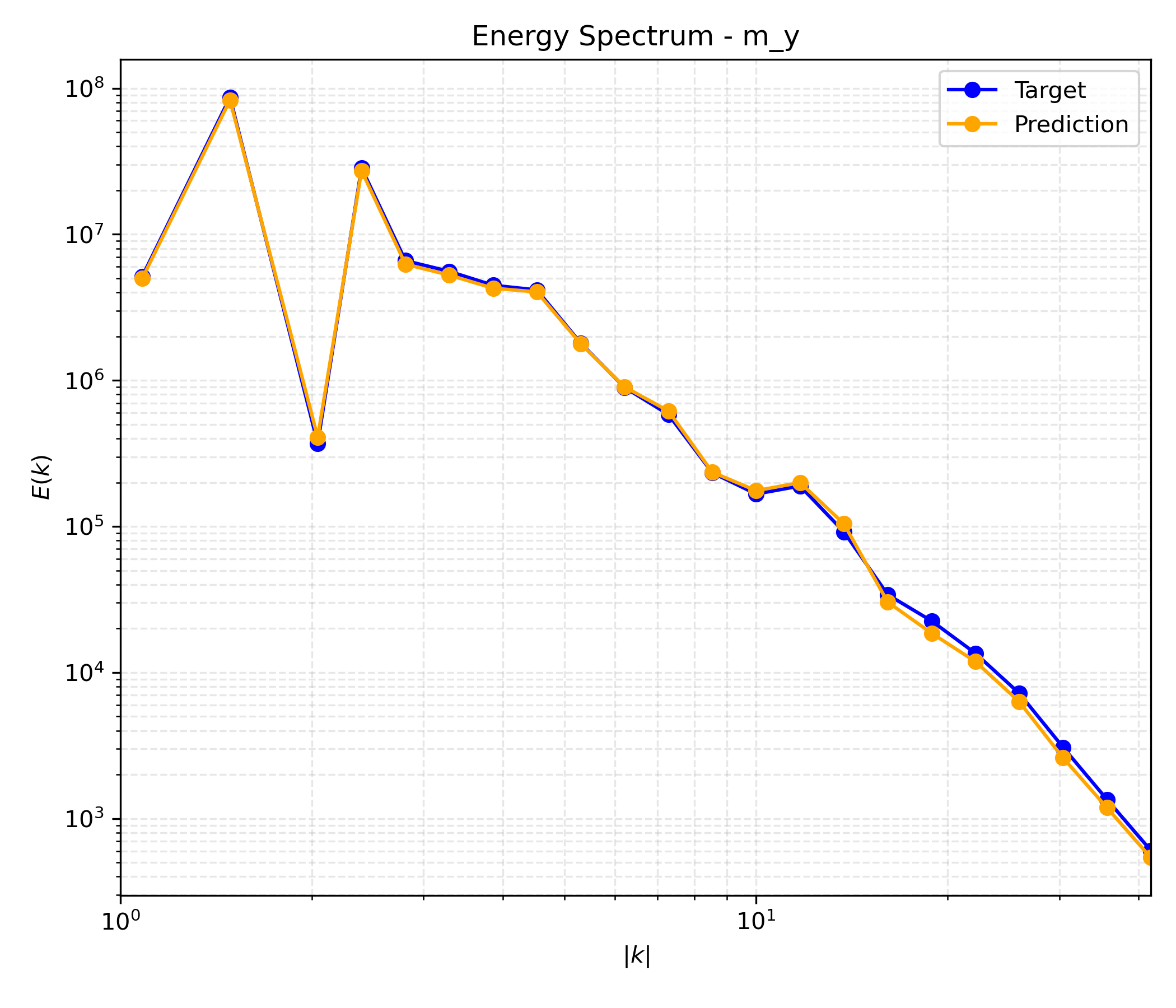}}\\[1ex]
  \subfloat[UVit $E$ at $(68,28)$.]{%
    \includegraphics[width=0.45\textwidth]{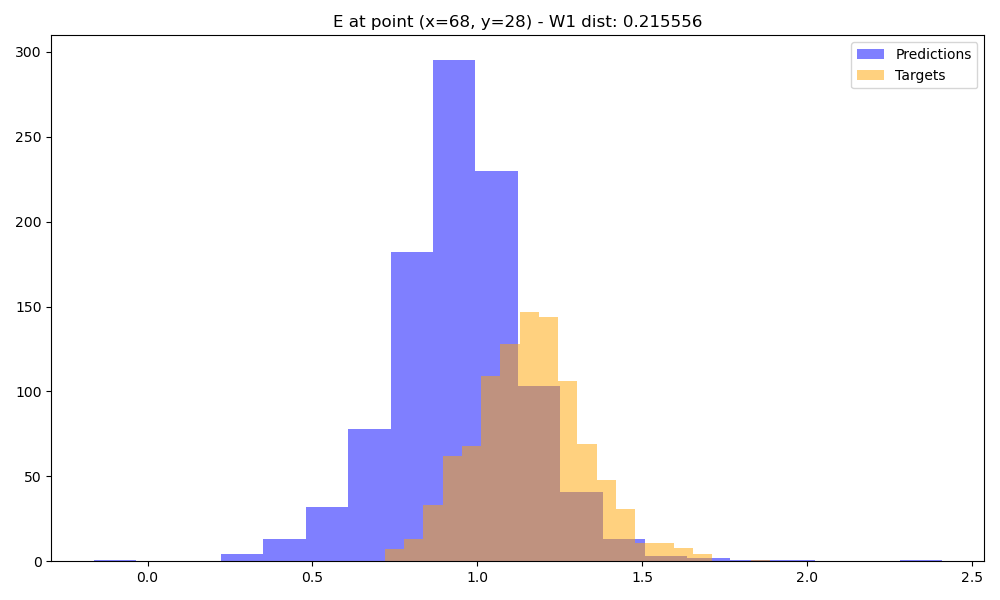}}
  \hfill
  \subfloat[RF $E$ at $(68,28)$.]{%
    \includegraphics[width=0.45\textwidth]{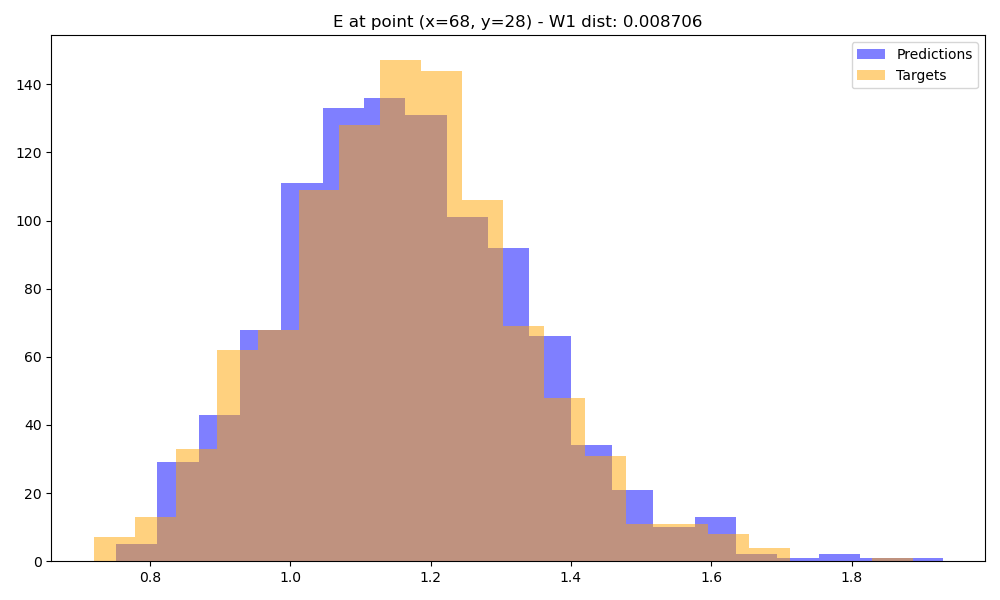}}
  \caption{Cloud‐shock: spectral and pointwise energy comparisons.}
  \label{fig:cs_spectrum_pointwise}
\end{figure}

In this appendix we provide a comprehensive set of qualitative comparisons between the baselines (GenCFD variants, UViT, FNO) models and our Rectified Flow (ReFlow) approach across three representative 2D fluid‐flow benchmarks (Richtmyer–Meshkov, shear layer roll‐up, and cloud-shock interaction).  For each task we show:

\begin{itemize}
    \item \textbf{Mean fields}  
    Ensemble averages reveal how each deterministic model diffuses fine‐scale features (UVit/FNO) versus perserving sharp interfaces (ReFlow, GenCFD variants).

    \item \textbf{Uncertainty maps}  
    Per‐pixel standard deviations illustrate whether uncertainty localizes in physically meaningful regions (e.g.\ shocks, shear interfaces) or remains overly diffuse.

    \item \textbf{Random samples}
    To qualitatively assess generative fidelity, we include representative random samples from ReFlow alongside outputs from selected baselines, specifically the deterministic FNO and the hybrid GenCFD \(\boldsymbol{\circ}\) FNO. The latter combines FNO’s low-frequency structural prediction with GenCFD’s generative modeling of fine-scale turbulence. This comparison highlights ReFlow’s ability to produce realistic, high-resolution flow fields that preserve physical coherence across different scales.

    \item \textbf{Spectral \& pointwise comparisons}  
    (Fig.~\ref{fig:rm_spectra_histograms}: Richtmyer–Meshkov global $E(k)$, local spectra, and pointwise density histograms;  
     Fig.~\ref{fig:sl_spectrum_pointwise_ux}: Shear‐layer global $u_x$ spectra and pointwise $u_x$ distributions;  
     Fig.~\ref{fig:cs_spectrum_pointwise}: Cloud–shock global $m_y$ spectra and pointwise energy distributions)  
    — combined Fourier‐space and histogram panels quantify multiscale fidelity and calibration of the predictive ensembles.
\end{itemize}

Together, these visualizations underscore ReFlow’s ability to maintain sharper physical structures, produce better‐calibrated uncertainties, and faithfully reproduce multiscale spectral statistics.  

\begin{figure}[t]
  \centering
  \includegraphics[width=\textwidth]{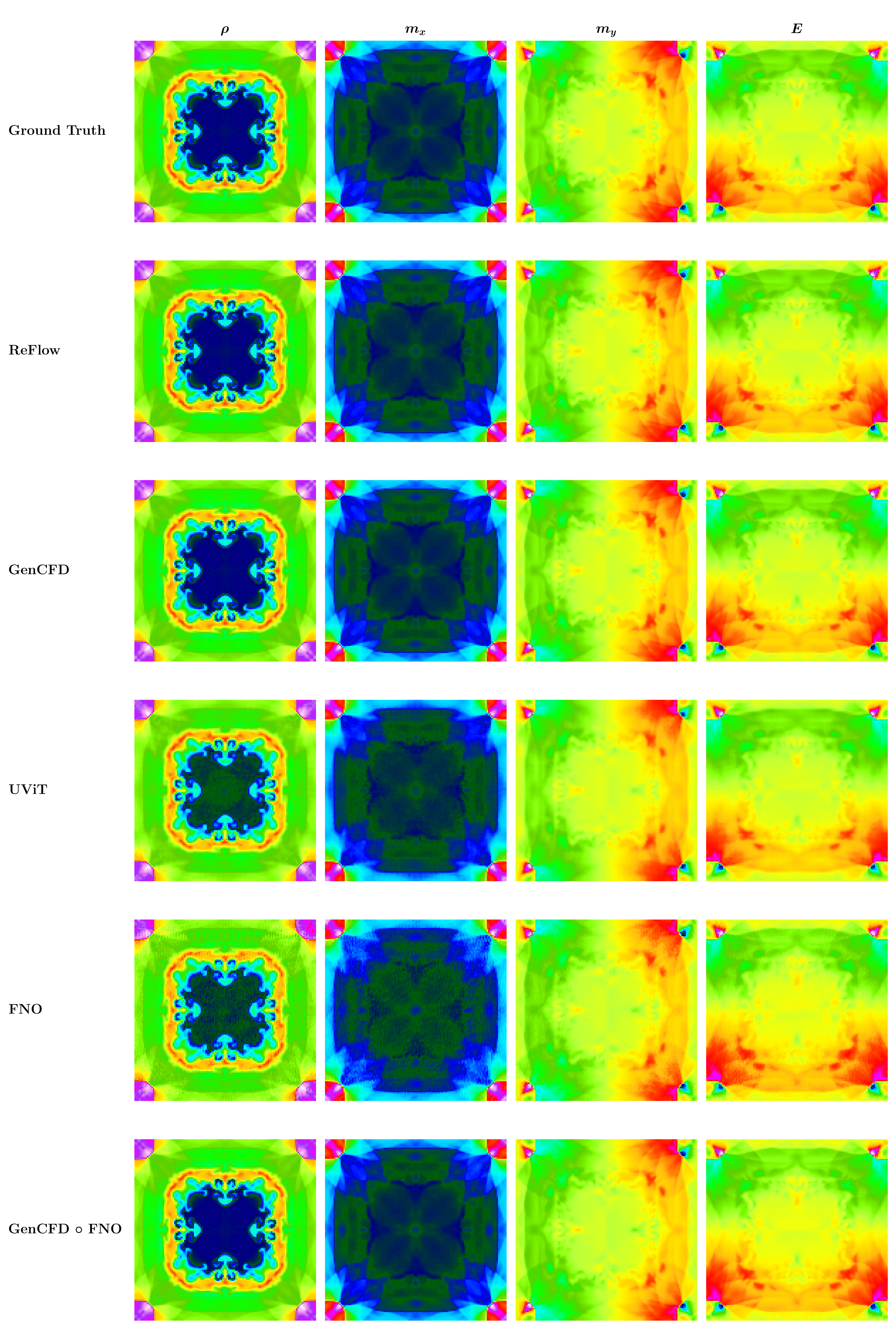}
  \caption{\textbf{Richtmyer–Meshkov Mean Field Comparison}: Mean velocity fields for the Richtmyer–Meshkov dataset, shown across all physical channels and models. In many cases, predictions closely resemble the ground truth, with differences often imperceptible, underscoring the overall high quality and consistency of the generated mean fields.}
  \label{fig:rm_mean_full}
\end{figure}


\begin{figure}[t]
  \centering
  \includegraphics[width=\textwidth]{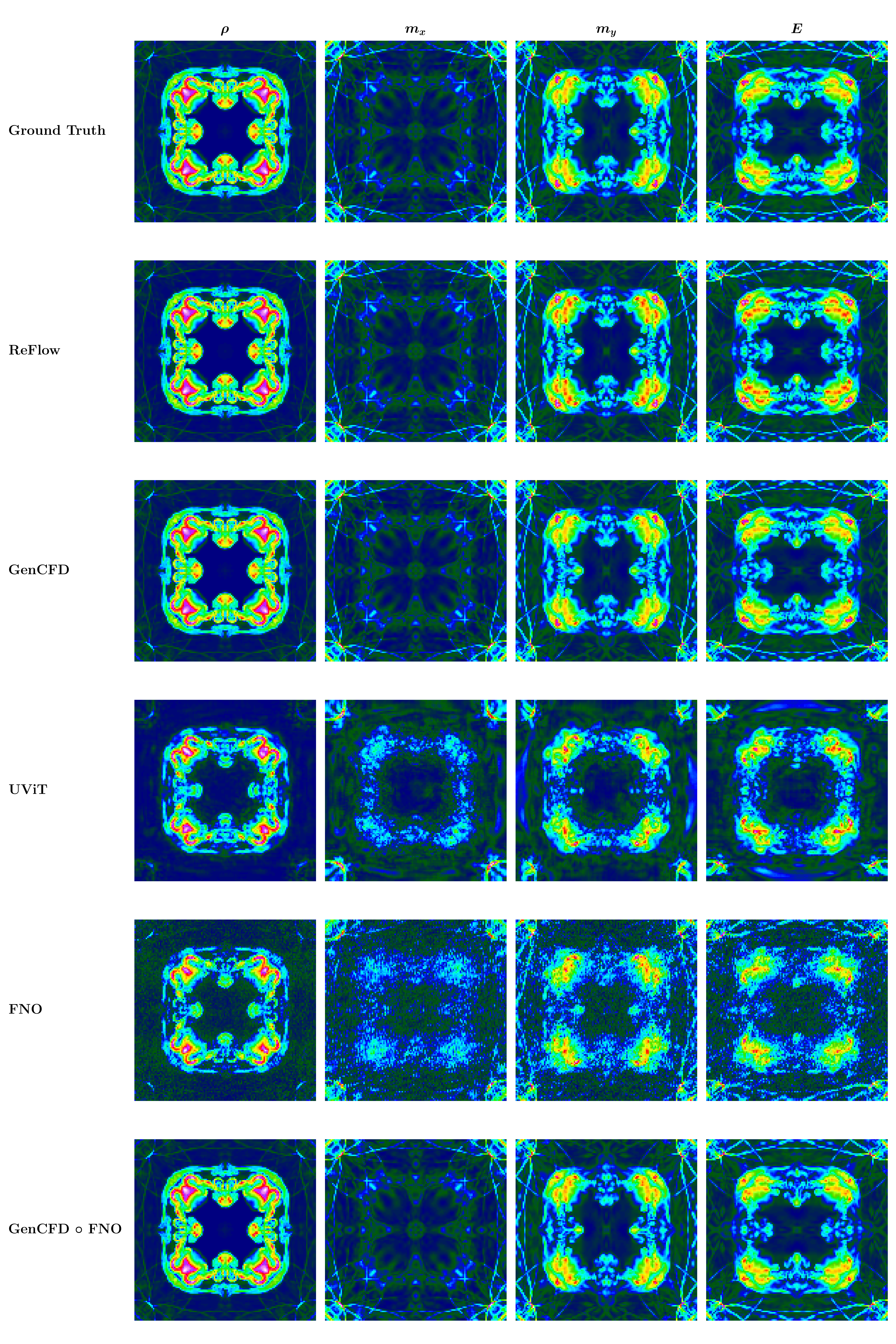}
  \caption{\textbf{Richtmyer–Meshkov Uncertainty Comparison}: Per-pixel standard deviation fields from the Richtmyer–Meshkov dataset across all models. Stochastic generative models capture spatially coherent uncertainty patterns, outperforming the deterministic baselines (FNO and UViT), which exhibit limited variability and poor uncertainty localization.}
  \label{fig:rm_std_full}
\end{figure}


\begin{figure}[t]
  \centering
  \includegraphics[width=\textwidth]{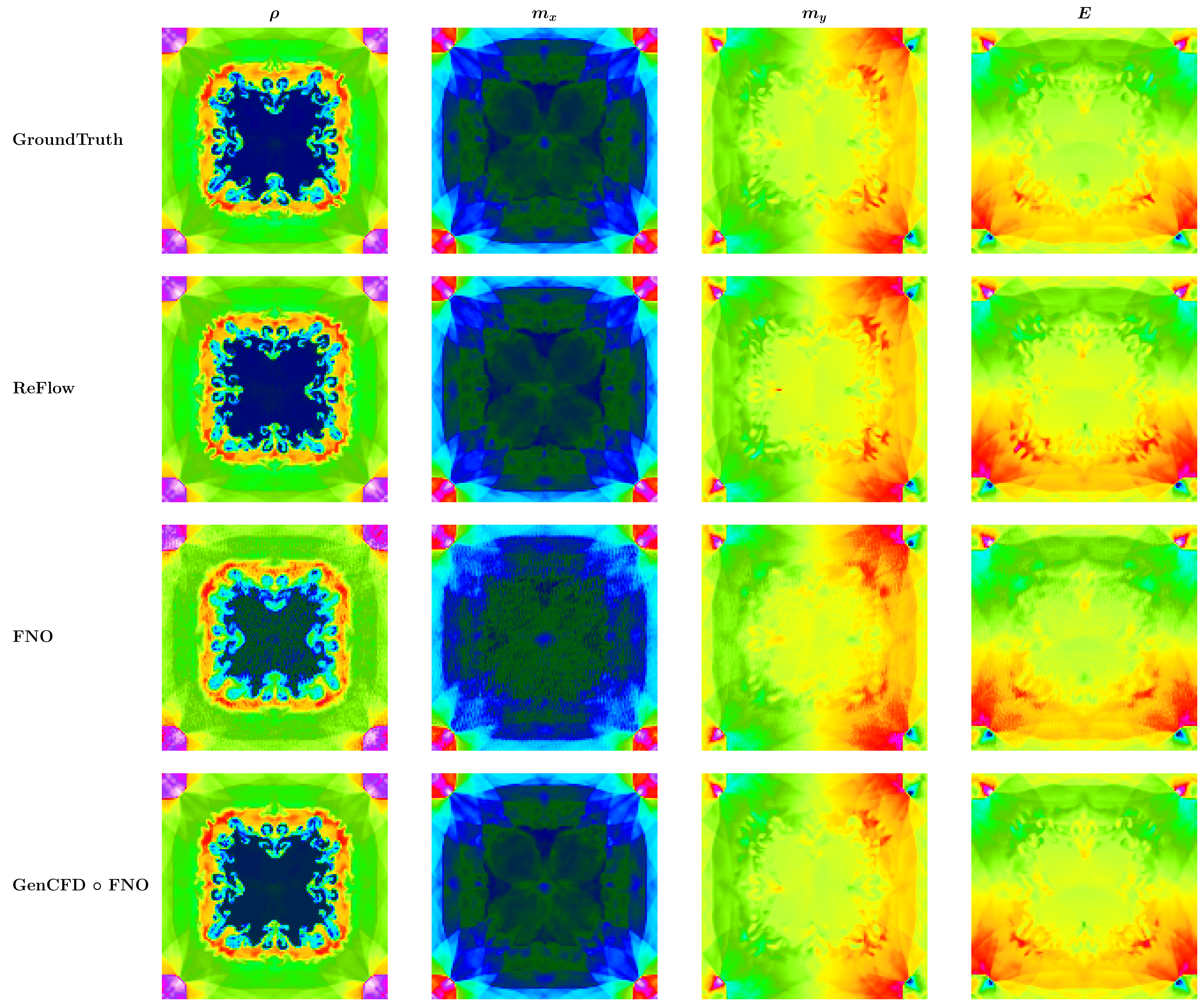}
  \caption{\textbf{Richtmyer–Meshkov Individual Sample Comparison}: A representative random test sample from the Richtmyer–Meshkov dataset, visualized across all four physical channels (\(\rho\), \(m_x\), \(m_y\), \(p\)). All models produce physically plausible results, with only little variation in fidelity to fine-scale structures. ReFlow offers the closest visual match to the ground truth, preserving coherent details and sharp interfaces. Notably, the GenCFD \(\boldsymbol{\circ}\) FNO hybrid enhances spectral richness by reconstructing high-frequency features based on a low-frequency FNO prior, illustrating its capacity to recover sharper modes even when starting from a coarse prediction.}

  \label{fig:rm_samples}
\end{figure}

\begin{figure}[t]
  \centering
  \subfloat[\textbf{mean-field $\mu$}]{%
    \includegraphics[width=0.45\textwidth]{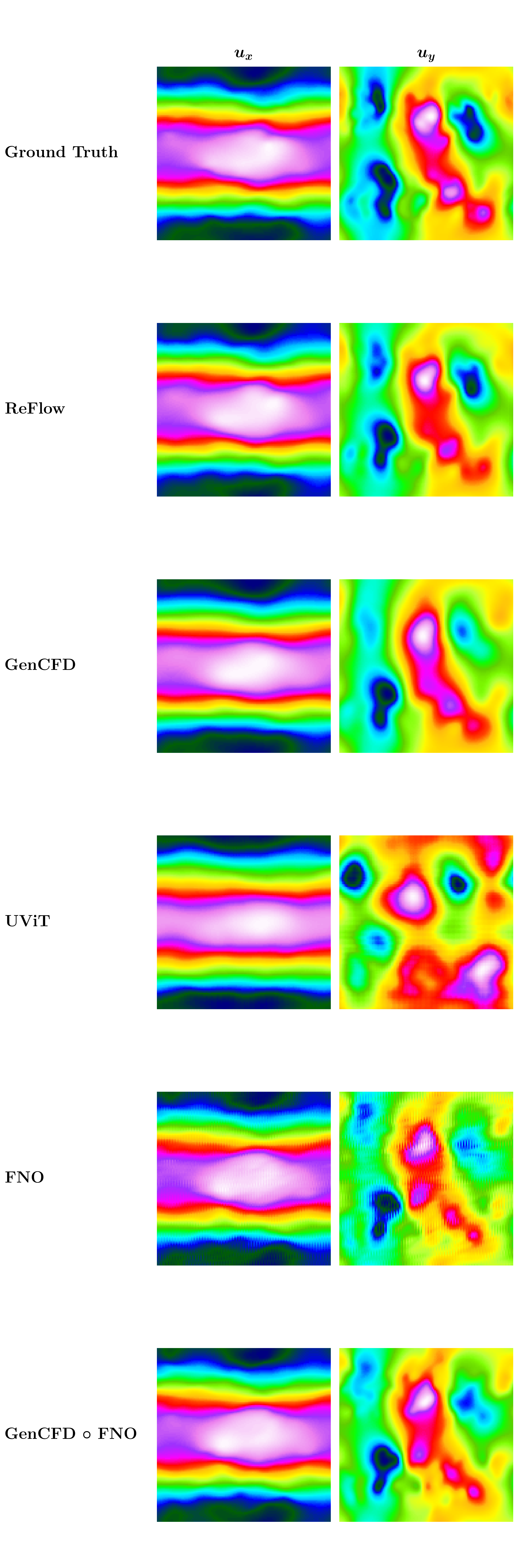}}
  \hfill
  \subfloat[\textbf{standard-deviation $\sigma$}]{%
    \includegraphics[width=0.45\textwidth]{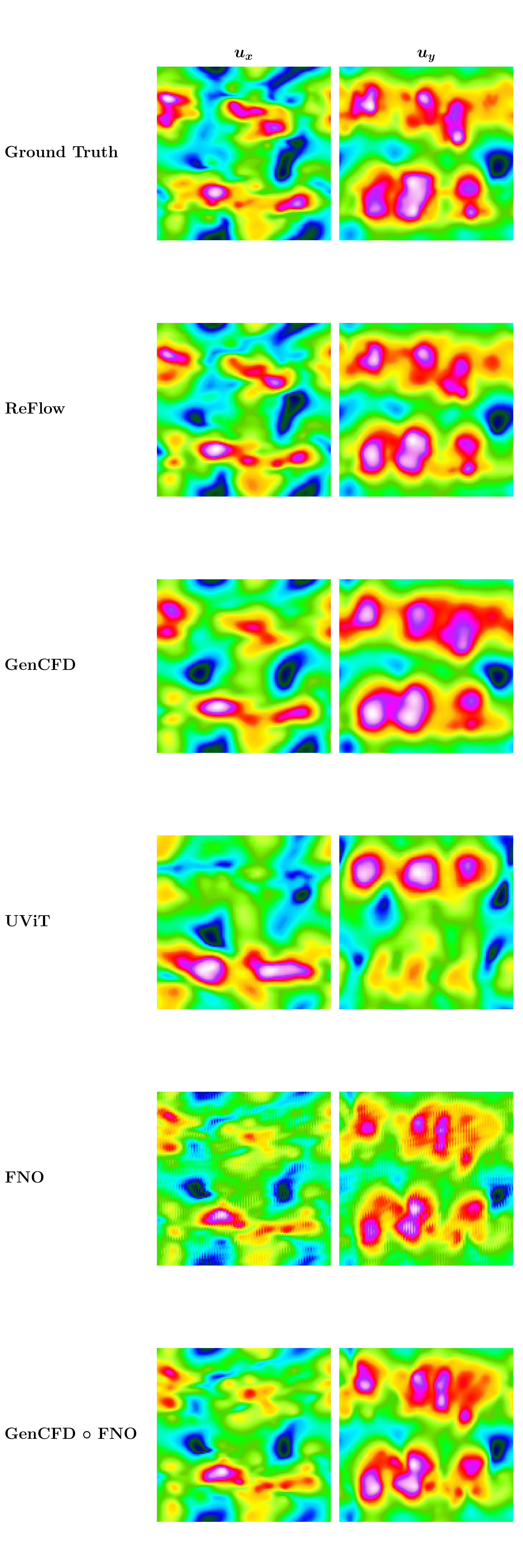}}
  \caption{\textbf{Shear‐Layer Mean and Standard Deviation Comparison}: Visualization of mean-field ($\mu$) and standard-deviation ($\sigma$) predictions for the Shear‐Layer dataset across both velocity components ($u_x$, $u_y$). For each model, the left half shows the predicted mean fields and the right half shows per-pixel standard deviations. The top row contains the reference ground truth. Despite the reduced channel complexity of this dataset, clear differences emerge in how well models capture coherent structures and localized uncertainties. ReFlow preserves sharper features and more accurately reflects the spatial distribution of uncertainty.}

  \label{fig:sl_mean_std}
\end{figure}


\begin{figure}[t]
  \centering
  \includegraphics[width=\textwidth]{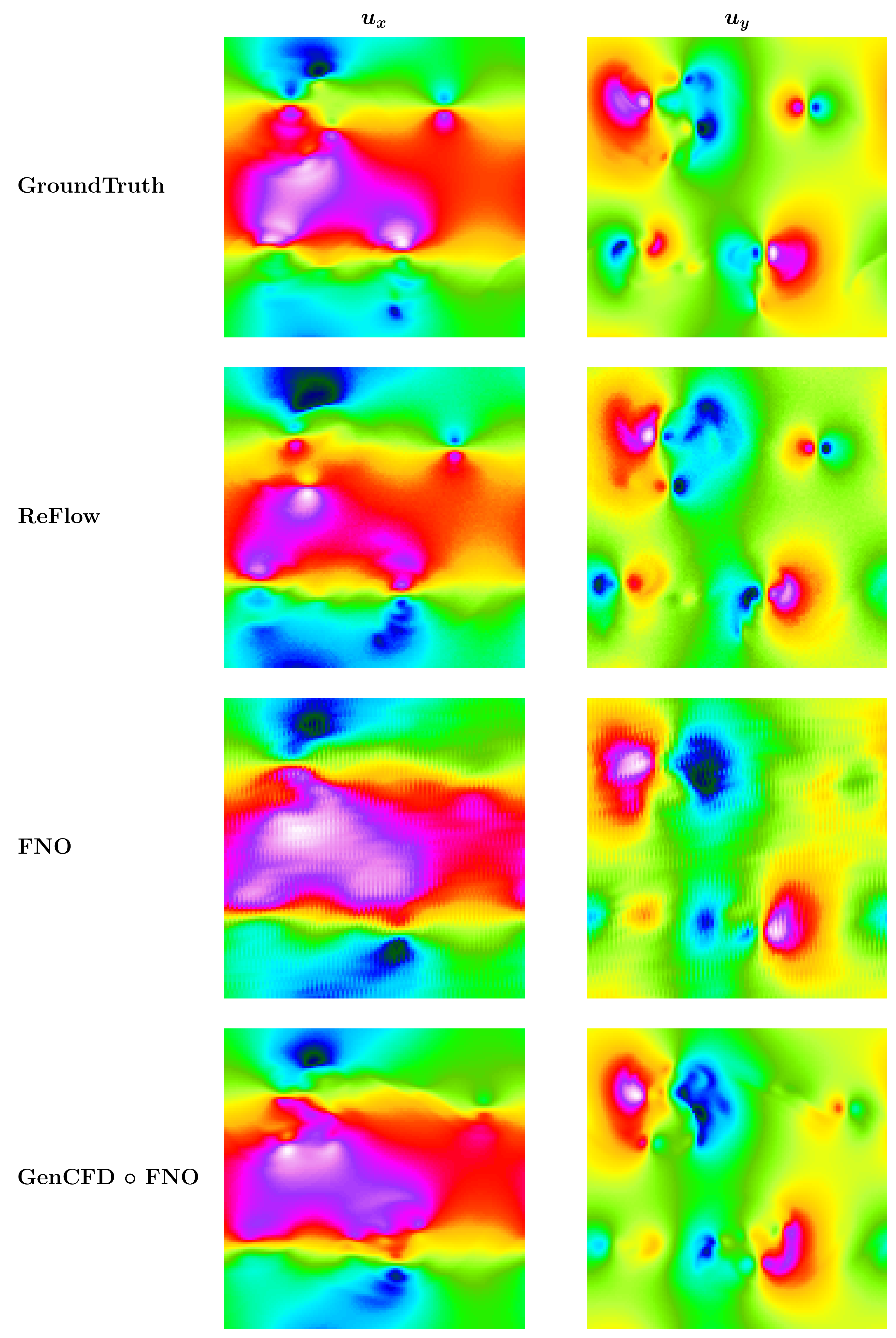}
  \caption{\textbf{Shear‐Layer Individual Sample Comparison}: A representative random test sample from the Shear‐Layer dataset, shown across both physical channels (\(u_x\), \(u_y\)). All models generate physically plausible results, though with varying degrees of detail and sharpness. ReFlow and GenCFD \(\boldsymbol{\circ}\) FNO most closely resemble the ground truth, preserving fine-scale structures and coherent interfaces. GenCFD \(\boldsymbol{\circ}\) FNO enhances high-frequency detail by leveraging the low-frequency prior from the initial FNO prediction.}
  \label{fig:sl_samples}
\end{figure}

\begin{figure}[t]
  \centering
  \includegraphics[width=\textwidth]{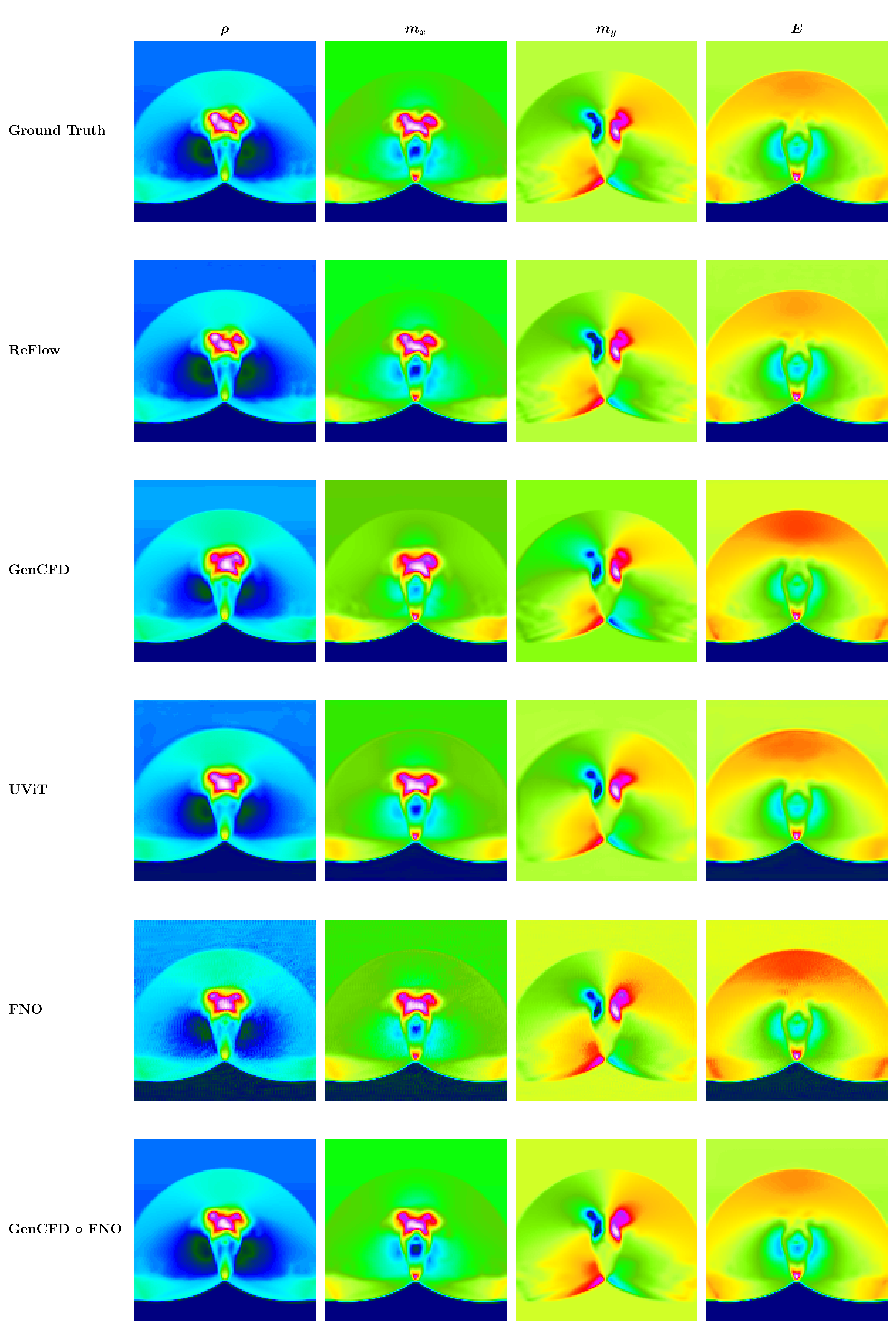}
  \caption{\textbf{Cloud-Shock Mean Field Comparison}: Mean velocity fields for the Cloud-Shock dataset, shown across all physical channels and models. In most cases, predictions closely resemble the ground truth.}
  \label{fig:cs_mean_full}
\end{figure}


\begin{figure}[t]
  \centering
  \includegraphics[width=\textwidth]{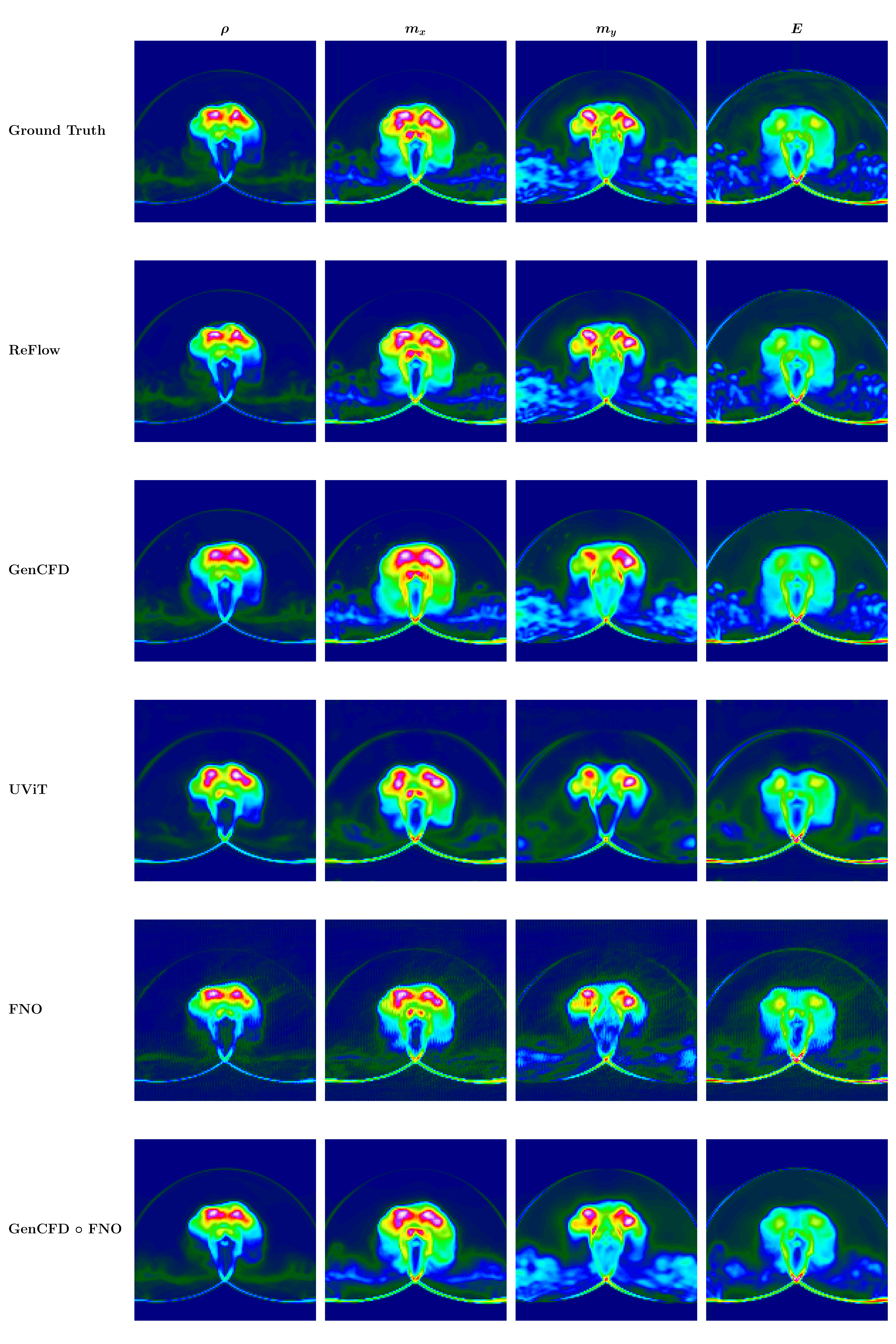}
  \caption{\textbf{Cloud–Shock Uncertainty Comparison}: Per-pixel standard deviation visualizations for the Cloud–Shock dataset across all models. While all methods capture key uncertainty patterns, stochastic models produce more localized and physically meaningful variability.}
  \label{fig:cs_std_full}
\end{figure}

\begin{figure}[t]
  \centering
  \includegraphics[width=\textwidth]{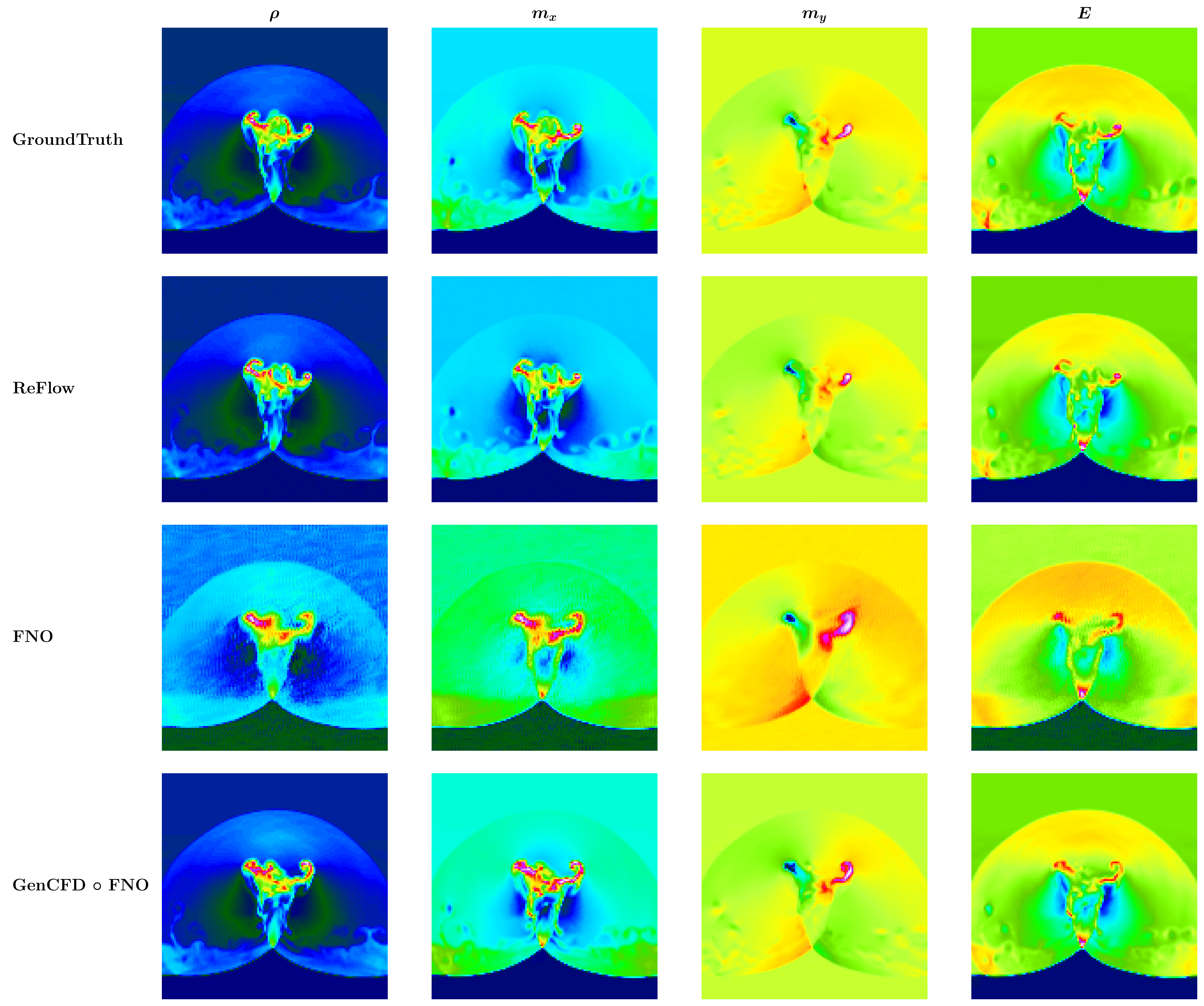}
  \caption{\textbf{Cloud–Shock Individual Sample Comparison}: A representative random test sample from the Cloud–Shock dataset, displayed across all physical channels. While all models produce plausible results, ReFlow and the hybrid GenCFD \(\boldsymbol{\circ}\) FNO most closely align with the ground truth, particularly in preserving fine-scale detail. The spectral reconstruction effect of the hybrid model is especially evident, recovering high-frequency structures from the coarse FNO prediction and enhancing the overall visual fidelity.}
  \label{fig:cs_samples}
\end{figure}

\end{document}